\documentclass[12pt,a4paper]{report}

\usepackage{thesis}
\usepackage{lmodern}
\usepackage[numbers,sort&compress]{natbib}
\usepackage[hidelinks]{hyperref}
\usepackage[left=2.5cm,top=2.5cm,right=2.5cm,bottom=2.5cm]{geometry}

\usepackage[utf8]{inputenc}

\usepackage{latexsym}
\usepackage{mathrsfs}
\usepackage{amsmath}
\usepackage{amssymb}
\usepackage{graphicx}
\usepackage{algorithm}
\usepackage{algpseudocode}
\usepackage{multirow}
\usepackage{CJKutf8}
\usepackage{booktabs}
\usepackage{lipsum}
\usepackage{enumitem}
\usepackage{tabto}
\usepackage{mathrsfs}

\usepackage{mathptmx} % times new roman font
\usepackage[T1]{fontenc}

\usepackage{tocloft}

\newcommand{\thesistitle}{\textbf{Multilingual Transfer Learning for Code-Switched Language and Speech Neural Modeling}}
\newcommand{\thesisauthor}{\textbf{Genta Indra Winata}}
\newcommand{\programname}{Electronic and Computer Engineering}
\newcommand{\departmentname}{Department of Electronic and Computer Engineering}
\newcommand{\thesisdate}{April 2021}
\newcommand{\signdate}{April 2021}
\newcommand{\thesisdegree}{PhD}

\newcommand{\supervisorinfo}{Prof. Pascale Fung, Thesis Supervisor \\ Department of Electronic and Computer Engineering}
\newcommand{\depheadinfo}{Prof. Bertram Shi \\ Head, Department of Electronic and Computer Engineering}

\begin{document}

\pagenumbering{roman}
\pagestyle{plain}
\addcontentsline{toc}{chapter}{Title Page}
\thispagestyle{empty}
\null\vskip0.5in
\begin{center}
  \begin{LARGE}
    \thesistitle
  \end{LARGE}
  \vfill
  \vspace{20mm}

  \begin{Large}
  by
  \end{Large}

  \vspace{4mm}
  \begin{Large}
  \thesisauthor 
  \end{Large}\\
  \vfill
  \vspace{20mm}
  \begin{large}
  A Thesis Submitted to\\
  The Hong Kong University of Science and Technology \\
  in Partial Fulfillment of the Requirements for\\
  the Degree of Doctor of Philosophy \\
  in the Department of \programname \\
  \vfill \vfill
  \thesisdate, Hong Kong
  \end{large}
  \vfill
%   Copyright © Genta Indra Winata 2020
\end{center}

\vfill

\newpage
\addcontentsline{toc}{chapter}{Authorization Page}
\null\skip0.2in
\begin{center}
{\bf \Large \underline{Authorization}}
\end{center}
\vspace{12mm}

I hereby declare that I am the sole author of the thesis.

\vspace{10mm}

I authorize the Hong Kong University of Science and Technology to lend this
thesis to other institutions or individuals for the purpose of scholarly research.

\vspace{10mm}

I further authorize the Hong Kong University of Science and Technology to
reproduce the thesis by photocopying or by other means, in total or in part, at the
request of other institutions or individuals for the purpose of scholarly research.

\vspace{30mm}

\begin{center}
\underline{~~~~~~~~~~~~~~~~~~~~~~~~~~~~~~~~~~~~~~~~~~~~~~~~~~~~~~~~~~~~~~~~~~~~~~}\\
~~~~\thesisauthor \\
~~~~\signdate

\end{center}

\newpage
\addcontentsline{toc}{chapter}{Signature Page}
\begin{center}
{\Large \thesistitle}\\
\vspace{5mm}
by\\
\vspace{3mm}
\thesisauthor\\
\vspace{5mm}
This is to certify that I have examined the above PhD thesis\\
and have found that it is complete and satisfactory in all respects,\\
and that any and all revisions required by\\
the thesis examination committee have been made.
\end{center}

\vspace{12mm}

\begin{center}
\underline{~~~~~~~~~~~~~~~~~~~~~~~~~~~~~~~~~~~~~~~~~~~~~~~~~~~~~~~~~~~~~~~~~~~~~~~~~~~ }\\
\supervisorinfo
\end{center}

\vspace{12mm}
\begin{center}
\underline{~~~~~~~~~~~~~~~~~~~~~~~~~~~~~~~~~~~~~~~~~~~~~~~~~~~~~~~~~~~~~~~~~~~~~~~~~~~ }\\
\depheadinfo
\end{center}

\vspace{4mm}

\textbf{Thesis Examination Committee}
\TabPositions{5cm}
\begin{enumerate}[itemsep=0.3mm]
    \item Prof. Pascale Fung \tab Department of Electronic and Computer Engineering
    \item Prof. Bertram Shi \tab Department of Electronic and Computer Engineering
    \item Prof. Qifeng Chen \tab Department of Electronic and Computer Engineering
    \item Prof. Yanqiu Song  \tab Department of Computer Science and Engineering
    \item Prof. Daisy Yan Du \tab Department of Humanities
    \item Prof. Thamar Solorio \tab Department of Computer Science, University of Houston
\end{enumerate}

\vspace{5mm}
\begin{center}
\departmentname\\
\vspace{1mm}
\signdate
\end{center}

\newpage
\addcontentsline{toc}{chapter}{Acknowledgments}
\centerline{{\bf \Large Acknowledgments}} \vspace{5mm} \noindent

Firstly, I would like to thank my supervisor, Professor Pascale Fung, for guiding me and sharing her experiences throughout my time as a Ph.D. student. It has given me a life-changing experience. Initially, I was very impressed with her talks on empathetic conversation agents and AI for the social good. She keeps inspiring me to push myself to achieve huge impacts on society by researching in the natural language processing field. I am very proud to be a part of the research lab in these four and a half years, and if I look back, I would have never expected myself to publish numerous papers in top conferences. 

Next, I would like to express my appreciation to Professor Bertram Shi and Professor Qifeng Chen for taking their time to be on my thesis supervision committee, and Professor Yanqiu Song and Professor Daisy Yan Du for serving as my thesis examining committee. I am also thankful to Professor Thamar Solorio, who has inspired me with many of her seminal works in code-switching, and it is my pleasure to have her on my thesis committee. I hope she enjoys reading my thesis and following my defense presentation. I want to give my gratitude to Tania Leigh Wilmshurst, who proofread my papers and thesis countless times, and gave me insightful and useful advice on academic writing. I would also like to thank Dr. Steven Hoi and Dr. Guangsen Wang for allowing me to gain research and development experience at Salesforce. Both of them gave me in-depth reviews of my work and the freedom to choose my research project during my internship.

In the last four years, I have had an exciting journey in the lab with amazing lab friends and colleagues. I want to thank Andrea Madotto, Zhaojiang Lin, and Chien-Sheng Wu for the many collaborations over the years, and who I also had the opportunity to work with on dialogue systems. I also want to thank Zihan Liu, with whom I have worked closely on multilingual, cross-lingual, and cross-domain research. I always enjoyed our time having discussions and working together. Thanks go to Samuel Cahyawijaya, who has tremendously helped me in exploring new ideas on speech and efficient models and who gave me moral support during tough times, and also to Peng Xu for our insightful discussions on language generation and for showing me his unbeatable perseverance at research. I would like to thank Professor Ayu Purwarianti and Sidik Soleman from Prosa.Ai, Xiaohong Li, Zhi Yuan Lim, and Syafri Bahar from Gojek, and my colleagues Bryan Wilie, Karissa Vincentio, and Rahmad Mahendra for the large collaboration on IndoNLU that has become among the most significant work on Indonesian NLP. Thanks also go to Onno Kampman, who I worked closely with on the virtual psychologist; to Yan Xu, Yejin Bang, Elham Barezi, Etsuko Ishii, Jamin Shin, Dan Su, Farhad Bin Siddique, Anik Dey, Emily Yang, and Hyeondey Kim who were my collaborators in one of my research projects; to Nayeon Lee for proofreading my thesis; and to Wenliang Dai, Tiezheng Yu, Ji Ho Park, Naziba Mostafa, Dario Bertero, Ziwei Ji, Zihao Qi, and many others for the invaluable research experiences. 

I would like to express my gratitude to my friends, Eveline Nathalia, Yuliana Sutjiadi-Sia, Budianto Huang, Benedict Wong, Ilona Christy Unarta, Kharis Daniel Setiasabda, Wilson Lye, and many others who have motivated me and sent me prayers during all these challenging times. I enjoyed every moment that we spent together, especially hiking. Finally, I want to send my love to my parents and sister, who are the support system that provides me unconditional support and love. There were many challenging times, and without them, it would have been impossible for me to overcome the challenges. I dedicate this thesis to them.

\newpage
\addcontentsline{toc}{chapter}{Table of Contents}
\tableofcontents

\newpage
\addcontentsline{toc}{chapter}{List of Figures}
\listoffigures

\newpage
\addcontentsline{toc}{chapter}{List of Tables}
\listoftables

\newpage
\addcontentsline{toc}{chapter}{Abstract}
\begin{center}
{\Large \thesistitle}\\
\vspace{20mm}
by \thesisauthor\\
%\vspace{15mm}
\departmentname\\
%\vspace{10mm}
The Hong Kong University of Science and Technology
\end{center}
\vspace{8mm}
\begin{center}
Abstract
\end{center}
\par
\noindent
Multilingualism is the ability of a speaker to communicate natively in more than one language. In multilingual communities, switching languages within a conversation, called code-switching, commonly occurs, and this creates a demand for multilingual dialogue and speech recognition systems to cater to this need. However, understanding code-switching utterances is a very challenging task for these systems because the model has to adapt to code-switching styles. 

Deep learning approaches have enabled natural language systems to achieve significant improvement towards human-level performance on languages with huge amounts of training data in recent years. However, they are unable to support numerous low-resource languages, mainly mixed languages. Also, code-switching, despite being a frequent phenomenon, is a characteristic only of spoken language and thus lacks transcriptions required for training deep learning models. On the other hand, conventional approaches to solving the low-resource issue in code-switching are focused on applying linguistic theories to the statistical model. The constraints defined in these theories are useful. Still, they cannot be postulated as a universal rule for all code-switching scenarios, especially for languages that are syntactically divergent, such as English and Mandarin. 

In this thesis, we address the aforementioned issues by proposing language-agnostic multi-task training methods. First, we introduce a meta-learning-based approach, meta-transfer learning, in which information is judiciously extracted from high-resource monolingual speech data to the code-switching domain. The meta-transfer learning quickly adapts the model to the code-switching task from a number of monolingual tasks by learning to learn in a multi-task learning fashion. Second, we propose a novel multilingual meta-embeddings approach to effectively represent code-switching data by acquiring useful knowledge learned in other languages, learning the commonalities of closely related languages and leveraging lexical composition. The method is far more efficient compared to contextualized pre-trained multilingual models. Third, we introduce multi-task learning to integrate syntactic information as a transfer learning strategy to a language model and learn where to code-switch.

To further alleviate the issue of data scarcity and limitations of linguistic theory, we propose a data augmentation method using Pointer-Gen, a neural network using a copy mechanism to teach the model the code-switch points from monolingual parallel sentences, and we use the augmented data for multilingual transfer learning. We disentangle the need for linguistic theory, and the model captures code-switching points by attending to input words and aligning the parallel words, without requiring any word alignments or constituency parsers. More importantly, the model can be effectively used for languages that are syntactically different, such as English and Mandarin, and it outperforms the linguistic theory-based models. 
% We find that the Pointer-Gen model outperforms the linguistic theory-based models for languages that are syntactically different.

In essence, we effectively tackle the data scarcity issue by introducing multilingual transfer learning methods to transfer knowledge from high-resource languages to the code-switching domain, and we compare their effectiveness with the conventional methods using linguistic theories.

\newpage
\pagenumbering{arabic}
\pagestyle{plain}
\chapter{Introduction}

\section{Motivation and Research Problem}
Multilingualism is the ability of a speaker to communicate effectively in more than one language. It is an important skill for people nowadays, and it is believed that multilingual speakers, in fact, outnumber monolingual speakers~\cite{tucker1999global}. In multilingual communities, an interesting phenomenon called code-switching occurs, in which people alternate between languages and mix them within a conversation or sentence~\cite{poplack1980sometimes}. This linguistic phenomenon shows the ability of multilingual people to effortlessly switch between two or more languages when communicating with each other~\cite{bullock2009cambridge}. Code-switching is often found in countries with immigrants who speak a non-English language as their native language, and learn English as their second language. For example, in 2017, around 18\% of the population of the United States was Hispanic, and many of them speak Spanish and English~\cite{anderson2007attitudes,rothman2007linguistic}, while in Southeast Asian countries such as Singapore, Malaysia, and Indonesia~\cite{stymne2020evaluating}, many people come from diasporas that speak Mandarin Chinese, other Chinese dialects, English, Malay, Arabic, and Indian languages~\cite{ng2005ethnic}, and it is very common to find them combining languages during conversation. Code-switching is used in many human-to-human communications in social media~\cite{vyas2014pos,barman2016part}, while companies also use code-switching in advertisements~\cite{luna2005advertising}, radio~\cite{stavans2004spanglish}, and television programs~\cite{alvarez1998s} as a marketing strategy thought to be more persuasive to bilinguals.

The term ``code-switching" has no clear definition accepted by all linguists. According to Myers~\cite{myers1997duelling}, code-switching is the use of two or more languages in the same conversation or in the same sentence of that turn. The distinction between code-mixing, code-switching, and lexical borrowing is not clear~\cite{bali2014borrowing}. In this thesis, we will not distinguish between code-switching and code-mixing as the terms are usually used interchangeably following the definition in~\cite{poplack2000sometimes}. The phenomenon of code-switching can occur at different linguistic levels. At the phrase level, code-switching can occur across sentences, which is called inter-sentential code-switching. An example is the following Mandarin-English utterance:

\vspace{5mm}

\begin{CJK*}{UTF8}{gbsn}
\indent\textbf{Utterance: }\textnormal{我 } \textnormal{不 } \textnormal{懂 } \textnormal{要 } \textnormal{怎么 } \textnormal{讲 } \textnormal{一 } \textnormal{个 } \textnormal{小时 }. Seriously I didn't have so much things to say.

\indent\textbf{Translation: }\textit{I don’t understand how to speak for an hour}. Seriously I didn’t have so

\indent much things to say.
\end{CJK*}

\vspace{5mm}

\noindent At the word-level, code-switching can occur within a sentence, where it is called intra-sentential code-switching. An example is the following Spanish-English utterance:

\vspace{5mm}

\indent\textbf{Utterance: }\textit{Walking Dead} le quita el apetito a cualquiera. 

\indent\textbf{Translation:} \textit{Walking Dead} takes away the appetite of anyone. 

\vspace{5mm}

\noindent In this context, ``Walking Dead" is an English television series title, and it does not represent the literal meaning. Meanwhile, some words with the same spelling may have entirely different meanings (e.g., cola in English and Spanish) \cite{winata2018bilingual}. Language identifiers are commonly used to solve the word ambiguity issue in mixed-language sentences. However, they may not reliably cover all code-switching cases, and create a bottleneck by requiring large-scale crowdsourcing to correctly annotate language identifiers in code-switching data. 
In a language pair like Indonesian-English~\cite{stymne2020evaluating}, the mixing may even be found at the subword level, where prefixes or suffixes are added, such as the following:

\vspace{5mm}

\indent \textbf{Utterance: }Kesehatannya memburuk since deaths \textit{daughternya}

\indent \textbf{Translation:} She is not doing so well since the death \textit{of her} daughter.

\vspace{5mm}

\begin{figure}[!ht]
    \centering
    \includegraphics[width=\linewidth]{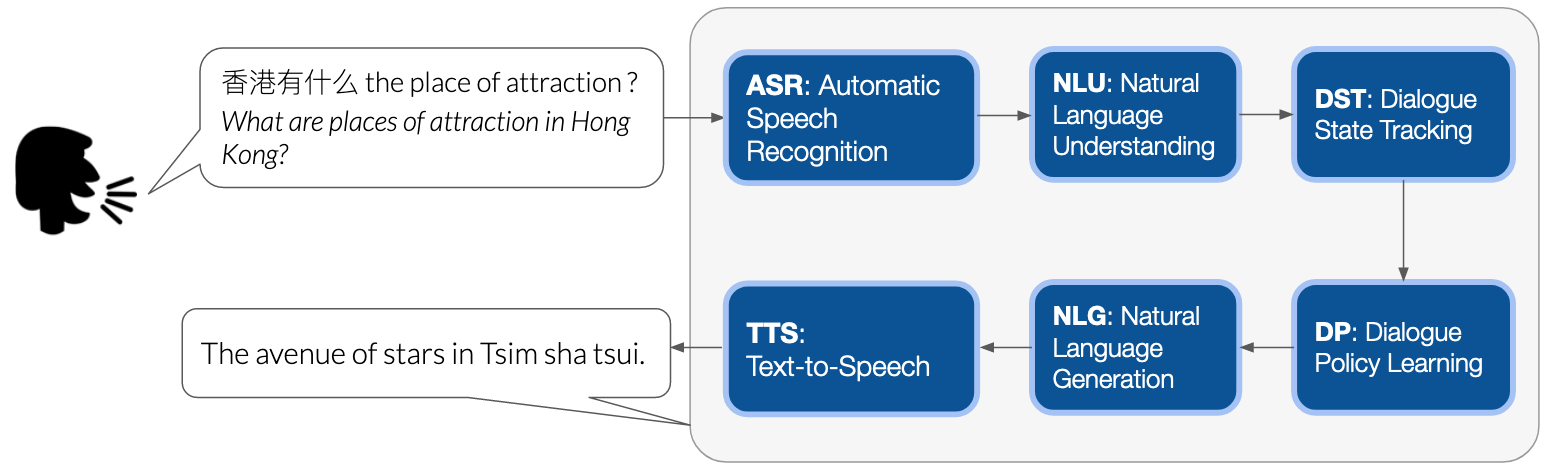}
    \caption{End-to-end flow of the goal-oriented dialogue system.}
  \label{fig:goal-oriented}
\end{figure}

Despite the enormous number of studies in natural language processing (NLP), only very few specifically focus on code-switching. However, NLP research on code-switching has been slowly growing due to increased interest in applications of multilinguality. The ultimate goal of research in multilinguality is to build conversational agents that are able to understand utterances from multilingual speakers and respond appropriately depending on the context. Figure~\ref{fig:goal-oriented} shows the pipeline of a goal-oriented dialogue system. First, the automatic speech recognition (ASR) module has to transcribe speech utterances into text to know what the user says. Then, the text is passed into the following modules to understand the text and generate an appropriate response to send back to the user. After the ASR, the natural language understanding (NLU) module captures important named entities and slots at the word level~\cite{liu2019zero}. These entities are then used in the dialogue state tracking (DST) module to remember the context of a dialogue~\cite{lin2020mintl}. Subsequently, the natural language generation (NLG) module uses the information extracted from the text and generates a response for the user, and the text-to-speech (TTS) module translates the text response into an audio signal.

The following are the major challenges to developing code-switching models:

\begin{itemize}
    \item[1.] \textbf{Incorporating linguistics theory:} Most research on code-switching focuses on finding constraints in the way monolingual grammars interact with each other to produce well-formed code-switched speech, and building code-switching grammars so that linguists can understand how code-switches are triggered. Using this knowledge, we can generate synthetic code-switching sentences as weak signals for the model, and thereby boost the performance of code-switching models~\cite{pratapa2018language,soto2020identifying,li2011asymmetric, li2012code}. 
    \item[2.] \textbf{Leveraging monolingual data:} Lack of data is a critical issue for training code-switching models. With the rise in the number of multilingual speakers, speech recognition systems that support different languages are in demand. However, most such systems are unable to support numerous low-resource languages, particularly mixed languages. The data scarcity of low-resource languages has been a major challenge for dialogue and speech recognition systems since they require a large amount of data to learn a robust model, and collecting training data is expensive and resource-intensive. A number of previous studies have used monolingual data as training signals for transfer learning, and these data can also be used in the form of pre-training.
    \item[3.] \textbf{Improving code-switching representations:} Learning a model to understand mixed language text or speech is very important to building better bilingual or multilingual systems that are robust to different language mixing styles. This will benefit dialogue systems and speech applications, such as virtual assistants~\cite{jose2020survey}. However, training a robust code-switching ASR model has been a challenging task for decades due to data scarcity. One way to enable low-resource language training is by first applying transfer learning methods that can efficiently transfer knowledge from high-resource languages~\cite{pan2009survey} and then generating synthetic speech data from monolingual resources~\cite{nakayama2018speech,winata2019code}. However, these methods are not guaranteed to generate natural code-switching speech or text. Another line of work explores the feasibility of leveraging large monolingual speech data in the pre-training, and applying fine-tuning on the model using a limited source of code-switching data, which has been found useful to improve performance~\cite{li2011asymmetric,winata2019code}. However, the transferability of these pre-training approaches is not optimized to extract useful knowledge from each individual languages in the context of code-switching, and even after the fine-tuning step, the model forgets the previously learned monolingual tasks. One of the most intuitive ideas to create a multilingual representation is using pre-trained multilingual language models, such as multilingual BERT~\cite{pires2019multilingual}, as a feature extractor. However, these models are not trained for code-switching, which makes them a poor option for this case. 
\end{itemize}

Traditionally, the approach to solving the low-resource issue in code-switching is to apply linguistic theories to the statistical model. Linguists have studied the code-switching phenomenon and proposed a number of theories, since code-switching is not produced indiscriminately, but follows syntactic constraints~\cite{poplack1978syntactic,pfaff1979constraints,poplack1980sometimes, belazi1994code}. Linguists have formulated various constraints to define a general rule for code-switching.
However, these constraints cannot be postulated as a universal rule for all code-switching scenarios, especially for languages that are syntactically divergent \cite{berk1986linguistic}, such as English and Mandarin, since they have word alignments with an inverted order. Variations of code-switching also exist, and many of them are influenced by traditions, beliefs, and normative values in the respective communities~\cite{bhatt2011code}. Studies describe that code-switching is dynamic across communities or regions and each has its own way to mix languages~\cite{chen2005social,wolfram2015american}. Thus, building a statistical code-switching model is challenging due to complexity in the grammatical structure and localization of code-switching styles. Another shortcoming of this approach is the limitation of syntactic parsers for mixed language sentences, which are currently unreliable. 

In this thesis, we address the different challenges mentioned above. We propose language-agnostic approaches that are not dependent on particular languages to improve the generalization of our models on code-switched data. We first introduce a multi-task learning to benefit from syntactic information in neural-based language models, so that the models share a syntactical representation of languages to leverage linguistic information and tackle the low-resource data issue. Then, we present two data augmentation methods to obtain synthetic code-switched training data by (1) aligning parallel sentences and applying linguistic constraints to check valid sentences and (2) using a copy-mechanism to learn how to generate code-switching sentences by sampling from the real distribution of code-switching data. The copy mechanism learns how to combine words from parallel sentences and identifies when to switch from one language to the other. We add the generated data on top of the training data and explore several fine-tuning strategies to improve code-switched language models. Next, we introduce new meta-embedding approaches to effectively transfer information from rich monolingual data to address the lack of code-switching data in different downstream NLP and speech recognition tasks. We introduce Meta-Transfer Learning to transfer-learn on a code-switched speech  recognition system in a low-resource setting by judiciously extracting  information from high-resource monolingual datasets. Finally, we propose a new representation learning method to represent code-switching data by learning how to transfer and acquire useful knowledge learned from other languages. 
% The final goal of this thesis is to introduce novel statistical approaches that have been investigated with end-to-end deep learning models. 

% First, we contextualize ideas from linguistic theory and incorporate them into neural networks using a copy mechanism to augment code-switching data that is useful for the speech recognition task. 

% Later, we also propose a new representation learning method to represent code-switching data by learning how to transfer and acquire useful learned knowledge from other languages. The final goal of this thesis is to introduce novel statistical approaches that have been investigated with end-to-end deep learning models.

\section{Thesis Outline}
The contents of this thesis are organized around code-switching, and our experiments are focused on code-switching NLP and speech tasks. The rest of the thesis is divided into six chapters and organized as follows:
\begin{itemize}
    \item Chapter 2 introduces the background and important related work on linguistic theories of code-switching. Then, we discuss applications of code-switching, such as language modeling, speech recognition, and sequence labeling, and we also explain how to compute code-switching complexity. This chapter presents the fundamentals required to understand the rest of the thesis.
    \item Chapter 3 examines approaches to training language models for code-switching by leveraging linguistic theories and neural network language models. We propose a data augmentation method to increase the variance of the corpus with linguistic theory and a model-based approach.
    \item Chapter 4 presents approaches to train language models in a multi-task training that leverages syntactic information. We train our model by jointly learning the language modeling task and part-of-speech 
    (POS) sequence tagging task on code-switched utterances. We incorporate language information into POS tags to create bilingual tags that distinguish between languages.
    % the multi-task syntax-aware method to improve code-switching language modeling by leveraging syntactic information. We further extend the method by generating synthetic code-switching data by using linguistic theories and copy mechanism. 
    \item Chapter 5 introduces approaches to train code-switching speech recognition by transfer learning methods. Our methods apply meta-learning by judiciously extracting information from high-resource monolingual datasets. The optimization conditions the model to retrieve useful learned information that is focused on the code-switching domain.
    % transfer learning methods using monolingual corpus to improve code-switched speech recognition.
    \item Chapter 6 discusses the state-of-the-art multilingual representation learning methods for code-switched named entity recognition (NER). We introduce meta-embeddings, considering the commonalities across languages and compositionality. We find that this method is language-agnostic, and it is very effective and efficient compared to large contextual language models on the code-switching domain. We also conduct a study to measure the effectiveness and efficiency of the multilingual models to see their capability and adaptability in the code-switching setting.
    \item Chapter 7 summarizes this thesis and the significance of the transfer learning approaches, and discusses possible future research directions.
\end{itemize}
\chapter{Background and Preliminaries}

\section{Overview}
In this chapter, we provide a literature review and background knowledge on the theoretical and computational linguistics aspects of code-switching that are fundamental to our work. We present the common linguistic models on code-switching that can later be applied to statistical models. We also introduce contemporary studies on NLP and speech recognition with code-switching data, such as language identification, language modeling, speech recognition, and sequence tagging. Lastly, we show metrics to evaluate and measure the code-switching complexity of a corpus.

\section{Linguistic Models of Code-Switching}

Early studies on code-switching formalize linguistic assumptions about how people learn to switch from one language to another by observing the grammar and syntax. The rules mainly compare the grammars of two languages, and find the asymmetric relations between them. They investigate trigger words that activates the switch, such as POS tags and proper nouns that have strong relationships with code-switching~\cite{soto2018role}. From the qualitative perspective, there are three well-established linguistic theories that are commonly used in the linguistic world~\cite{sitaram2019survey}: (1) the free morpheme constraint, 
(2) the matrix language-frame model, and (3) the equivalence constraint. These three theories will be discussed in the following. 
% We also describe the constraint-free approach, such as null grammar.

\subsection{Free Morpheme Constraint}
Morphemes can be classified into two types: free morphemes and bound morphemes. Free morphemes are those that can stand alone to function as words, while bound morphemes are those that can only be attached to another part of a word. \citet{poplack1980sometimes} proposes that code-switching will occur in any constituent in discourse if that constituent is not a bound morpheme. Basically, this work states a condition that a code-switch may occur between a free and a bound morpheme if and only if the bound morpheme is phonologically integrated into the language. For example, in Spanish-English,\footnote{The examples are taken from https://www.ukessays.com/essays/languages/grammatical-constraints-on-code-switching.php}the utterance

\vspace{5mm}

\indent \textbf{Utterance: }And what a tertuliait was, Dios mio!

\indent \textbf{Translation:} And what a \textit{gathering it} was, \textit{my God!}.

\vspace{5mm}

\noindent is acceptable under the free morpheme constraint, since ``Dios mio" is a bound morpheme, unlike the following sentence:

\vspace{5mm}

\indent \textbf{Utterance: }*Estaba type-ando su ensayo.

\indent \textbf{Translation:} \textit{She was} type-\textit{ing her essay}.

\subsection{Equivalence Constraint}
\begin{CJK*}{UTF8}{gbsn}
We define the dominant language as the matrix language and the contributing language as the embedded language. According to the equivalence constraint (EC) theory, code-switches will tend to occur at points in a discourse where the juxtaposition of the matrix language (\textit{$L_1$}) and embedded language (\textit{$L_2$}) elements does not violate a syntactic rule of either language~\cite{poplack1980sometimes,poplack2013sometimes}. For example, we have an example of parallel sentences in English and Chinese:
\vspace{5mm}

\indent \textbf{\textit{$L_1$} (English): }this is actually belonged to simplified chinese

\indent \textbf{\textit{$L_2$} (Chinese):} \text{这个} \text{其实} \text{是} \text{属于} \text{简体} \text{中文}.

\vspace{5mm}

\begin{figure}[!ht]
    \centering
    \includegraphics[width=0.7\linewidth]{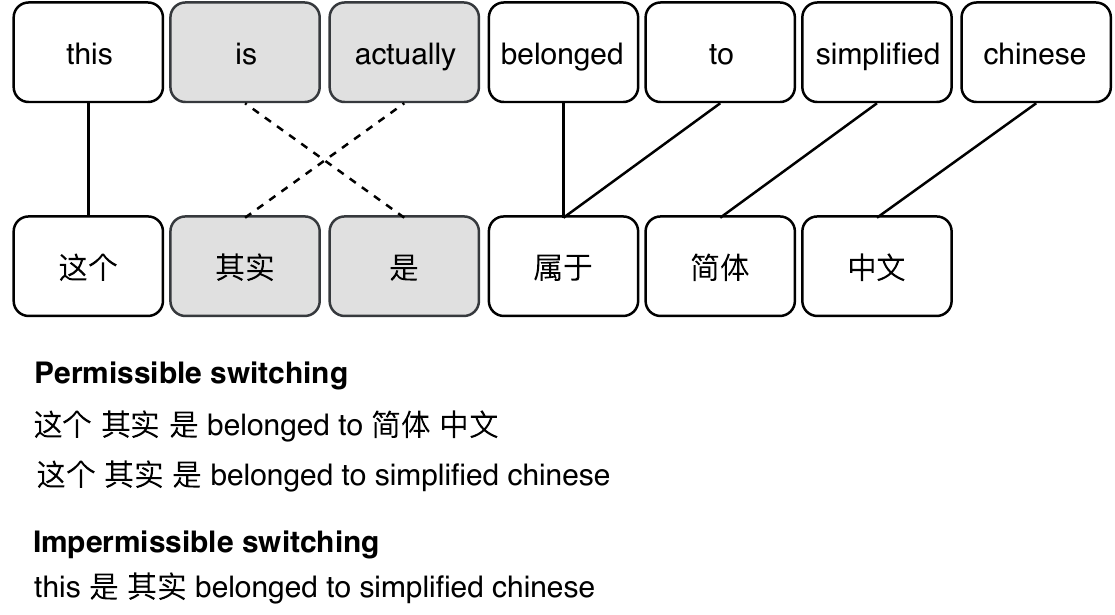}
    \caption{Example of equivalence constraint using aligner on English-Chinese data.}
    \label{fig:eq-contraint-example}
\end{figure}

\noindent We align words for which the constituents from the two languages map onto each other~\cite{poplack1980sometimes}. Figure~\ref{fig:eq-contraint-example} shows an example where the switches are permissible.  Solid lines show the alignment between the matrix language (top) and the embedded language (bottom).~The dotted lines denote impermissible switching.

We can also interpret the EC theory by defining context-free grammars \textit{$G_1$} and \textit{$G_2$} of \textit{$L_1$} and \textit{$L_2$}, respectively. This assumes that every non-terminal category in G1 has a corresponding non-terminal category in G2, as stated by~\citet{pratapa2018language}. In their implementation, the sentences are parsed using a monolingual syntactic parser. 

\vspace{5mm}

\indent \textbf{\textit{$L_1$} (English):} She lives in a white house

\indent \textbf{\textit{$L_2$} (Spanish):} Elle vive en una casa blanca.

\vspace{5mm}

\begin{figure}[!ht]
    \centering
    \includegraphics[width=\linewidth]{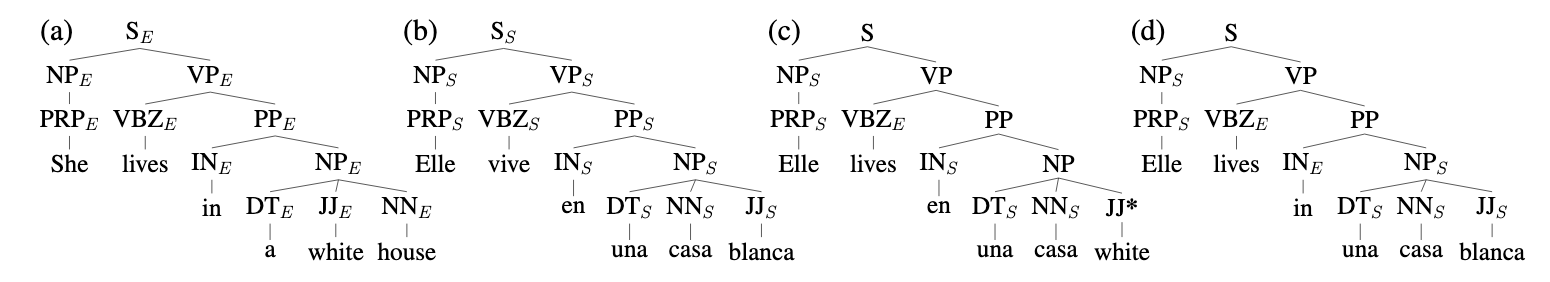}
    \caption{Example of equivalence constraint on English-Spanish using syntactic parsers~\cite{pratapa2018language}.}
    \label{fig:eq-contraint-example-english-spanish}
\end{figure}

\noindent Figure~\ref{fig:eq-contraint-example-english-spanish} represents the sentences above in a tree form, where (a) is the original English sentence, (b) the original Spanish sentence, (c) the incorrectly code-mixed sentence, and (d) the correctly code-mixed sentence that satisfies the EC theory.

However, this method relies on the quality of the parser, and it is difficult to apply to languages with distant grammatical structures, like English and Chinese, due to the large difference in the syntactic tags.

% Figure~\ref{fig:eq-contraint-example} shows an example where the switch between ``is" and ``是" or ``actually" and ``其实" is invalid since there is an overlap between alignments. Formally, we can define alignments between an $L_1$ sentence and an $L_2$ sentence comprise a source vector with indices~$u_t = \{a_1, a_2, ..., a_m\} \in \mathbb{W}^m$ that has a corresponding target vector $v_t = \{b_1, b_2, ..., b_m\} \in \mathbb{W}^m$, where $u$ is a sorted vector of indices in an ascending order. The alignment between $a_i$ and $b_i$ does not satisfy the constraint if there exists a pair of $a_j$ and $b_j$ where ($a_i < a_j$ and $b_i  > b_j$) or ($a_i > a_j$ and $b_i  < b_j$). If the switch occurs at this point, it changes the grammatical order in both languages; thus, this switch is not acceptable. 

\end{CJK*}

\subsection{Matrix Language-Frame Model} 
The matrix language-frame (MLF) model states the grammatical procedures to produce code-switching sentences~\cite{myers1995social}. The model defines the matrix language as the primary language (L1) and embedded language as the embedded language (L2) that is inserted into the morphosyntactic frame of the matrix language. The MLF model defines three different constituents: L1 islands, L2 islands, and both L1 and L2 constituents. This framework also introduces the blocking hypothesis, which states that any embedded language morpheme, which is not congruent with the matrix language is blocked.

\section{End-to-end Speech Recognition}

The ASR system is the first step in the pipeline of systems in applications such as conversational agents, so any errors made by the ASR system can propagate through the wider system and lead to failures in interactions. Attempts have been made to approach the problem of code-switched ASR from the acoustic, language, and pronunciation modeling perspectives. The research on speech recognition has been further developed from modularized systems into end-to-end systems that combine acoustic, pronunciation, and language models into a single monolithic model. The recent advancement in sequence-to-sequence models has shown promising results in training monolingual ASR systems. Two common architectures are widely used for end-to-end speech recognition: connectionist temporal classification (CTC)~\cite{graves2006connectionist} and encoder-decoder models~\cite{ChanJLV15,dong2018speech,winata2019code,winata2020lightweight}. Further explorations on end-to-end architectures have been made by jointly training both of these architectures in a multi-task hybrid fashion~\cite{KimHW16,Winata2020AdaptandAdjustOT}, and they have found that these architectures improve the performance of the overall model. 
% In this thesis, we will focus on the Encoder-Decoder model that is used in the following chapters.

% One approach is to identify the language boundaries and subsequently use a monolingual ASR system to recognize monolingual fragments [102]. Another approach runs multiple recognizers in parallel with a LID system and uses scores from all the systems for decoding speech [103]. In [56], no LID system is used - instead, two recognizers in English and Malay are run in parallel, and the hypotheses pro
\subsection{RNN-based Encoder-Decoder Model}
The first-ever proposed end-to-end encoder-decoder ASR model, Listen-Attend-Spell (LAS)~\cite{chan2016listen}, uses a recurrent neural network (RNN) as its encoder and decoder components. The LAS model introduces two components, a listener and speller, with the former acting as an acoustic model that accepts filter bank spectra as inputs, and the latter working as the decoder that emits character outputs. 

\paragraph{Listener}
The listener uses a bidirectional long short-term memory (BiLSTM)~\cite{hochreiter1997long,graves2013hybrid} with a pyramidal structure to reduce the length of the audio frames. The pyramid BiLSTM reduces the time resolution by a factor of 2, similar to Clockwork RNN~\cite{koutnik2014clockwork}, to address the slow convergence issue. We concatenate the outputs at consecutive steps of the current layer before feeding the concatenation to the next layer as follows:
\begin{align}
    h_i^j = \text{BiLSTM}(h)_{i-1}^j,[h_{2i}^{j-1},h_{2i+1}^{j-1}]),
\end{align}
where $\mathbf{h}=(h_1,h_2,...h_{U-1},h_U)$ is the high-level representation, and $U\leq T$, where $T$ is the length of the input sequence.

\paragraph{Speller}
The speller is an attention-based long short-term memory (LSTM) that produces a probability distribution over the next character, conditioned on previous characters. The further equations are as follows:
\begin{align}
    c_i = \text{Attention}(s_i, \mathbf{h}),\\
    s_i = \text{RNN}(s_{i-1}, y_{i-1}, c_{i-1}),\\
    P(y_i|\mathbf{x},y_{<i}) = \text{Output}(s_i,c_i).
\end{align}
At each time step, the attention mechanism will generate a context vector $c_i$, and the context $c_i$ contains the information of the acoustic signal that is required to generate the next character. Then, the RNN module will generate a new decoder state $s_i$ and pass it to the output layer to generate a character distribution $P(y_i|\mathbf{x},y_{<i})$. The model can be trained jointly to maximize the log probability as follows:
\begin{align}
\max_{\theta}\sum_{i}{\log P(y_i|\mathbf{x},y^*_{<i};\theta)},
\end{align}
where $y^*_{<i}$ is the true labels of the previous characters. We always give transcripts for predicting the next character. In the inference time, the model finds the most likely character sequence given the audio frame inputs. The decoding process is to search the most likely sequence of given words:
\begin{align}
    \hat{\mathbf{y}} = \text{arg} \max_{\mathbf{y}} \log P(\mathbf{y}|\mathbf{x}).
\end{align}
We add a start of sentence token, \texttt{<SOS>}, at each time step, and apply a beam-search to collect a set of hypotheses that have an end of sentence token \texttt{<EOS>}. We rescore our beams with a language model score $P_{\text{LM}}(\mathbf{y})$ to improve the predictions as follows:
\begin{align}
    s(\mathbf{y}|\mathbf{x}) = \alpha P(\mathbf{y}|\mathbf{x}) + \beta \log P_{\text{LM}}(\mathbf{y}),
\end{align}
where $\alpha$ is the ASR model weight and $\beta$ is the language model weight.

\subsection{Transformer-based Encoder-Decoder Model}
\begin{figure}[!ht]
    \centering
    \includegraphics[width=0.45\linewidth]{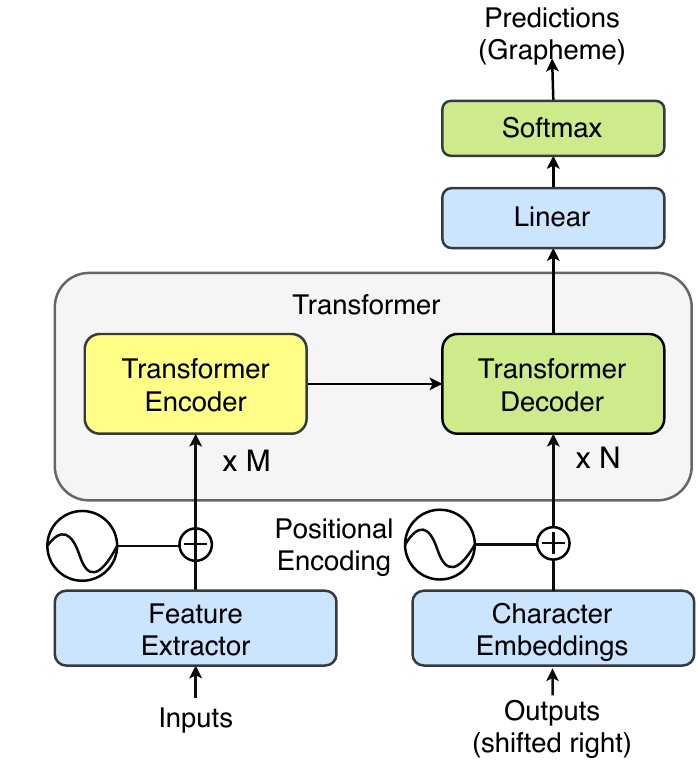}
    \caption{Transformer ASR model architecture.}
  \label{fig:transformer-asr}
\end{figure}

In recent studies~\cite{dong2018speech,winata2019effectiveness,winata2019code}, transformers replace RNN modules in the ASR system since it is more efficient in terms of time and memory complexity. The RNN model has issues in parallelization, and it has to do a recursive function through the sequence length. The transformer is able to cut this dependency, allows the input to be processed in parallel, and uses an attention mechanism to attend to all tokens. Similar to the RNN model, the transformer-based model also predicts target tokens and accepts speech features as input. The transformer-based ASR architecture is illustrated in Figure~\ref{fig:transformer-asr}.
In the following, we describe the structure of the encoder and decoder layer.

\begin{figure}[!ht]
\centering
  \includegraphics[width=0.3\linewidth]{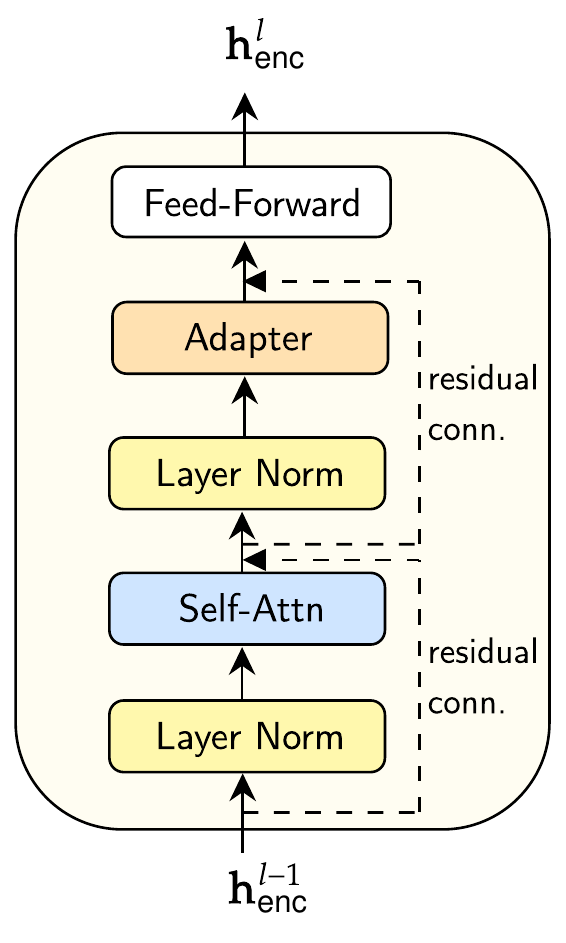} 
  \caption{Encoder transformer structure.}
  \label{fig:enc}
\end{figure}

\paragraph{Encoder Layer}
Figure~\ref{fig:enc} shows the structure of the encoder layer. An adapter layer is added  after the layer norm and self-attention module. We also apply two residual connections after both the self-attention layer and the adapter layer: 
\begin{align}
    \mathbf{o} &= \text{SelfAttn}(\text{LayerNorm}(\mathbf{h}_{\text{enc}}^{l-1})) + \mathbf{h}_{\text{enc}}^{l-1},\\
    \mathbf{h}_{\text{enc}}^l &= \text{FeedForward}(\text{LayerNorm}(\mathbf{o}) + \mathbf{o}),
\end{align}
where $\mathbf{h}_\text{enc}^{l-1}$ is the encoder hidden states of the previous layer $l-1$ and $\mathbf{h}_{\text{enc}}^l$ is the output of the encoder layer.

\begin{figure}[!ht]
\centering
  \includegraphics[width=0.3\linewidth]{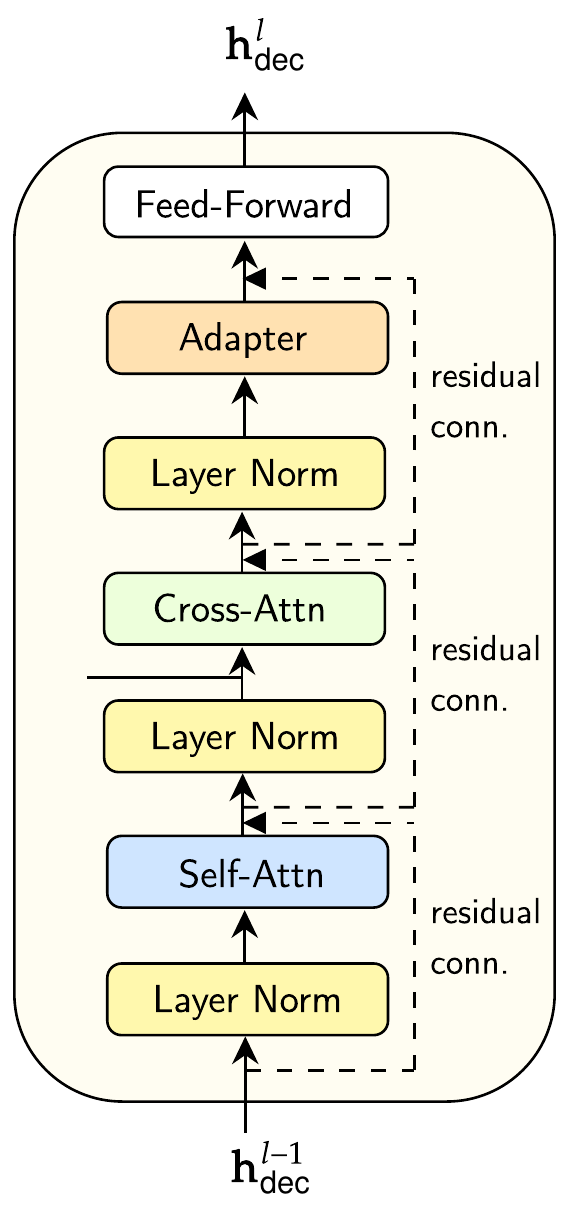} 
  \caption{Decoder transformer structure.}
  \label{fig:dec}
\end{figure}

\paragraph{Decoder Layer}
Figure~\ref{fig:dec} shows the structure of the decoder layer with the adapter. We place the adapter layer after the cross-attention model:
\begin{align}
    \mathbf{o}_1 &= \text{SelfAttn}(\text{LayerNorm}(\mathbf{h}_{\text{dec}}^{l-1})) + \mathbf{h}_{\text{dec}}^{l-1},\\
    \mathbf{o}_2 &= \text{CrossAttn}(\mathbf{h}_{\text{enc}}, \text{LayerNorm}(\mathbf{o}_1)) + \mathbf{o}_1,\\
    \mathbf{h}_{\text{dec}}^{l+1} &= \text{FeedForward}(\text{LayerNorm}(\mathbf{o}_2) + \mathbf{o}_2),
\end{align}
where $\mathbf{h}_{\text{dec}}^{l-1}$ is the decoder hidden states of the previous layer, and $\mathbf{h}_{\text{dec}}^{l}$ is the output of the current layer.

The transformer model has three main components: a scaled dot-product attention, multi-head attention, and positional-wise feed-forward network.

\paragraph{Scaled Dot-Product Attention}
The attention function is described as a mapping between a query and a set of key-value pairs to an output~\cite{vaswani2017attention}. The weight calculated from the function is assigned to the value. Particularly for the scaled dot-product attention, we apply dot products of the queries of dimension $d_q$ and keys of dimension $d_k$. Then, we divide each of them by $\sqrt{d_k}$ and apply the softmax function to compute the weights. We use the weights to weigh the values of dimension $d_v$. Normally, we set $d_q = d_k$, and $d_{model}$ is divisible by the number of heads $h$.
\begin{align}
    \text{Attention}(Q,K,V) = \text{Softmax}(\frac{QK^T}{\sqrt{d_k}})V.
\end{align}

\paragraph{Multi-head Attention} Instead of using a single-head attention, we can apply multiple $h$ heads that learn the linear projections of the query, key, and value. As suggested by~\citet{vaswani2017attention}, multi-head attention is more beneficial in learning representation subspaces at different positions:
\begin{align}
    \text{MultiHead}(Q,K,V) &= \text{Concat}(\text{head}_1,...,\text{head}_h) W^O,\\
    \text{head}_i &= \text{Attention}(QW^Q_i, KW^K_i, VW^V_i),
\end{align}
where the projection matrices $W_i^Q \in \mathbb{R}^{d_{model}
\times d_q}$, $W_i^K \in \mathbb{R}^{d_{model}
\times d_k}$, $W_i^V \in \mathbb{R}^{d_{model}
\times d_v}$, and $W_i^O \in \mathbb{R}^{d_{h d_v}
\times d_{model}}$. 

\paragraph{Position-Wise Feed-Forward Network}

After the attention layers in the encoder and decoder, we add a linear layer that is applied to each position and a ReLU activation. The layer will project the input into $d_{model}$. During the implementation, we can use two convolutions with a kernel size of 1:
\begin{align}
\text{FFN}(x) = \text{max}(0,xW_1+b_1)W_2+b_2.
\end{align}
\begin{figure*}[!ht]
  \centering
  \includegraphics[width=0.75\linewidth]{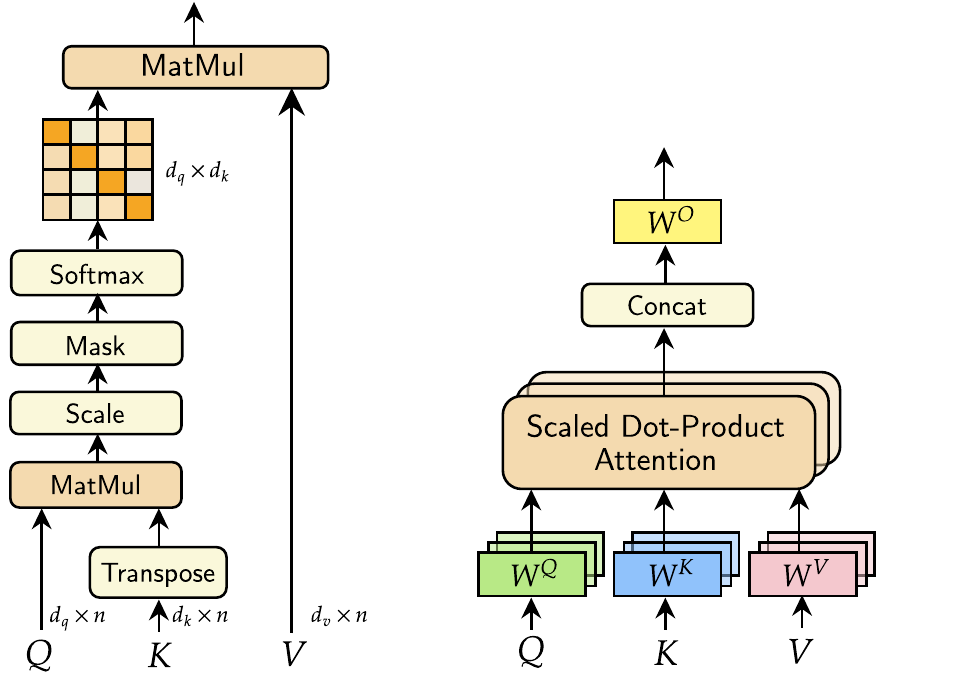}
  \caption{Transformer components. (left) Scaled dot-product attention. (right) Multi-head attention.}
  \label{fig:transformer}
\end{figure*}
For the training, we follow the same training objective as the RNN-based model, and we apply the same inference procedure.
% \paragraph{Phone merging}
% ~\cite{vu2012first,sivasankaran2018phone}
% \paragraph{Transliteration} ~\cite{emond2018transliteration}
% \paragraph{Semi-Supervised Learning}
% ~\cite{nakayama2018speech}
% \paragraph{End-to-end ASR}
% ~\cite{winata2018towards}

\subsection{Code-Switched Speech Recognition}
Since code-switching is a spoken language phenomenon, it is important for ASR systems to be able to handle code-switching speech. The initial attempts at code-switched ASR systems, focus on building HMM-based systems, where the ability to recognize the language speech frames is based on finding the language boundaries using language identification (LID). The detected monolingual frames are then passed into monolingual ASR systems~\cite{chan2004detection,weiner2012integration}. One way to identify the language boundaries is to recognize the monolingual speech fragments~\cite{chan2004detection}. Another approach to detecting code-switching speech is to map phone sets of code-switched language pairs into IPA mappings and construct a bilingual phone set and clustered phones. This approach has been conducted on Mandarin-English~\cite{yu2003chinese}, Ukrainian-Russian~\cite{lyudovyk2014code}, and Hindi-English~\cite{sivasankaran2018phone}. By applying this method, we can train code-switched ASR in a single model.

\subsection{Evaluation Metrics} 
Word error rate (WER) is the standard metric for evaluating speech recognition systems, and it is determined as follows:
\begin{align}
    WER = \frac{INS + DEL + SUB}{N} \times 100\%,
\end{align}
where $INS$, $DEL$, and $SUB$ are the number of insertions, deletions, and substitutions, respectively, and $N$ is the number of words in the reference. However, for some languages, such as Chinese, or in the mixed language setting, character error rate (CER) is used instead of WER. This metric calculates the distance between two sequences as the Levenshtein distance, and the idea of using it is to minimize the distance between two tokens or the number of insertions, deletions, or substitutions required to change one token into another. Computing CER is very similar to WER, but instead of computing a word as the token, we use a character. For our code-switched ASR, we compute the overall CER and also show the individual CERs for each language.

\section{Language Modeling}

Language modeling (LM) is a fundamental task in NLP and speech systems, most notably in ASR. Basically, an LM score indicates how well the model predicts a sequence of words. A higher score means a sequence is more likely to appear in the corpus. In the ASR task, LM is utilized to enhance the speech recognition results by re-scoring the hypothesis and maximizing the posteriors. The LM assigns a joint probability over all possible word sequences $\mathbf{w}$, $P_{\text{LM}}(\mathbf{w})$. We can compute the probability by using Bayes' theorem:
\begin{align}
    P_{\text{LM}}(\mathbf{w}) = P(w_1,w_2,...,w_n) = \prod_{i=1}^{n} P(w_i|w_{<i}) = P(w_1) P(w_2|w_1) ... P(w_i|w_1,w_2,...w_{i-1}).
\end{align}

\subsection{n-gram Language Modeling}
The n-gram LM applies the Markov assumption to estimate the probability of future units without looking too far into context. The assumption is that the probability of a word only depends on the previous word. The general equation for the n-gram approximation is:
\begin{align}
    P(w_1^n) \approx \prod_{k=1}^n P(w_k|w_{k-1}).
\end{align}
To compute the probability of a word $w$, we can sum all n-gram counts that start with that word, and normalize the sum by the unigram count for word $w_{n-1}$ so the number of counts lies between 0 and 1:
\begin{align}
    P(w_n|w_{n-1})=\frac{C(w_{n-1}w_n)}{C(w_{n-1})},
\end{align}
where $C(.)$ is the number of word combinations appearing in the data. However, this method still suffers from issues with unknown words in a test set in an unseen context. For example, when $n=3$, the probability of the word is conditioned on the previous two words. Thus, we compute the probability as follows:
\begin{align}
    P(w_i|w_{i-1},w_{i-2})=\frac{C(w_i,w_{i-1},w_{i-2})}{C(w_{i-1},w_{i-2})}.
\end{align}

Assigning an unknown token with zero probability is similar to ignoring the token; thus, it is not the best choice. We can apply a smoothing approach to assign a small probability to unknown tokens using smoothing techniques, such as Kneser-Ney~\cite{james2000modified}, Laplace, or Backoff smoothing.

\subsection{Neural-Based Language Modeling}
Instead of counting the frequency and using a fixed window, as in the n-gram LM, we can enlarge the window by using an RNN. The RNN model memorizes the context of the previous word and passes it to the next-token prediction. Here, we describe two neural-based LMs using an RNN and LSTM~\cite{schmidhuber1997long}, an extension of the RNN model.

\subsubsection{RNN language model}
\begin{figure*}[!ht]
  \centering
  \includegraphics[width=\linewidth]{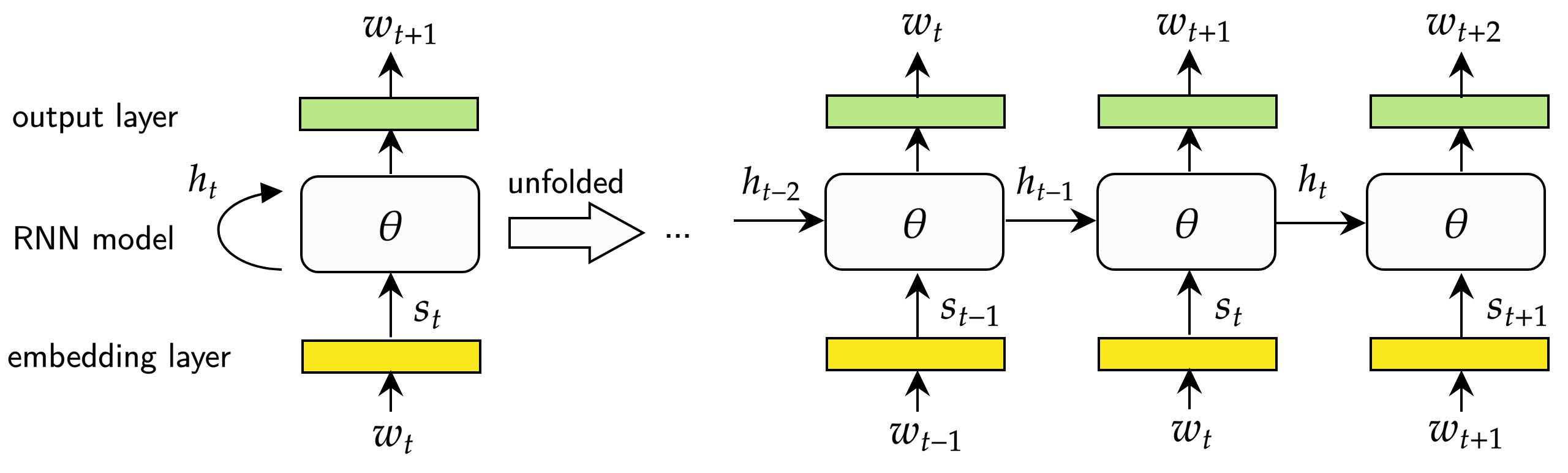}
  \caption{RNN language model.}
  \label{fig:rnnlm}
\end{figure*}

Neural-based language models normally use an RNN~\cite{kombrink2011recurrent,mikolov2011extensions,winata2019effectiveness} as the base model. Figure~\ref{fig:rnnlm} shows the unidirectional language model architecture, and we denote the model as $\theta$. Here we will also denote our input sequences as  $\mathbf{x} = (x_1, x_2, ..., x_{n-1}, x_n)$. 

\noindent A vector $\mathbf{s}_t$ is generated by mapping token $\mathbf{x}_t$ to the embedding layer $E$. The model generates an output vector $\mathbf{o}_t \in \mathbb{R}^{
V|}$, where $|V|$ is the size of the vocabulary. Vector $\mathbf{h}_t$ represents the output value in the hidden layer from the previous time step and the vector is projected to the dimension as the size of the vocabulary. The softmax function is applied to $\mathbf{o}_t$ for normalization to output the probability distribution. The RNN model is trained using backpropagation. The outputs of the layer are computed as follows:
\begin{align}
    \mathbf{s}_t &= E\mathbf{x}_t,\\
    \mathbf{h}_t &= \sigma(\mathbf{W}_h\mathbf{h}_{t-1}+ \mathbf{W}_i s_{t} + b_1),\\
    \mathbf{o}_t &=\mathbf{W}_o\mathbf{h}_t + b_2,\\
    \mathbf{o}_t &= \text{Softmax}(\mathbf{o}_t),\\
    \text{Softmax}(z) &= \frac{e^{z}}{\sum_{k}e_k^{z}},
\end{align}
where $\mathbf{W}_h$ and $\mathbf{W}_i$ are the learned weights. 
\subsubsection{LSTM language model}
LSTM language models~\cite{schmidhuber1997long,sundermeyer2012lstm,winata2019effectiveness}
are parameterized with two large matrices, $\mathbf{W}_i$, and $\mathbf{W}_h$. The LSTM captures long-term dependencies in the input and avoids the exploding/vanishing gradient problems of the standard RNN. The gating layers control the information flow within the network and decide which information to keep, discard, or update in the memory. The following recurrent equations show the LSTM dynamics:
\begin{align}
    \small
    \label{eq:LSTM_gate}
    \begin{pmatrix} 
        \mathbf{i}_t \\ 
        \mathbf{f}_t \\ 
        \mathbf{o}_t \\ 
        \mathbf{\hat{c}}_t 
    \end{pmatrix}
    &= 
    \begin{pmatrix} 
        \sigma \\ 
        \sigma \\ 
        \sigma \\ 
        \text{tanh}
    \end{pmatrix}
    % \odot
    \begin{pmatrix} 
        \mathbf{W}_i & \mathbf{W}_h
    \end{pmatrix}
    \begin{pmatrix} 
        \mathbf{s}_t \\
        \mathbf{h}_{t-1}
    \end{pmatrix},
\\
    \small
    \mathbf{W}_i &= 
    \begin{pmatrix} 
        \mathbf{W}_i^i \\ 
        \mathbf{W}_i^f \\ 
        \mathbf{W}_i^o \\ 
        \mathbf{W}_i^c 
    \end{pmatrix},
    \mathbf{W}_h = 
    \begin{pmatrix} 
        \mathbf{W}_h^i \\ 
        \mathbf{W}_h^f \\ 
        \mathbf{W}_h^o \\ 
        \mathbf{W}_h^c 
    \end{pmatrix},\\
    \mathbf{o}_t &= \text{Softmax}( W_o\mathbf{o}_t),
\end{align}
where $\mathbf{s}_t \in \mathbb{R}^{n_{inp}}$  and $\mathbf{h}_t \in \mathbb{R}^{n_{dim}}$ at time $t$. Here, $\sigma(\cdot)$ and $\odot$ denote the sigmoid function and element-wise multiplication operator, respectively. The model parameters can be summarized in a compact form with $\theta = [\mathbf{W}_i, \mathbf{W}_h]$, where $\mathbf{W}_{i}\in \mathbb{R}^{4*n_{inp} \times 4*n_{dim} }$, which is the input matrix, and $\mathbf{W}_{h} \in \mathbb{R}^{4*n_{dim} \times 4*n_{dim} }$, which is the hidden matrix. Note that we often refer to $\mathbf{W}_i$ as additive recurrence and $\mathbf{W}_h$ as multiplicative recurrence. Then, $\mathbf{o}_t \in \mathbb{R}^{|V|}$ is generated by the model, and we apply the softmax function to compute the probability distribution.

\subsection{Code-Switched Language Modeling}
It is very well known that the larger the dataset we use for training, the better the language model we can get. However, in the code-switching domain, it is very hard to collect high-quality data and it is very expensive to acquire. One possible source of data is social media, but these data still need to be annotated. There are several approaches to annotate code-switched sentences using LID and syntactic information. We can classify the current research directions into four categories:

\paragraph{Linguistic features}
Linguistic features are commonly used in language model training~\cite{adel2013recurrent,sreeram2018exploiting} and are generally
a combination of both semantic and syntactic features~\cite{adel2015syntactic}, which give more semantic and syntactic information to the model. ~\citet{adel2015syntactic} show that the features can be used as code-switch triggers, such as POS, brown clusters, and open class words.

\paragraph{Data augmentation}
The goal of this approach is to increase the amount of code-switching data by generating new samples to address the data limitation issue. Traditionally in computational linguistics, we depend on linguistic theories to generate more code-switching samples. The theories are used to constraint the generation so that the results are more natural and human-like. The first work to study the use of linguistic theories for this purpose is by~\citet{li2012code}, who used the assumption of the equivalence constraint~\cite{li2012code,pratapa2018language,winata2018learn} and functional head constraint~\cite{li2014language}. Instead of using linguistic theories, other studies have developed methods that use a neural-based approach for data augmentation to learn the code-switching data distribution, such as SeqGAN~\cite{chang2019code} and an encoder-decoder model with a copy mechanism~\cite{winata2019code}.

\paragraph{Transfer learning methods} Leveraging out-of-domain data in the training is a common technique to improve performance and generalization in the code-switching domain. One of the approaches is to apply pre-training on monolingual data before fine-tuning the data with code-switching data using multi-task learning~\cite{li2012code,winata2019code}.
% , and a meta-learning based method, Meta-Transfer Learning.
% The model will be discussed later in Chapter 4~\cite{winata-etal-2020-meta}.

\paragraph{Deep learning architectures}
A number of studies have applied deep learning models to code-switched models. The first neural-based language model was proposed by~\citet{adel2013recurrent} using an RNNLM model. They later~\cite{adel2013combination} proposed an ensemble model to combine factoid RNNLM with n-gram language models. Further work on deep learning for code-switching was done by~\citet{choudhury2017curriculum}, who introduce curriculum learning for training code-switched models to train a network with monolingual training instances,~\citet{garg2018dual}, who propose DualRNN, a model with two RNN components that focus on each of two languages, and~\citet{chandu2018language}, who use language information in an RNN-based language model to help learn code-switching points. 
% In this thesis, we will introduce Syntax-Aware Multi-Task Learning~\cite{winata2018code}, which will be further discussed in Chapter 3.

\subsection{Evaluation Metrics}
The perplexity (PPL) is a standard metric to evaluate a language model. It measures how likely a sequence is to occur~\cite{clarkson1999towards,lee2021towards}, and is formulated as follows:
\begin{equation}
    \text{PPL}(\mathbf{x}) = P(\mathbf{x}_1,\mathbf{x}_2,...,\mathbf{x}_N)^{-\frac{1}{N}} = \sqrt[N]{\frac{1}{P(\mathbf{x}_1,\mathbf{x}_2,...,\mathbf{x}_N)}},
\end{equation}
\noindent where $\text{x} = (\mathbf{x}_1, \mathbf{x}_2, ..., \mathbf{x}_N)$ is a word sequence and $N$ is the sequence length. In the context of code-switching, we calculate PPL differently for Chinese and English. In Chinese, we use characters, while in English we use words. The reason is that some Chinese words are not well tokenized by the Chinese tokenizer, as mentioned by~\citet{garg2018code} and~\citet{winata2019code}, and tokenization results are not consistent. Using characters instead of words in Chinese can alleviate word boundary issues. The PPL is calculated by taking the exponential of the sum of losses. To show the effectiveness of our approach in calculating the probability of the switching, we split the perplexity computation into monolingual segments \textbf{(en-en)} and \textbf{(zh-zh)}, and code-switching segments \textbf{(en-zh)} and \textbf{(zh-en)}.

\section{Sequence Labeling}

Sequence labeling is the key task for NLU. The task is very useful in many NLP applications, for instance, in a conversational virtual agent to detect slot values or named entities from the user inputs. Notably, in code-switching, the sequence labeling task is more challenging since there might be ambiguity in the semantics, and learning the representation for code-switched sentences is not trivial. Sequence labeling comprises a variety of sub-tasks, such as LID, NER, and POS tagging, which will be discussed below.

\subsection{Language Identification}
LID is a task to identify the language of each word within an utterance. Identifying language in code-switched data is crucial since the boundary separates two sub-utterances with different languages. Conventional LID systems operate at the sentence level, which leads to the requirement of word-level LID. Multiple cues, including acoustic, prosodic, and phonetic features are useful features for LID~\cite{lyu2008language}. A couple of shared tasks, EMNLP 2014~\cite{solorio2014overview} and EMNLP 2016~\cite{molina2016overview}, have played an essential role in establishing datasets for LID.  

\subsection{Named Entity Recognition}
NER datasets for code-switching are similar to LID datasets, with word-level annotations. The NER data for code-switching are tweets crawled from online social media for Spanish-English and Arabic-English, and they were compiled and released as a shared task in ACL 2018~\cite{aguilar2018named}. In the shared task,~\citet{attia2018ghht} augment convolutional-based character embeddings and external resources, such as gazetteers and brown clusters into a BiLSTM with a conditional random field (CRF) model, while~\citet{geetha2018tackling} also incorporate external resources like gazetteers and cross-lingual embeddings, such as MUSE~\cite{lample2017unsupervised}.~\citet{winata2018bilingual} introduce a set of pre-processing methods to reduce the out-of-vocabulary (OOV) rate on the code-switching data, and bilingual character embeddings using an RNN model. 
% This model will be further described in Chapter 5.

\subsection{Part-of-Speech Tagging}
POS tagging datasets consist of code-switched sentences tagged at the word-level with POS information. Similar to NER, the datasets are collected from social media on Spanish-English data~\cite{solorio2008part} and Hindi-English data~\cite{singh2018twitter}. This task is an important linguistic component that is used for constituency and dependency parsing. In this task, ~\citet{vyas2014pos} use a CRF-based tagger and a Twitter POS tagger in order to tag sequences of mixed language, while~\citet{solorio2008part} explore monolingual resources to train taggers.
\section{Representation Learning in NLP}

\subsection{Word Embeddings}

Learning a representation through embedding is a fundamental technique to capture latent word semantics \cite{clark2015vector}. In the early stages of research on this topic, Word2Vec embeddings~\cite{mikolov2013efficient, mikolov2013distributed} were the first model to be proposed to the NLP community for word-level representation learning, and interestingly, the embeddings can captures semantics of a word in the form of a vector. The Word2Vec model is used to map words in the discrete space into a vector representation in the continuous space. The goal of building the embeddings is to learn high-quality word vectors from large datasets. There are two training techniques to train word representations: The Skip-gram and continuous bag of words (CBOW) model~\cite{mikolov2013efficient}. The Skip-gram is used to predict the context words for a given word by training, while CBOW applies a reverse technique to Skip-gram: it predicts the word by providing the context as input. 

Another type of word embedding is GloVe~\cite{pennington2014glove}, which leverages statistical information by training only on the nonzero elements in a word-word co-occurrence matrix, instead of on the entire sparse matrix or individual context windows. The model produces a word vector space with a meaningful sub-structure. Standard word vectors ignore the rich structures of languages, and do not consider rare or misspelled words; thus, FastText~\cite{bojanowski2017enriching,mikolov2018advances,grave2018learning} was proposed to address the issue. If a word is unknown, the representations are formed by summing all vectors. To better capture the morphology of the language and address the out-of-vocabulary (OOV) issue, subwords and characters have been commonly used to replace words. One of the commonly used subword embeddings is BPEmb~\cite{heinzerling2018bpemb}, that utilizes byte pair encoding (BPE), a variable-length encoding that views the text as a sequence of symbols and iteratively merges the most frequent symbol pair into a new symbol.

\subsection{Contextualized Language Models}

A deep contextualized model has been proposed to learn representations~\cite{peters2018deep}. Different from word embeddings, it uses a BiLSTM to capture context-dependent aspects of the word semantics to perform well on sequence labeling tasks such as NER and POS tagging. Recently, the BERT model~\cite{devlin2019bert} advanced the state-of-the-art on various NLP tasks. It uses attention models to capture multi-directional representations, instead of using a shallow bidirectional left-to-right and right-to-left, as well as WordPiece~\cite{wu2016google} as the vocabulary to address the OOV issue. A multilingual extension of BERT, mBERT, has shown itself to be very effective on multilingual and cross-lingual NLP tasks~\cite{liu2020attention,liu2020exploring,liu2020cross,liu2020importance}, speech recognition~\cite{Winata2020AdaptandAdjustOT}, and low-resource languages~\cite{wilie2020indonlu}. This multilingual model helps to improve the generalization of different languages since they are trained on large monolingual datasets. However, training a model using only monolingual datasets does not help in code-switching tasks since the model is not informed of the interlingual alignments.

\subsection{Representation Learning for Code-Switching Tasks}

To represent a code-switched sentence, we need a bilingual or multilingual model that understands both languages and has the cross-lingual alignment between words in different languages that have similar semantics. Following are the current methods addressing code-switching representation:

\paragraph{Bilingual Correlation-Based Embeddings (BiCCA)}
\citet{faruqui2014improving} propose canonical correlation analysis, which measures the linear relationship between two embeddings from different languages, finds the two best projection vectors with respect to the correlation, and projects the embeddings into the same dimension space. These word representations are more suitable for encoding a word's semantics than its syntactic information~\cite{faruqui2014improving}.

\paragraph{Bilingual Compositional Model (BiCVM)}
~\citet{hermann-blunsom-2014-multilingual} propose to learn a multilingual representation using parallel sentences and minimize the energy between semantically similar sentences. They use a noise-constrastive large-margin update to ensure that representations of non-aligned sentences have a certain margin between each other.

\paragraph{BiSkip}
~\citet{luong-etal-2015-bilingual} extend the Skip-gram~\cite{mikolov2013distributed} to the bilingual setting, where the model learns to predict word contexts cross-lingually, or in other words, to align words in L1 to corresponding words in L2, and vice versa. 

\paragraph{MUSE}
~\citet{lample2017unsupervised}and~\citet{lample2018word} propose to learn bilingual word embeddings by an unsupervised method that can be used for unsupervised machine translation and cross-lingual tasks. The model learns a mapping from the source to target space using adversarial training and Procustes alignment~\cite{schonemann1966generalized}.

\paragraph{Synthetic Data (GCM) Embeddings}
\citet{pratapa-etal-2018-word} propose to train Skip-gram embeddings from synthetic code-switched data generated by~\citet{pratapa2018language}. They show that GCM improves syntactic and semantic code-switching tasks.

\paragraph{Char2Subword} Recently, ~\citet{aguilar2020char2subword} have proposed the Char2Subword module that builds representations from characters out of the subword vocabulary and they use the module to replace subword embeddings. This model is also robust to typographical errors, such as the misspellings, inflection and casing that are mainly found in social media text. They provide a comprehensive empirical study on a variety of code-switching tasks, such as LID, NER, and POS tagging. 

\section{Complexity Metrics}
Two metrics are commonly used for computing the amount of code-switching in a sequence: the switch-point fraction (SPF) and code mixing index (CMI).

\subsection{Switch-Point Fraction}
The SPF calculates the number of switch-points in a sentence divided by the total number of word boundaries \cite{pratapa2018language}.~We define \textit{switch-point} as a point within a sentence at which the languages of words on either side are different. It is formulated as follows:
\begin{equation}
    SPF(W) = \frac{P(W)}{N(W)},
\end{equation}
\noindent where $N(W)$ is the number of tokens of utterance $W$, and $P(W)$ is the number of code-switching points in utterance $W$.

\subsection{Code Mixing Index} The CMI counts the number of switches in a corpus \cite{gamback2014measuring}. It can be computed at the utterance level by finding the most frequent language in the utterance and then counting the frequency of the words belonging to all other languages present. We compute this metric at the corpus level by averaging all the sentences in a corpus.
The computation is shown as follows:
\begin{equation}
    CMI(W) = \frac{N(W) - max(\ell_i\in \ell\{t_{\ell_i}(W)\}) + P(W)}{N(W)},
\end{equation}
where $CMI(W)$ is the score of the code mixing index of utterance $W$, $N(W)$ is the number of tokens of utterance $W$, $t_{\ell_i}$ is the tokens in language $\ell_i$, and $P(W)$ is the number of code-switching points in utterance $W$. We compute this metric at the corpus level by averaging the values for all sentences.
\chapter{Linguistically-Driven Code-Switched Language Modeling}

% Language model has been a important research on code-switching since it is mainly used in speech recognition model for the ASR decoding process. 

% There has been prior research on language modeling for code-switching, such as using syntactic and language information.
The lack of data problem is the main issue for building a robust code-switching model. 
Although, a vast number of monolingual data are available in social media, they do not follow the same patterns as code-switching speech. Traditionally, this issue is tackled by generating synthetic code-switched data using assumptions from linguistic constraints. However, relying on linguistic constraints may only benefit language pairs that have been heavily investigated in linguistic research. Linguistic constraints are also dependent on syntactic parsers that are unreliable on distant languages, such as English and Mandarin.

In this chapter, we introduce transfer learning approaches that use monolingual data for domain adaptation, and a small number of code-switching data for fine-tuning. We propose language-agnostic computational approaches to generate code-switching data that can be extended to any language pairs using two different methods: (1) leveraging the equivalence constraint to validate the generated data without any external parsers, and (2) learning the data distribution of code-switching data using a neural-based model with a copy mechanism. Then, we take the augmented data to train a language model.
% and use the trained LM model to re-score ASR models.
% In this chapter, we will this chapter into three main sections: (1) we introduce data augmentation method using equivalent constraint, (2) we introduce our neural language model architecture to augment data, and (3) we summarize this chapter.

\section{Model Description}
\subsection{Data Augmentation Using Equivalence Constraint}
\begin{CJK*}{UTF8}{gbsn}
Existing methods of data augmentation apply EC theory to generate code-switching sentences. The methods in~\citet{li2012code} and~\citet{pratapa2018language} may suffer performance issues as they receive erroneous results from the word aligner and POS tagger, causing misclassification that affects the quality of the alignment. Thus, these approaches are not reliable or effective. Recently, ~\citet{garg2018code} proposed a SeqGAN-based model which learns how to generate new synthetic code-switching sentences. However, the distribution of the generated sentences is very different from real code-switching data, which leads to underperformance. Studies on the EC~\cite{poplack1980sometimes,poplack2013sometimes} show that code-switching only occurs where it does not violate the syntactic rules of either language.~\citet{pratapa2018language} apply the EC in English-Spanish LM with a strong assumption. However, we are working with English and Chinese, which have distinctive grammar structures (e.g., POS tags), so applying a constituency parser would give us erroneous results. Thus, we simplify sentences into a linear structure, and we allow lexical substitution on non-crossing alignments between parallel sentences.
\begin{figure}[!ht]
    \centering
    \includegraphics[width=0.7\linewidth]{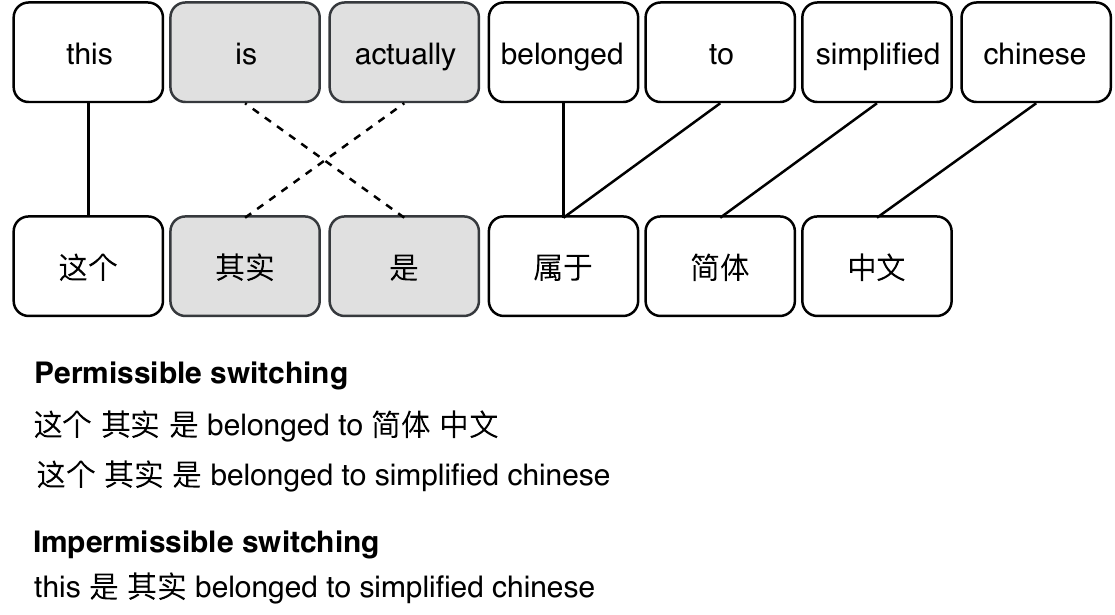}
    \caption{Example of equivalence constraint.}
    \label{fig:eq-contraint}
\end{figure}

Figure~\ref{fig:eq-contraint} shows an example of the EC in English and Chinese. Solid lines show the alignment between the matrix language (top) and the embedded language (bottom).~The dotted lines denote impermissible switching. Alignments between an $L_1$ sentence and an $L_2$ sentence comprise a source vector with indices~$u_t = \{a_1, a_2, ..., a_m\} \in \mathbb{W}^m$ that has a corresponding target vector $v_t = \{b_1, b_2, ..., b_n\} \in \mathbb{W}^n$, where $u$ is a vector of indices sorted in an ascending order. The alignment between $a_i$ and $b_i$ does not satisfy the constraint if there exists a pair of $a_j$ and $b_j$ where ($a_i < a_j$ and $b_i  > b_j$) or ($a_i > a_j$ and $b_i  < b_j$). If the switch occurs at this point, it changes the grammatical order in both languages; thus, this switch is not acceptable. During the generation step, we allow any switches that do not violate the constraint. We propose to generate synthetic code-switching data by the following steps:
\begin{enumerate}
\item Align the $L_1$ sentences $Q$ and $L_2$ sentences $E$ using \texttt{fast\_align}\footnote{The code implementation can be found at https://github.com/clab/fast\_align.}~ \cite{N13-1073}.~We use the mapping from the $L_1$ sentences to the $L_2$ sentences.
\item Permute alignments from step (1) and use them to generate new sequences by replacing the phrase in the $L_1$ sentence with the aligned phrase in the $L_2$ sentence.
\item Evaluate generated sequences from step (2) if they satisfy the EC theory.%, we refer to this as \textit{\textbf{EQ}}.
\end{enumerate}
\end{CJK*}

\subsection{Neural-Based Data Augmentation}

We introduce a neural-based code-switching data generator model using the pointer generator model (Pointer-Gen)~\cite{SeeP17-1099} to learn code-switching constraints from a limited source of code-switching data and leverage their translations in both languages~\cite{winata2019code,winata2018learn}. Intuitively, the copy mechanism can be formulated as an end-to-end solution to copy words from parallel monolingual sentences by aligning and reordering the word positions to form a grammatical code-switching sentence. We remove the dependence on the aligner or tagger, and generate new sentences with a similar distribution to the original dataset.

Initially, Pointer-Gen~\cite{SeeP17-1099} was proposed to learn when to copy words directly from the input to the output in text summarization, and it has since been successfully applied to other NLP tasks, such as comment generation \cite{lin2019learning}. Pointer-Gen leverages the information from the input to ensure high-quality generation, especially when the output sequence consists of elements from the input sequence, such as code-switching sequences.
\begin{figure*}[!t]
  \centering
  \includegraphics[width=\linewidth]{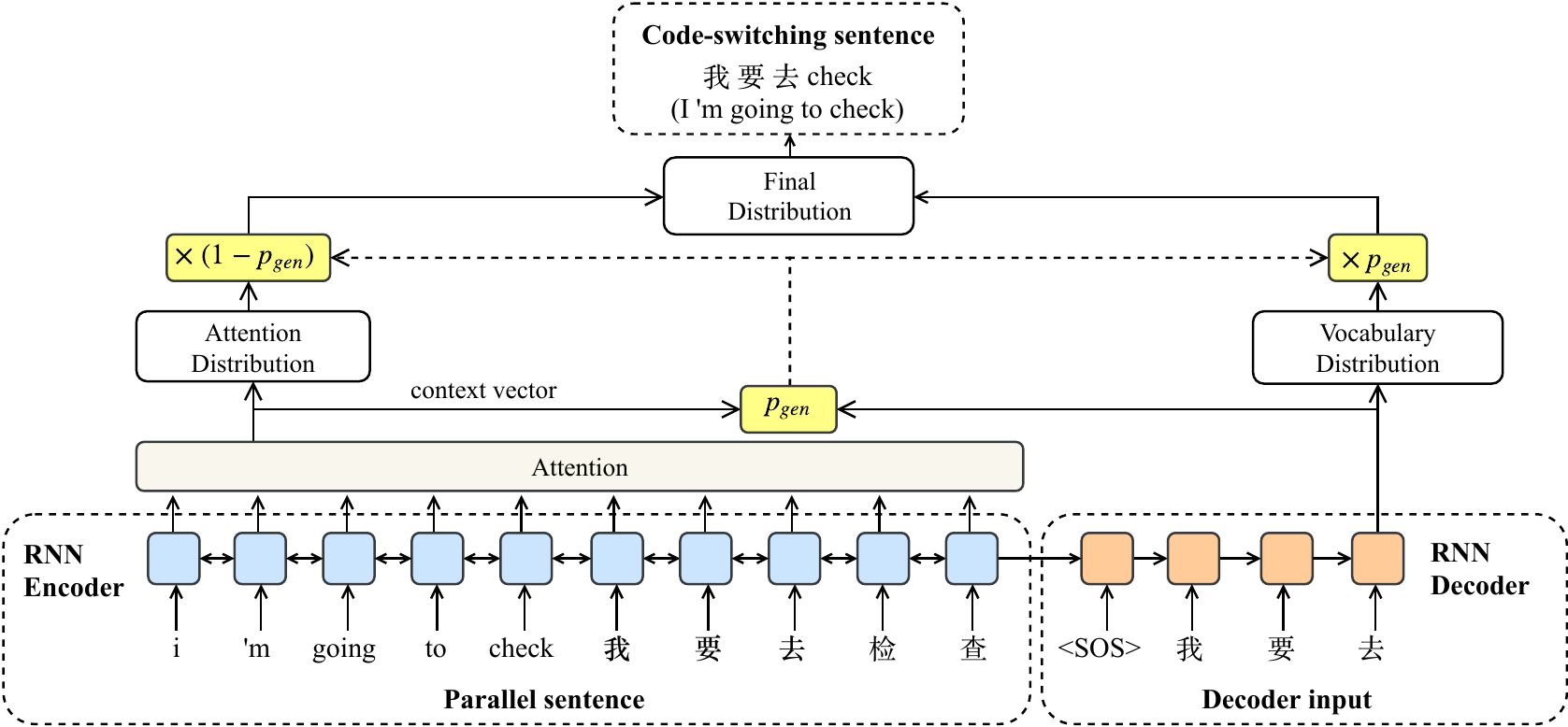}
  \caption{\textbf{Pointer-Gen} model, which includes an RNN encoder and RNN decoder. The parallel sentence is the input of the encoder, and in each decoding step, the decoder generates a new token.}
  \label{fig:pointer_generator}
\end{figure*}
We propose to use Pointer-Gen by leveraging parallel monolingual sentences to generate code-switching sentences. The approach is depicted in Figure \ref{fig:pointer_generator}. Pointer-Gen is trained from concatenated sequences of parallel sentences to generate code-switching sentences, constrained by code-switching text. The words of the input are fed into the encoder. Intuitively, the copy mechanism can be formulated as an end-to-end solution to copy words from parallel monolingual sentences by aligning and reordering the word positions to form a grammatical code-switching sentence. This method removes the dependence on the aligner or tagger, and generates new sentences with a similar distribution to the original dataset. Interestingly, this method can learn the alignment effectively without a word aligner or tagger. As an additional advantage, we demonstrate its interpretability by showing the attention weights learned by the model that represent the code-switching constraints.

We use a BiLSTM, which produces hidden state $h_t$ in each step $t$. The decoder is a unidirectional LSTM receiving the word embedding of the previous word. For each decoding step, a generation probability $p_{gen}$ $\in$ [0,1] is calculated, which weights the probability of generating words from the vocabulary, and copying words from the source text.~$p_{gen}$ is a soft gating probability to decide whether to generate the next token from the decoder or to copy the word from the input instead. The attention distribution $a_t$ is a standard attention with general scoring~\cite{luong2015effective}. It considers all encoder hidden states to derive the context vector. The vocabulary distribution $P_{voc}(w)$ is calculated by concatenating the decoder state $s_t$ and the context vector $h_t^*$:
\begin{equation}
p_{gen} = \sigma (w_{h^*}^T h_t^* + w_s^T s_t + w_x^T x_t + b_{ptr}),
\end{equation}
where $w_{h^*}, w_s, $ and $w_x$ are trainable parameters and $b_{ptr}$ is the scalar bias. The vocabulary distribution $P_{voc}(w)$ and the attention distribution $a^t$ are weighted and summed to obtain the final distribution $P(w)$, which is calculated as follows:
\begin{equation}
P(w) = p_{gen} P_{voc}(w) + (1 - p_{gen})\sum_{i:w_i=w}{a_i^t}.
\end{equation}
We use a beam search to select the $N$-best code-switching sentences.

\subsection{Language Modeling}
We generate data using the EC theory and Pointer-Gen. We also compare our methods with SeqGAN~\cite{garg2018code} as baseline.
% We generate code-switching sentences using three methods: EC theory, SeqGAN, and Pointer-Gen. 
% We take SeqGAN as our baseline. 
To find the best way of leveraging the generated data, we compare training strategies as follows:
\begin{align*}
&\textbf{(1)} \text{ rCS}, \textbf{(2a)} \text{ EC},
\textbf{(2b)} \text{ SeqGAN},\\ &\textbf{(2c)} \text{ Pointer-Gen}, \textbf{ (3a)} \text{ EC \& rCS},\\ 
&\textbf{(3b)} \text{ SeqGAN \& rCS}, \textbf{ (3c)} \text{ Pointer-Gen \& rCS}\\
&\textbf{(4a)} \text{ EC} \rightarrow \text{rCS} \textbf{ (4b)} \text{ SeqGAN} \rightarrow \text{rCS},\\ &\textbf{(4c)} \text{ Pointer-Gen} \rightarrow \text{rCS}
\end{align*}
\textbf{(1)} is training with real code-switching data;~\textbf{(2a--2c)} are training with only augmented data; \textbf{(3a--3c)} are training with the concatenation of augmented data with rCS; and~\textbf{(4a--4c)} are running a two-step training, first training the model only with augmented data and then fine-tuning with rCS. Our early hypothesis is that the results from \textbf{(2a)} and 
\textbf{(2b)} will not be as good as the baseline, but when we combine them, they will outperform the baseline. We expect the result of \textbf{(2c)} to be on par with \textbf{(1)}, since Pointer-Gen learns patterns from the rCS dataset, and generates sequences with similar code-switching points. 

We use a stacked LSTM model to train our language models. The LSTM model is trained using a standard cross-entropy to predict the next token. 

% \subsection{Speech Recognition}
% We train our speech recognition model using 

\section{Experimental Setup}

\subsection{Datasets}
In this section, we use three datasets. First, we use speech data from SEAME (South East Asia Mandarin-English),~\cite{SEAME2015}, a conversational
English-Mandarin Chinese code-switching speech corpus that consists of spontaneously spoken interviews and conversations. This dataset consists of two phases. Phase I is the first data collected in the process, while Phase II data consists of all data from Phase I with additional data collected afterwards. Thus, we choose the Phase II dataset. Table~\ref{data-statistics-phase-2} shows the statistics of the dataset, which is split by speaker ID. We tokenize the tokens in the transcription using the Stanford NLP toolkit~\cite{manning-EtAl:2014:P14-5}. The other two datasets are monolingual speech datasets, HKUST~\cite{liu2006hkust}, comprising spontaneous Mandarin Chinese telephone speech recordings, and Common Voice, an open-accented English dataset collected by Mozilla.\footnote{The dataset is available at https://voice.mozilla.org/.} We split Chinese words into characters to avoid word boundary issues, and generate $L_1$ sentences and $L_2$ sentences by translating the training set of SEAME Phase II into English and Chinese using the Google NMT system.\footnote{https://translate.google.com} Then, we use them to generate 270,531 new pieces of code-switching data, which is thrice the number of the training sets.

\begin{table}[!ht]
\centering
\caption{Data statistics of SEAME Phase II.}
\resizebox{0.45\textwidth}{!}{
\begin{tabular}{@{}rccc@{}}
\toprule
\multicolumn{1}{l}{} & \multicolumn{1}{c}{\textbf{Train}} & \multicolumn{1}{c}{\textbf{Dev}} & \multicolumn{1}{c}{\textbf{Test}} \\ \midrule
\multicolumn{1}{r}{\# Speakers}               & \multicolumn{1}{c}{138} & \multicolumn{1}{c}{8} & \multicolumn{1}{c}{8} \\
\multicolumn{1}{r}{\# Duration (hr)} & \multicolumn{1}{c}{100.58} & \multicolumn{1}{c}{5.56} & \multicolumn{1}{c}{5.25} \\ 
\multicolumn{1}{r}{\# Utterances} & \multicolumn{1}{c}{90,177} & \multicolumn{1}{c}{5,722} & \multicolumn{1}{c}{4,654} \\ 
\multicolumn{1}{r}{\# Tokens} & \multicolumn{1}{c}{1.2M} & \multicolumn{1}{c}{65K} & \multicolumn{1}{c}{60K} \\ 
% \multicolumn{1}{|r|}{\begin{tabular}[c]{@{}r@{}}\# Tokens\\ Preprocessed\end{tabular}} & \multicolumn{1}{c|}{978K} & \multicolumn{1}{c|}{53K} & \multicolumn{1}{c|}{48K} \\ \hline
% \multicolumn{1}{|r|}{Avg. segment} & \multicolumn{1}{c|}{4.21} & \multicolumn{1}{c|}{3.59} & \multicolumn{1}{c|}{3.99}
% \\ \hline
% \multicolumn{1}{|r|}{Avg. switches} & \multicolumn{1}{c|}{1.82}                    & \multicolumn{1}{c|}{1.78}                  & \multicolumn{1}{c|}{1.76}               \\ \hline
\multicolumn{1}{r}{CMI} & \multicolumn{1}{c}{0.18}& \multicolumn{1}{c}{0.22} & \multicolumn{1}{c}{0.19} \\ 
\multicolumn{1}{r}{SPF} & \multicolumn{1}{c}{0.15} & \multicolumn{1}{c}{0.19} & \multicolumn{1}{c}{0.17} \\ \bottomrule
\end{tabular}
}
\label{data-statistics-phase-2}
\end{table}

% \subsection{Training Strategies}
\subsection{Training}
In this section, we present the settings we use to generate code-switching data, and train our language model and end-to-end ASR.

\paragraph{SeqGAN} We implement the SeqGAN model using a PyTorch implementation as our baseline,\footnote{To implement SeqGAN, we use code from https://github.com/suragnair/seqGAN.} and use our best trained language model baseline as the generator in SeqGAN. We sample 270,531 sentences from the generator, thrice the number of the code-switched training data (with a maximum sentence length of 20).

\paragraph{Pointer-Gen} The pointer-generator model has 500-dimensional hidden states. We use 50k words as our vocabulary for the source and target, and optimize the training by Stochastic Gradient Descent (SGD) with an initial learning rate of 1.0 and decay of 0.5. We generate the three best sequences using beam search with five beams, and sample 270,531 sentences, thrice the number of the code-switched training data.

\paragraph{EC} We generate 270,531 sentences, thrice the number of the code-switched training data. To make a fair comparison, we limit the number of switches to two for each sentence to get a similar number of code-switches (SPF and CMI) to Pointer-Gen.

\paragraph{Language Model} In this work, we focus on sentence generation, so we evaluate our data with the same two-layer LSTM language model for comparison. It is trained using a two-layer LSTM with a hidden size of 200 and is unrolled for 35 steps. The embedding size is equal to the LSTM hidden size for weight tying \cite{press2017using}. We optimize our model using SGD with an initial learning rate of 20. If there is no improvement during the evaluation, we reduce the learning rate by a factor of 0.75. In each step, we apply a dropout to both the embedding layer and recurrent network. The gradient is clipped to a maximum of 0.25. We optimize the validation loss and apply an early stopping procedure after five iterations without any improvements. In the fine-tuning step of training strategies \textbf{(4a--4c)}, the initial learning rate is set to 1.

\subsection{Evaluation Metrics}
We use token-level perplexity (PPL) to measure the performance of our models. For the language model, we calculate the PPL of characters in Mandarin Chinese and words in English. The reason is that some Chinese words inside the SEAME corpus are not well tokenized, and tokenization results are not consistent. Using characters instead of words in Chinese can alleviate word boundary issues. The PPL is calculated by taking the exponential of the sum of losses. To show the effectiveness of our approach in calculating the probability of the switching, we split the PPL computation into monolingual segments \textbf{(en-en)} and \textbf{(zh-zh)}, and code-switching segments \textbf{(en-zh)} and \textbf{(zh-en)}.

\begin{table}[!ht]
\caption{Statistics of the generated data. The table shows the number of utterances and words, code-switches ratio, and percentage of new n-grams.}
\centering
\resizebox{0.6\textwidth}{!}{
\begin{tabular}{rccc}
\toprule
\multicolumn{1}{l}{} & \textbf{EC} & \textbf{SeqGAN} & \textbf{Pointer-Gen} \\ \midrule
\# Utterances & 270,531 & 270,531 & 270,531 \\ 
\# Words & 3,040,202 & 2,981,078 & 2,922,941  \\ 
% Avg. switches & 3.57 & 3.51 & 1.82 \\ \hline
new unigram & 13.63\% & 34.67\% & 4.67\%\\ 
new bigram & 69.43\% & 80.33\% & 46.57\% \\
new trigram & 99.73\% & 141.56\% & 69.38\% \\ 
new four-gram & 121.04\% & 182.89\% & 85.07\% \\ 
CMI & 0.25 & 0.13 & 0.25 \\ 
SPF & 0.17 & 0.2 & 0.17 \\ \toprule
\end{tabular}
}
\label{generated-sequences}
\end{table}

\section{Results and Discussion}
\subsection{Code-Switched Data Generation}
\begin{CJK*}{UTF8}{gbsn}
As shown in Table \ref{generated-sequences}, we generate new n-grams including code-switching phrases. This leads us to a more robust model, trained with both generated data and real code-switching data. We can see clearly that Pointer-Gen-generated samples have a distribution more similar to the real code-switching data compared with \textit{SeqGAN}, which shows the advantage of our proposed method. Table \ref{data-statistics-word-trigger} shows the most common English and Mandarin Chinese POS tags that trigger code-switching.~The distribution of word triggers in the Pointer-Gen data are similar to the real code-switching data, indicating our model's ability to learn similar code-switching points. Nouns are the most frequent English word triggers. They are used to construct an optimal interaction by using cognate words and to avoid confusion.~Also, English adverbs such as ``then" and ``so" are phrase or sentence connectors between two language phrases for intra-sentential and inter-sentential code-switching. On the other hand, Chinese transitional words such as the measure word ``个" or associative word ``的" are frequently used as inter-lingual word associations.

\begin{table*}[!ht]
\caption{The most common English and Mandarin Chinese part-of-speech tags that trigger code-switching. We report the frequency ratio from \textbf{Pointer-Gen}-generated sentences compared to the real code-switching data. We also provide an example for each POS tag.}
\centering
\resizebox{0.95\textwidth}{!}{
\begin{tabular}{cc|ccl}
\toprule
\multicolumn{2}{c}{\textbf{rCS}} & \multicolumn{3}{c}{\textbf{Pointer-Gen}} \\ \midrule
\textbf{POS tags} & \textbf{ratio} & \textbf{POS tags} & \multicolumn{1}{c}{\textbf{ratio}} & \multicolumn{1}{l}{\textbf{example}} \\ \midrule
\multicolumn{5}{c}{\textbf{English}} \\ \midrule
\begin{tabular}[c]{@{}c@{}}NN \\ (noun)\end{tabular} & 56.16\% & \begin{tabular}[c]{@{}c@{}}NN \\ (noun)\end{tabular} & \multicolumn{1}{r}{55.45\%} & \begin{tabular}[c]{@{}c@{}}\text{那} \text{个} \textbf{{consumer}} \text{是} \text{不} \\ (that consumer is not)\end{tabular} \\ \midrule
\begin{tabular}[c]{@{}c@{}}RB \\ (adverb)\end{tabular} & 10.34\% & \begin{tabular}[c]{@{}c@{}}RB \\ (adverb)\end{tabular} & \multicolumn{1}{r}{10.14\%} & \begin{tabular}[c]{@{}l@{}}okay \textbf{{so}} \text{其} \text{实} \\ (okay so its real)\end{tabular} \\ \midrule
\begin{tabular}[c]{@{}c@{}}JJ \\ (adjective)\end{tabular} & 7.04\% & \begin{tabular}[c]{@{}c@{}}JJ \\ (adjective)\end{tabular} & \multicolumn{1}{r}{7.16\%} & \begin{tabular}[c]{@{}l@{}}\text{我} \text{很} \textbf{{jealous}} \text{的} \text{每} \text{次} \\ (i am very jealous every time)\end{tabular} \\ \midrule
\begin{tabular}[c]{@{}c@{}}VB \\ (verb)\end{tabular} & 5.88\% & \begin{tabular}[c]{@{}c@{}}VB \\ (verb)\end{tabular} & \multicolumn{1}{r}{5.89\%} & \begin{tabular}[c]{@{}l@{}}\textbf{{compared}} \text{这} \text{个} \\ (compared to this)\end{tabular} \\ \midrule
\multicolumn{5}{c}{\textbf{Chinese}} \\ \midrule
\begin{tabular}[c]{@{}c@{}}VV \\ (other verbs)\end{tabular} & 23.77\% & \begin{tabular}[c]{@{}c@{}}VV \\ (other verbs)\end{tabular} & 23.72\% & \begin{tabular}[c]{@{}l@{}}\text{讲} \text{的} \text{要} \textbf{{用}} microsoft word \\ (i want to use microsoft word)\end{tabular} \\ \midrule
\begin{tabular}[c]{@{}c@{}}M \\ (measure word)\end{tabular} & 16.83\% & \begin{tabular}[c]{@{}c@{}}M \\ (measure word)\end{tabular} & 16.49\% & \begin{tabular}[c]{@{}l@{}}\text{我} \text{们} \text{有} \text{这} \textbf{{个}} god of war \\ (we have this god of war)\end{tabular} \\ \midrule
\begin{tabular}[c]{@{}c@{}}DEG \\ (associative)\end{tabular} & 9.12\% & \begin{tabular}[c]{@{}c@{}}DEG \\ (associative)\end{tabular} & 9.13\% & \begin{tabular}[c]{@{}l@{}}\text{我} \text{们} \textbf{{的}} result \\ (our result)\end{tabular} \\ \midrule
\begin{tabular}[c]{@{}c@{}}NN \\ (common noun)\end{tabular} & 9.08\% & \begin{tabular}[c]{@{}c@{}}NN \\ (common noun)\end{tabular} & 8.93\% & \begin{tabular}[c]{@{}l@{}}\text{我} \text{应} \text{该} \text{不} \text{会} \text{讲} \textbf{{话}} because intimidated by another\\ (i shouldn’t talk because intimidated by another)\end{tabular} \\ \bottomrule
\end{tabular}
}
\label{data-statistics-word-trigger}
\end{table*}
\end{CJK*}

\subsection{Language Modeling}
In Table \ref{lm-results}, we can see the perplexities of the test set evaluated on different training strategies. Pointer-Gen consistently performs better than state-of-the-art models such as \textit{EC} and \textit{SeqGAN}. Comparing the results of models trained using only generated samples, \textbf{(2a-2b)} leads to the undesirable results that are also mentioned by Pratapa et al.~\cite{pratapa2018language}, but this does not apply to Pointer-Gen \textbf{(2c)}. We can achieve similar results with the model trained using only real code-switching data, rCS. This demonstrates the quality of our data generated using Pointer-Gen. In general, combining any generated samples with real code-switching data improves the language model performance for both code-switching segments and monolingual segments. Applying concatenation is less effective than the two-step training strategy. Moreover, applying the two-step training strategy achieves state-of-the-art performance. 

\begin{table*}[!ht]
\centering
\caption{Results of perplexity (PPL) on a validation set and test set for different training strategies. We report the overall PPL, and code-switching points \textbf{(en-zh)} and \textbf{(zh-en)}, as well as the monolingual segments PPL  \textbf{(en-en)} and  \textbf{(zh-zh)}.}
\resizebox{\textwidth}{!}{
\begin{tabular}{lcccccccccc}
\toprule
\multirow{2}{*}{\textbf{Training Strategy}} & \multicolumn{2}{c}{\textbf{Overall}} & \multicolumn{2}{c}{\textbf{en-zh}} & \multicolumn{2}{c}{\textbf{zh-en}} & \multicolumn{2}{c}{\textbf{en-en}} & \multicolumn{2}{c}{\textbf{zh-zh}}  \\ \cmidrule{2-11} 
  & \textbf{valid} & \textbf{test} & \textbf{valid} & \textbf{test} & \textbf{valid} & \textbf{test} & \textbf{valid} & \textbf{test} & \textbf{valid} & \textbf{test} \\ \midrule
\multicolumn{11}{l}{\textit{Only real code-switching data}} \\ \midrule
\text{(1)} rCS & 72.89 & 65.71 & 7411.42 & 7857.75 & 120.41 & 130.21 & 29.31 & 29.61 & 244.88 & 246.71 \\ \midrule
\multicolumn{11}{l}{\textit{Only generated data}} \\ \midrule
\text{(2a)} EC & 115.98 & 96.54 & 32865.62 & 30580.89 & 107.22 & 109.10 & 28.24 & 28.2 & 1893.77 & 1971.68 \\ 
% \text{(2b)} EC & 96.21 & 82.69 & 11910.54 & 11236.51 & 154.03 & 156.61 & 33.13 & 32.57 & 447.41 & 438.84 \\
\text{(2b)} SeqGAN & 252.86 & 215.17 & 33719 & 37119.9 & 174.2 & 187.5 & 91.07 & 88 & 1799.74 & 1783.71 \\
% \text{(2c)} SeqGAN & 263.95 & 219.65 & 49534.84 & 56196.51 & 160.46 & 172.39 & 90.45 & 85.66 & 1787.57 & 1730.79 \\
\text{(2c)} Pointer-Gen & 72.78 & 64.67 & 7055.59 & 7473.68 & 119.56 & 133.39 & 27.77 & 27.67 & 234.16 & 235.34 \\ \midrule
\multicolumn{11}{l}{\textit{Concatenate generated data with real code-switching data}} \\ \midrule
\text{(3a)} EC \& rCS & 70.33 & 62.43 & 8955.79 & 9093.01 & 130.92 & 139.06 & 26.49 & 26.28 & 227.57 & 242.30 \\
\text{(3b)} SeqGAN \& rCS & 77.37 & 69.58 & 8477.44 & 9350.73 & 134.27 & 143.41 & 30.64 & 30.81 & 260.89 & 264.28 \\
% \text{(3b)} EC \& rCS & 70.96 & 63.20 & 8719.49 & 8697.98 & 138.69 & 148.01 & 27.21 & 27.20 & 223.85 & 231.64 \\
% \text{(3b)} SeqGAN \& rCS & 75.21 & 67.62 & 8170.75 & 8677.63 & 127.5 & 139.13 & 29.86 & 30.05 & 252.92 & 256.82 \\
\text{(3c)} Pointer-Gen \& rCS & 68.49 & 61.57 & 7146.08 & 7667.82 & 127.50 & 139.06 & 26.75 & 26.96 & 218.27 & 226.60\\ \midrule
% \text{(4b)} EC $\rightarrow$ rCS & 68.64 & 61.44 & 8008.73 & 8075.36 & \textbf{112.15} & \textbf{116.68} & 26.55 & 26.80 & 228.02 & 233.74 \\
% \textbf{(4c)} \textbf{Pointer-Gen $\rightarrow$ CS} & \textbf{66.75} & \textbf{60.16} & \textbf{6670.17} & \textbf{7168.45} & 113.47 & 125.75 & 26.65 & \textbf{26.66} & \textbf{217.67} & \textbf{222.63} \\ \hline
% \text{(4c)} SeqGAN $\rightarrow$ rCS & 70.61 & 64.03 & 6950.02 & 7694.2 & 114.82 & 122.84 & 28.5 & 28.73 & 236.94 & 244.62 \\
\multicolumn{11}{l}{\textit{Pretrain with generated data and fine-tune with real code-switching data}} \\ \midrule
\text{(4a)} EC $\rightarrow$ rCS & 68.46 & 61.42 & 8200.78 & 8517.29 & 101.15 & 107.77 & 25.49 & 25.78 & 247.30 & 258.95 \\
\text{(4b)} SeqGAN $\rightarrow$ rCS & 70.61 & 64.03 & 6950.02 & 7694.2 & 114.82 & 122.84 & 28.50 & 28.73 & 236.94 & 244.62 \\
\textbf{(4c)} \textbf{Pointer-Gen $\rightarrow$ rCS} & \textbf{66.08} & \textbf{59.74} & \textbf{6620.76} & \textbf{7172.42} & \textbf{114.53} & \textbf{127.12} & \textbf{26.36} & \textbf{26.40} & \textbf{216.02} & \textbf{222.49} \\ \bottomrule
\end{tabular}
}
\label{lm-results}
\end{table*}

% \subsection{Speech Recognition}
% We evaluate our proposed sentence generation method on an end-to-end ASR system. Table \ref{asr-evaluation} shows the CER of our ASR systems, as well as the individual CER on each language. Based on the experimental results, pretraining is able to reduce the error rate by 1.64\%, as it corrects the spelling mistakes in the prediction.~After we add LM (rCS) to the decoding step, the error rate can be reduced to 32.25\%. Finally, we replace the LM with LM (Pointer-Gen $\rightarrow$ rCS), and it further decreases the error rate by 1.18\%.

% \begin{table}[!h]
% \caption{ASR evaluation, showing the performance on all sequences: Overall, English segments (en), and Mandarin Chinese segments (zh).}
% \centering
% \resizebox{0.65\textwidth}{!}{
% \begin{tabular}{lccc}
% \toprule
% \textbf{Model} & \textbf{Overall} & \textbf{en} & \textbf{zh}\\ \midrule
% % Baseline & 34.40\% & 41.70\% & 35.84\% \\
% Baseline & 34.40\% & 41.79\% & 35.94\% \\
% + Pre-training & 32.76\% & 40.06\% & 32.44\% \\
% \hspace{3mm}\text{+ LM (rCS)} & 32.25\% & 39.45\% & 31.90\% \\ \midrule
% % \hspace{3mm}\textbf{+ LM (4c)} & \text{31.75\%} & \text{39.03\%} & \text{31.37\%} \\ \hline
% % \hspace{3mm}\textbf{+ LM (4c)} & \text{31.55\%} & \text{39.24\%} & \text{31.05\%} \\ \hline
% % \hspace{3mm}\textbf{+ LM (4c)} & \text{31.31\%} & \text{38.66\%} & \text{30.93\%} \\ \hline
% \hspace{3mm}\textbf{+ LM (Pointer-Gen $\rightarrow$ rCS)} & \textbf{31.07\%} & \textbf{38.39\%} & \textbf{30.85\%} \\ \bottomrule
% \end{tabular}
% }
% \label{asr-evaluation}
% \end{table}
\subsection{Analysis}
\begin{figure*}[!ht]
  \centering
  \includegraphics[width=\linewidth]{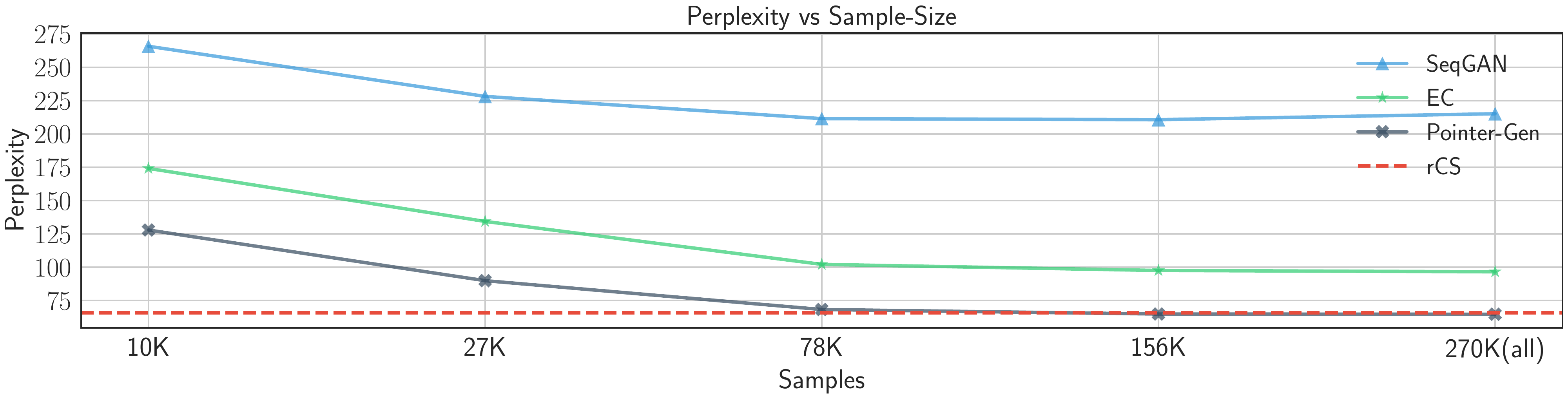}
  \caption{Results of perplexity (PPL) on different numbers of generated samples. The graph shows that Pointer-Gen attains a close performance to the real training data, and outperforms \textit{SeqGAN} and \textit{EC}.}
  \label{fig:data_effects}
\end{figure*}

\paragraph{Effect of Data Size}
To understand the importance of data size, we train our model with varying amounts of generated data. Figure \ref{fig:data_effects} shows the PPL of the models with different amounts of generated data. An interesting finding is that our model trained with only 78K samples of Pointer-Gen data (same number of samples as rCS) achieves a similar PPL to the model trained with only rCS, while SeqGAN and EC have a significantly higher PPL. We can also see that 10K samples of Pointer-Gen data are as good as 270K samples of EC data. In general, the number of samples is positively correlated with an improvement in performance.

\begin{CJK*}{UTF8}{gbsn}
\begin{figure*}[!ht]
  \centering
  \includegraphics[width=\linewidth]{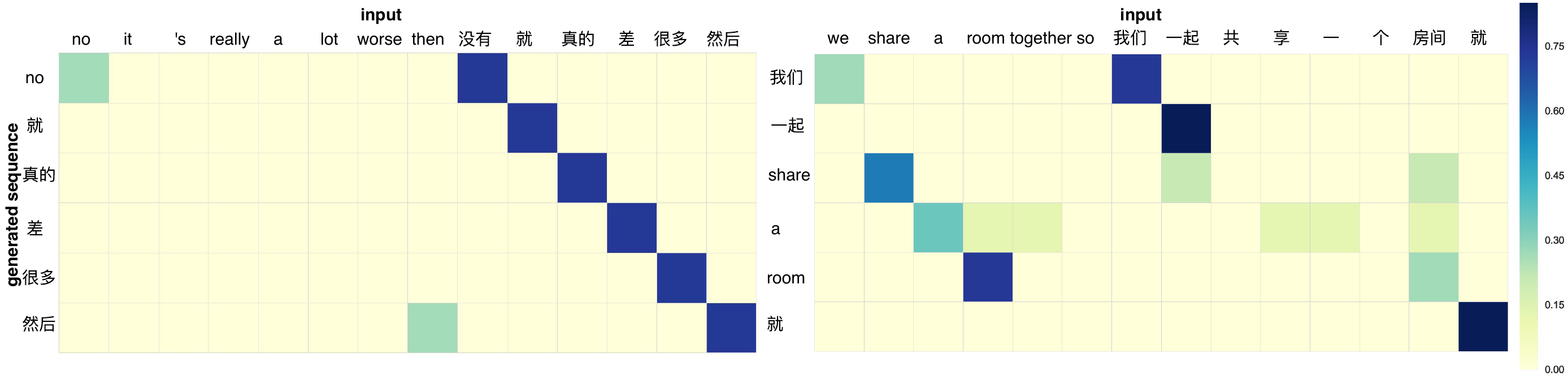}
  \caption{The visualization of pointer-generator attention weights on input words in each time-step during the inference time. The y-axis indicates the generated sequence, and the x-axis indicates the word input. In this figure, we show the code-switching points.}
  \label{fig:attention}
\end{figure*}
\end{CJK*}

\paragraph{Model Interpretability}
\begin{CJK*}{UTF8}{gbsn}
We can interpret a Pointer-Gen model by extracting its attention matrices and then analyzing the activation scores. We show the visualization of the attention weights in Figure \ref{fig:attention}.~The square in the heatmap corresponds to the attention score of an input word. In each time-step, the attention scores are used to select words to be generated. As we can observe in the figure, in some cases, our model attends to words that are translations of each other, for example, the words (``no",``没有"), (``then",``然后") , (``we",``我们"), (``share", ``一起"), and (``room",``房间"). This indicates the model can identify code-switching points, word alignments, and translations without being given any explicit information.
\end{CJK*}

\section{Short Summary}
In this chapter, we propose two methods for generating synthetic code-switched sentences using EC and Pointer-Gen. The former method alleviates the dependency on syntactic parsers that can be applied to other language pairs, and the latter method learns how to copy words from parallel corpora. Interestingly, Pointer-Gen also captures code-switching points by attending to input words and aligning the parallel words, without requiring any word alignments or constituency parsers. More importantly, it can be effectively used for languages that are syntactically different, such as English and Mandarin. Our language model trained using Pointer-Gen outperforms the EC theory-based models and SeqGAN model.
% ~We also show that the learned language model can be used to improve the performance of an end-to-end ASR system.
\chapter{Syntax-Aware Multi-Task Learning for Code-Switched Language Modeling}

LM using only word lexicons is not adequate to learn the complexity of code-switching patterns, especially in a low-resource setting. In the linguistic world, code-switching patterns can be found by observing the syntactic features that are useful information to identify code-switch points, as they are not produced randomly.~\citet{poplack1978syntactic} shows that there are higher proportions of certain types of switches in the presence of syntactic features. Therefore, we conjecture that the syntactic features are highly correlated to the code-switching triggers. 

In this chapter, we propose a multi-task learning framework for the code-switching LM task, which is able to leverage syntactic features such as language information and POS tags~\cite{winata2018code}. Using syntactic features allows the model to learn shared grammatical information that constrains the next word prediction. The main contribution of this work is two-fold. First, a multi-task learning model is proposed to jointly learn the LM task and POS sequence tagging task on code-switched utterances. Second, we incorporate language information into POS tags to create bilingual tags --- The tags distinguish between Chinese and English. The POS tag features are shared with the language model and enrich the features to learn better where to switch. 

\section{Model Description}
\begin{figure}[t]
  \centering
  \includegraphics[width=0.62\linewidth]{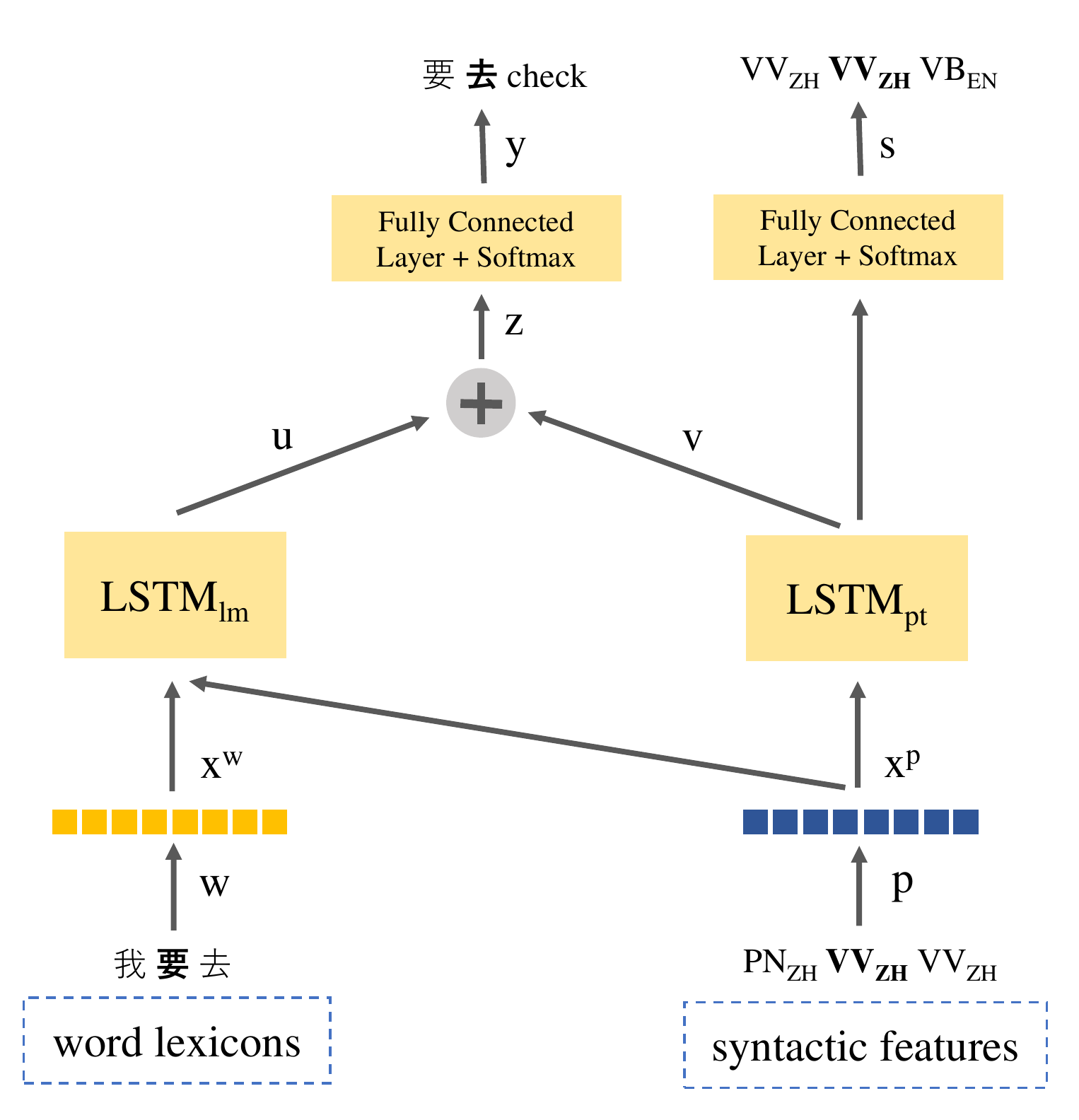}
  \caption{Multi-task learning framework.}
  \label{fig:multi-task-model}
\end{figure}

\begin{CJK*}{UTF8}{gbsn}
The proposed multi-task learning consists of two NLP tasks: LM and POS sequence tagging. Figure \ref{fig:multi-task-model} illustrates our multi-task learning extension to the recurrent language model. We use LSTM~\cite{hochreiter1997long} in our model instead of the standard RNN. The LM task is a standard next word prediction, and the POS tagging task is to predict the next POS tag. The POS tagging task shares the POS tag vector and the hidden states to the LM task, but it does not receive any information from the other loss. Let $w_t$ be the word lexicon in the document and $p_t$ be the POS tag of the corresponding $w_t$ at index $t$. They are mapped into embedding matrices to get their $d$-dimensional vector representations $x_t^w$ and $x_t^p$. The input embedding weights are tied with the output weights. We concatenate $x_t^w$ and $x_t^p$ as the input of the $\textnormal{LSTM}_{lm}$. The information from the POS tag sequence is shared with the language model through this step:
\[ u_t = \textnormal{LSTM}_{lm}(x_t^{w} \oplus x_t^{p}, u_{t-1}), \]
\[ v_t = \textnormal{LSTM}_{pt}(x_t^{p}, v_{t-1}), \]

\noindent where $\oplus$ denotes the concatenation operator, and $u_t$ and $v_t$ are the final hidden states of $\textnormal{LSTM}_{lm}$ and $\textnormal{LSTM}_{pt}$ respectively. $u_t$ and $v_t$, the hidden states from both LSTMs, are then summed before predicting the next word. Then, we project the vector to the vocabulary space by multiplying it with learned parameters $W_{lm}^O$ and bias $b_{lm}$:
\begin{align}
z_t &= u_t + v_t,\\
y_t & = W_{lm}^O (z_t) + b_{lm},\\
%\[ y_t = \textnormal{Softmax}(z_t) \]
y_t &= \frac{e^{y_t}}{\sum_{j=1}^T e^{y_j}} \textnormal{, where } j = 1 \textnormal{, .., }T.
\end{align}
\noindent For the POS tagging task, the model also learns a learned weight $W_{pt}^O$ to project vector $v$ to the POS label distribution $s$ as follows:
\begin{align}
    s_t &= W_{pt}^O (v_t) + b_{pt},\\
    s_t &= \frac{e^{s_t}}{\sum_{j=1}^T e^{s_j}} \textnormal{, where } j = 1 \textnormal{, .., }T.
\end{align}
%The language model is implemented using LSTM unit.
% \begin{align*} 
% \textbf{i}_t &= \sigma(\textbf{W}_t \textbf{v}_t + \textbf{U}_i \textbf{h}_{t-1} + \textbf{b}_i) \\
% \textbf{f}_t &= \sigma(\textbf{W}_f \textbf{v}_t + \textbf{U}_f \textbf{h}_{t-1} + \textbf{b}_f) \\
% \textbf{o}_t &= \sigma(\textbf{W}_o \textbf{v}_t + \textbf{U}_o \textbf{h}_{t-1} + \textbf{b}_i) \\
% \hat{\textbf{c}}_t &= \textnormal{tanh}(\textbf{W}_f \textbf{v}_t + \textbf{U}_c \textbf{h}_{t-1} + \textbf{b}_c) \\
% \textbf{c}_t &= f_t \odot \textbf{c}_{t-1} + \textbf{i}_t \odot \hat{\textbf{c}}_t \\
% \textbf{h}_t &= \textbf{o}_t \odot \textnormal{tanh}(\textbf{c}_t)
% \end{align*}

% \begin{align*}
% i_t &= \sigma(W_{ii} x_t + b_{ii} + W_{hi} h_{(t-1)} + b_{hi}) \\
% f_t &= \sigma(W_{if} x_t + b_{if} + W_{hf} h_{(t-1)} + b_{hf}) \\
% g_t &= \tanh(W_{ig} x_t + b_{ig} + W_{hg} h_{(t-1)} + b_{hg}) \\
% o_t &= \sigma(W_{io} x_t + b_{io} + W_{ho} h_{(t-1)} + b_{ho}) \\
% c_t &= f_t \odot c_{(t-1)} + i_t \odot g_t \\
% h_t &= o_t \odot \tanh(c_t)
% \end{align*}

%where $\odot$ denotes element-wise product, $i_t$, $f_t$, $o_t$ are input, forget and output gate activations at time t. 

The word distribution of the next word $y_{t}$ is normalized using the softmax function. The model uses cross-entropy losses as error functions $\mathcal{L}_{lm}$ and $\mathcal{L}_{pt}$ for the LM task and POS tagging task, respectively. We optimize the multi-objective losses using the Back Propagation algorithm, and we perform a weighted linear sum of the losses for each individual task:
\begin{align}
\mathcal{L}_{total} = p \mathcal{L}_{lm} + (1-p) \mathcal{L}_{pt},
\end{align}
\noindent where $p$ is the weight of the loss in training.%, and $\mathcal{L}_{lm}$ and $\mathcal{L}_{pt}$ are cross-entropy losses.

\section{Experimental Setup}

% In this section, we present the experimental settings for this task as follows:

\subsection{Dataset}
SEAME (South East Asia Mandarin-English), a conversational
Mandarin-English code-switching speech corpus, consists of spontaneously spoken interviews and conversations \cite{SEAME2015}. Our dataset (LDC2015S04) is the most up-to-date version of the Linguistic Data Consortium (LDC) database. However, the statistics are not identical to those from~\citet{lyu2010analysis}. The corpus consists of two phases. In Phase I, only selected audio segments were transcribed, while in Phase II, most of the audio segments were transcribed. For each \textit{language island}, a phrase within the same language, we extract POS tags iteratively using the Chinese and English Penn Tree Bank Parser~\cite{tseng2005morphological, toutanova2003feature}. There are 31 English POS tags and 34 Chinese POS tags. Chinese words are distinguishable from English words since they have a different encoding. We add language information to the POS tag label to discriminate POS tags between the two languages. For our pre-processing steps, we tokenize English and Chinese words using the Stanford NLP toolkit~\cite{manning-EtAl:2014:P14-5}. Second, all hesitations and punctuation are removed except apostrophes, for example, ``let's" and ``it's." Table \ref{data-statistics-phase-1} and Table \ref{data-statistics-phase-2} show the statistics of the SEAME Phase I and II corpora. Table \ref{trigger-words} shows the most common trigger POS tags for the Phase II corpus.
% According to the authors, it was not possible to restore the original dataset. The authors only used Phase I corpus.

\begin{table}[!htb]
\centering
\caption{Data statistics in SEAME Phase I.}
\label{data-statistics-phase-1}
\begin{tabular}{@{}rccc@{}}
\hline \toprule
\multicolumn{1}{r}{} & \multicolumn{1}{c}{\textbf{Train set}} & \multicolumn{1}{c}{\textbf{Dev set}} & \textbf{Test set} \\ \midrule
\multicolumn{1}{r}{\# Speakers} & \multicolumn{1}{c}{139} & \multicolumn{1}{c}{8} & 8 \\ 
\multicolumn{1}{r}{\# Utterances} & \multicolumn{1}{c}{45,916} & \multicolumn{1}{c}{1,938} & 1,228                         \\ 
\multicolumn{1}{r}{\# Tokens} & \multicolumn{1}{c}{762K} & \multicolumn{1}{c}{31K} & 17K                        \\ 
\multicolumn{1}{r}{\begin{tabular}[c]{@{}r@{}}Avg. segments length\end{tabular}} & \multicolumn{1}{c}{3.67} & \multicolumn{1}{c}{3.68}         & 3.18          \\ 
\multicolumn{1}{r}{Avg. switches}                    & \multicolumn{1}{c}{3.60}           & \multicolumn{1}{c}{3.47}         & 3.67          \\ \bottomrule
\end{tabular}
\end{table}

% \begin{table}[!htb]
% \centering
% \caption{Data statistics in SEAME Phase II.}
% \label{data-statistics-phase-2}
% \begin{tabular}{@{}rccc@{}}
% \toprule
% \multicolumn{1}{l}{}                                                                & \multicolumn{1}{c}{\textbf{Train set}} & \multicolumn{1}{c}{\textbf{Dev set}} & \multicolumn{1}{c}{\textbf{Test set}} \\ \midrule
% \multicolumn{1}{r}{\# Speakers}                                                   & \multicolumn{1}{c}{138}                     & \multicolumn{1}{c}{8}                     & \multicolumn{1}{c}{8}                      \\ 
% \multicolumn{1}{r}{\# Utterances}                                                 & \multicolumn{1}{c}{78,815}                  & \multicolumn{1}{c}{4,764}                 & \multicolumn{1}{c}{3,933}                  \\ 
% \multicolumn{1}{r}{\# Tokens}                                                     & \multicolumn{1}{c}{1.2M}                    & \multicolumn{1}{c}{65K}                   & \multicolumn{1}{c}{60K}                    \\ 
% \multicolumn{1}{r}{\begin{tabular}[c]{@{}r@{}}Avg. segment length\end{tabular}} & \multicolumn{1}{c}{4.21}                    & \multicolumn{1}{c}{3.59}                  & \multicolumn{1}{c}{3.99}                   \\ 
% \multicolumn{1}{r}{Avg. switches} & \multicolumn{1}{c}{2.94}                    & \multicolumn{1}{c}{3.12}                  & \multicolumn{1}{c}{3.07}                   \\ \bottomrule
% \end{tabular}
% \end{table}

\begin{table}[!ht]
\centering
\caption{Code-switching trigger words in SEAME Phase II.}
\label{trigger-words}
\begin{tabular}{@{}lclc@{}}
\toprule
\textbf{POS Tag} & \multicolumn{1}{c}{\textbf{Freq}} & \multicolumn{1}{c}{\textbf{POS Tag}} & \multicolumn{1}{c}{\textbf{Freq}}\\ \midrule
$\textnormal{VV}_{ZH}$ & \multicolumn{1}{c}{107,133} & $\textnormal{NN}_{EN}$ & \multicolumn{1}{c}{31,031} \\ 
$\textnormal{AD}_{ZH}$ & \multicolumn{1}{c}{97,681} & $\textnormal{RB}_{EN}$ & \multicolumn{1}{c}{12,498} \\ 
$\textnormal{PN}_{ZH}$ & \multicolumn{1}{c}{92,117} & $\textnormal{NNP}_{EN}$ & \multicolumn{1}{c}{11,734} \\ 
$\textnormal{NN}_{ZH}$ & \multicolumn{1}{c}{45,088} & $\textnormal{JJ}_{EN}$ & \multicolumn{1}{c}{5,040} \\ 
$\textnormal{VA}_{ZH}$ & \multicolumn{1}{c}{27,442} & $\textnormal{IN}_{EN}$ & \multicolumn{1}{c}{4,801} \\ 
$\textnormal{CD}_{ZH}$ & \multicolumn{1}{c}{20,158} & $\textnormal{VB}_{EN}$ & \multicolumn{1}{c}{4,703} \\ \bottomrule
\end{tabular}
\end{table}

%We analyze POS TAG trigger on our training set.

% \begin{table}[!htb]
% \centering
% \caption{My caption}
% \label{my-label}
% \begin{tabular}{|l|l|l|}
% POS Tag & occurrences & CS-rate \\
%         &            &         \\
%         &            &        
% \end{tabular}
% \end{table}

\subsection{Training} 
We train our LSTM models with different hidden sizes [200, 500]. All LSTMs have two layers and are unrolled for 35 steps. The embedding size is equal to the LSTM hidden size. A dropout regularization~\cite{srivastava2014dropout} is applied to the word embedding vector and POS tag embedding vector, and to the recurrent output~\cite{gal2016theoretically} with values between [0.2, 0.4]. We use a batch size of 20 in training. An end of sentence (EOS) tag is used to separate every sentence. We use SGD and start with a learning rate of 20, and if there is no improvement during the evaluation, we reduce the learning rate by a factor of 0.75. The gradient is clipped to a maximum of 0.25. For multi-task learning, we use different weight loss hyper-parameters $p$ in the range of [0.25, 0.5, 0.75]. We tune our model with the development set and evaluate our best model using the test set, taking PPL as the final evaluation metric, which is calculated by taking the exponential of the error in the negative log-form:
\begin{align}
\textnormal{PPL}(w) = P(\mathbf{w}_1,\mathbf{w}_2,...,\mathbf{x}_N)^{-\frac{1}{N}} = \sqrt[N]{\frac{1}{P(\mathbf{w}_1,\mathbf{w}_2,...,\mathbf{w}_N)}}.
% e^{\mathcal{L}_{total}}.
\end{align}

\subsection{Evaluation}
We evaluate our proposed method on both SEAME Phase I and Phase II. We compare our method with existing baselines only on SEAME Phase I, since all of those models were only evaluated on the Phase I dataset. The baselines are as follows:
\paragraph{RNNLM} The baseline model trained using RNNLM~\cite{mikolov2011rnnlm}.\footnote{ http://www.fit.vutbr.cz/~imikolov/rnnlm/}
\paragraph{wFLM} The factored language model proposed by~\citet{adel2015syntactic} using FLM open class clusters, brown clusters, part-of-speech and the trigram language model.
\paragraph{FI + OF} The model from~\citet{adel2013recurrent} that uses a factorized RNNLM with POS as input and is trained with a multi-task objective to predict the next word and language.
\paragraph{RNNLM + FLM} The model from~\citet{adel2013combination} that combines FLM with RNNLM by language model interpolation.

% We can also compute the joint probability of a sequence $P(w)$ by the following:
% \[ P(w) = \prod_{t=1}^T P(w_t|w_{t-1}, w_{t-2}, ..., w_1) \]

\section{Results and Discussion}
\begin{figure*}[!ht]
  \centering
  \includegraphics[width=\linewidth]{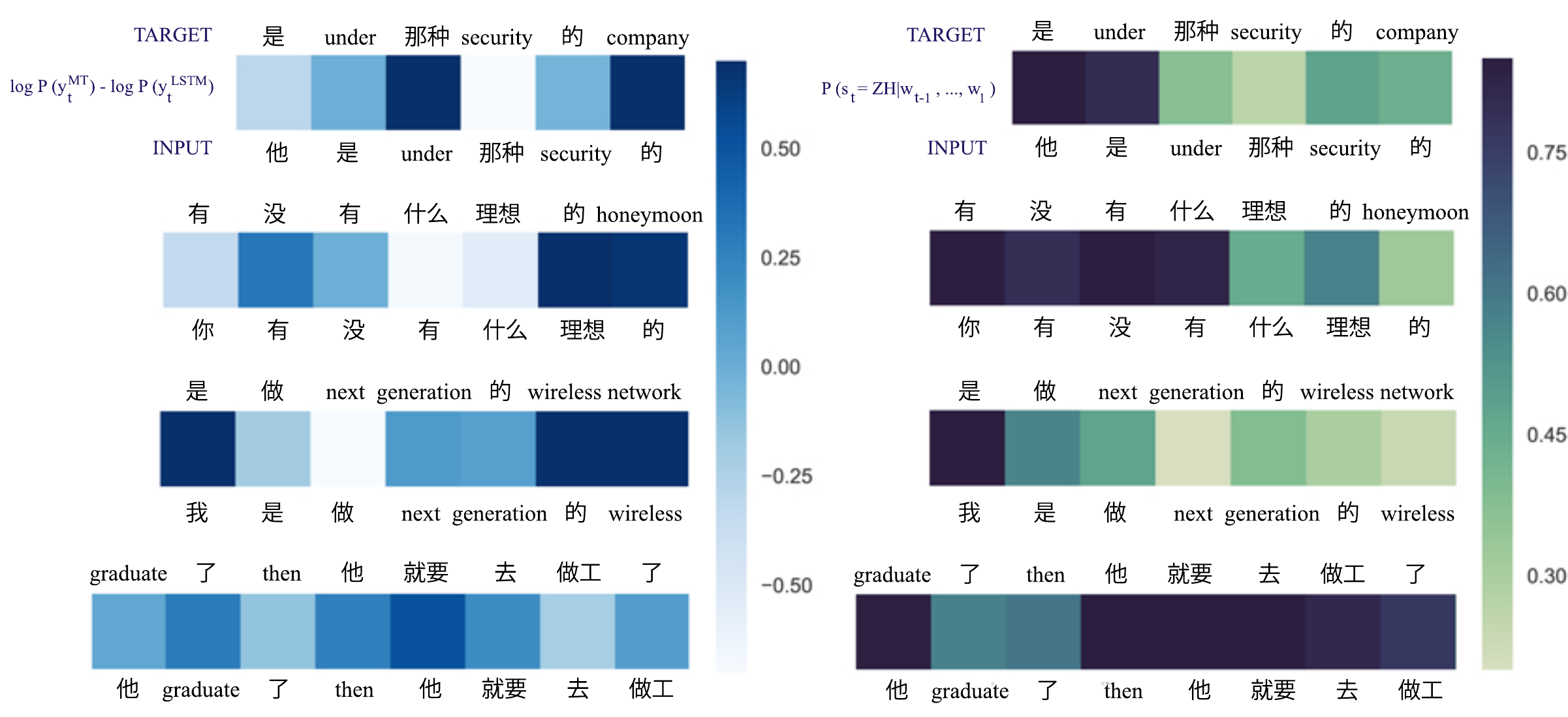}
  \caption{Prediction examples in Phase II. \textbf{Left:} Each square shows the target word's log probability improvement with the multi-task model compared to the LSTM model (darker color is better).
  \textbf{Right:} Each square shows the probability that the next POS tag is Chinese (darker color represents higher probability).}
  \label{fig:language}
\end{figure*}

Table \ref{results-weighted-loss-phase-1} and Table \ref{results-weighted-loss-phase-2} show the results of multi-task learning with different values of the hyper-parameter $p$. We observe that the multi-task model with $p = 0.25$ achieves the best performance. We compare our multi-task learning model against the RNNLM and LSTM baselines. The baselines correspond to RNNs that are trained with word lexicons. Table \ref{results-phase-1} and Table \ref{results-phase-2} present the overall results from the different models. The multi-task model performs better than the LSTM baseline by 9.7\% PPL in Phase I and 7.4\% PPL in Phase II. The performance of our model in Phase II is also better than the RNNLM (8.9\%) and far better than the one presented in~\cite{adel2013combination} in Phase I. Moreover, the results show that adding a shared POS tag representation to $\textnormal{LSTM}_{lm}$ does not hurt the performance of the LM task. This implies that the syntactic information helps the model to better predict the next word in the sequence. To further verify this hypothesis, we conduct two analyses by visualizing our prediction examples, as shown in Figure~\ref{fig:language}.

\begin{table}[!tb]
\centering
\caption{Multi-task results with different weighted loss hyper-parameters in SEAME Phase I.}
\label{results-weighted-loss-phase-1}
\begin{tabular}{@{}cccc@{}}
\toprule
\textbf{\begin{tabular}[c]{@{}c@{}}Hidden\\ size\end{tabular}}  & \multicolumn{1}{c}{\textbf{$p$}} & \multicolumn{1}{c}{\textbf{\begin{tabular}[c]{@{}c@{}}PPL Dev\end{tabular}}} & \multicolumn{1}{c}{\textbf{\begin{tabular}[c]{@{}c@{}}PPL Test\end{tabular}}} \\ \midrule
\multirow{3}{*}{200} & \multicolumn{1}{c}{\textbf{0.25}} & \multicolumn{1}{c}{\textbf{180.90}}    & \multicolumn{1}{c}{\textbf{178.18}} \\
 & \multicolumn{1}{c}{0.50} & \multicolumn{1}{c}{182.60} & \multicolumn{1}{c}{178.75}\\
 & \multicolumn{1}{c}{\textbf{0.75}} & \multicolumn{1}{c}{\textbf{180.90}} & \multicolumn{1}{c}{\textbf{178.18}} \\ \midrule
\multirow{3}{*}{500} & \multicolumn{1}{c}{\textbf{0.25}} & \multicolumn{1}{c}{\textbf{173.55}} & \multicolumn{1}{c}{\textbf{174.96}} \\
 & \multicolumn{1}{c}{0.50} & \multicolumn{1}{c}{175.23} & \multicolumn{1}{c}{173.89} \\
 & \multicolumn{1}{c}{0.75} & \multicolumn{1}{c}{185.83} & \multicolumn{1}{c}{178.49} \\ \bottomrule
\end{tabular}
\end{table}

\begin{table}[!tb]
\centering
\caption{Multi-task results with different weighted loss hyper-parameters in SEAME Phase II.}
\label{results-weighted-loss-phase-2}
\begin{tabular}{@{}cccc@{}}
\toprule
\textbf{\begin{tabular}[c]{@{}c@{}}Hidden\\ size\end{tabular}}  & \multicolumn{1}{c}{\textbf{$p$}} & \multicolumn{1}{c}{\textbf{\begin{tabular}[c]{@{}c@{}}PPL Dev\end{tabular}}} & \multicolumn{1}{c}{\textbf{\begin{tabular}[c]{@{}c@{}}PPL Test\end{tabular}}} \\ \midrule
\multirow{3}{*}{200} & \multicolumn{1}{c}{\textbf{0.25}} & \multicolumn{1}{c}{\textbf{149.68}}    & \multicolumn{1}{c}{\textbf{149.84}} \\
 & \multicolumn{1}{c}{0.5} & \multicolumn{1}{c}{150.92} & \multicolumn{1}{c}{152.38} \\ 
 & \multicolumn{1}{c}{0.75} & \multicolumn{1}{c}{150.32} & \multicolumn{1}{c}{151.22} \\ \midrule
\multirow{3}{*}{500} & \multicolumn{1}{c}{\textbf{0.25}} & \multicolumn{1}{c}{\textbf{141.86}} & \multicolumn{1}{c}{\textbf{141.71}} \\ 
 & \multicolumn{1}{c}{0.5} & \multicolumn{1}{c}{144.18} & \multicolumn{1}{c}{144.27} \\
 & \multicolumn{1}{c}{0.75} & \multicolumn{1}{c}{145.08} & \multicolumn{1}{c}{144.85} \\ \bottomrule
% \multirow{3}{*}{1000} & \textbf{0.25} & \textbf{x}    & \textbf{x} \\ \cline{2-4}
%  & 0.5 & x & x\\ \cline{2-4}
%  & 0.75 & x & x \\ \hline
\end{tabular}
\end{table}
\begin{table}[!tb]
\centering
\caption{Language model results in SEAME Phase I}
\label{results-phase-1}
\resizebox{0.54\textwidth}{!}{
\begin{tabular}{@{}lcc@{}}
\toprule
\textbf{Model} & \multicolumn{1}{c}{\textbf{\begin{tabular}[c]{@{}c@{}}PPL Dev\end{tabular}}} & \multicolumn{1}{c}{\textbf{\begin{tabular}[c]{@{}c@{}}PPL Test\end{tabular}}} \\ \midrule
RNNLM \cite{adel2013recurrent} & \multicolumn{1}{c}{246.60} & \multicolumn{1}{c}{287.88}\\ 
wFLM \cite{adel2015syntactic} & \multicolumn{1}{c}{238.86} & \multicolumn{1}{c}{245.40}\\ 
FI + OF \cite{adel2013recurrent} & \multicolumn{1}{c}{219.85} & \multicolumn{1}{c}{239.21}\\ 
RNNLM + FLM \cite{adel2013combination} & \multicolumn{1}{c}{177.79} & \multicolumn{1}{c}{192.08}\\ 
LSTM & \multicolumn{1}{c}{190.33} & \multicolumn{1}{c}{185.91} \\ 
\textnormal{+ syntactic features} & \multicolumn{1}{c}{178.51} & \multicolumn{1}{c}{176.57} \\ \midrule
\textbf{Multi-task} & \multicolumn{1}{c}{\textbf{173.55}} & \multicolumn{1}{c}{\textbf{174.96}} \\ \bottomrule
\end{tabular}
}
\end{table}
\begin{table}[!tb]
\centering
\caption{Language model results in SEAME Phase II}
\label{results-phase-2}
\resizebox{0.52\textwidth}{!}{
\begin{tabular}{@{}lcc@{}}
\toprule
\textbf{Model} & \multicolumn{1}{c}{\textbf{\begin{tabular}[c]{@{}c@{}}PPL Dev\end{tabular}}} & \multicolumn{1}{c}{\textbf{\begin{tabular}[c]{@{}c@{}}PPL Test\end{tabular}}} \\ \midrule
RNNLM \cite{adel2013recurrent} & \multicolumn{1}{c}{178.35} & 171.27 \\
LSTM & \multicolumn{1}{c}{150.65} & 153.06 \\
\textnormal{+ syntactic features} & \multicolumn{1}{c}{147.44} & 148.38 \\ \midrule
\textbf{Multi-task} & \multicolumn{1}{c}{\textbf{141.86}} & \textbf{141.71} \\ \bottomrule
\end{tabular}
}
\end{table}

To measure the target word's log probability with the multi-task model compared to the standard LSTM model, we calculate the log probability difference between the two models. According to Figure~\ref{fig:language}, in most cases, the multi-task model improves the prediction of the monolingual segments, and particularly at code-switching points, such as ``under", ``security", ``generation'', ``then", ``graduate", ``他", and ``的". It also shows that the multi-task model is more precise in learning where to switch the language. Meanwhile, Table \ref{trigger-words} shows the relative frequency of the trigger POS tag. The word ``then" belongs to $\textnormal{RB}_{EN}$, which is one of the most common trigger words in the list. Furthermore, the target word prediction is significantly improved in most of the trigger words. We report the probability that the next produced POS tag is Chinese. It is shown that the words ``then", ``security", ``了'', ``那种'', ``做'', and ``的''  tend to switch the language context within the utterance. However, it is very hard to predict all the cases correctly. This may be due to the fact that without any switching, the model still creates a correct sentence.

\section{Short Summary}
In this chapter, we propose a multi-task learning approach for code-switched LM that leverages syntactic information. The multi-task learning models achieve the best performance and outperform the LSTM baseline with 9.7\% and 7.4\% improvement in PPL for the Phase I and Phase II SEAME corpus, respectively. This implies that by training two different NLP tasks together, the models can correctly learn the correlation between them. Indeed, the syntactic information helps the model to be aware of code-switching points, and it improves the performance over the LM. Finally, we conclude that multi-task learning has good potential for code-switching LM research.
\end{CJK*}
\chapter{Multi-Task Learning for End-to-End Code-Switched Speech Recognition}

ASR is the first step in the pipeline of a conversational agent and positively affects the response of the overall system. Any errors made by the ASR system will propagate through other modules and lead to failures in the conversation. Thus, robust ASR systems are crucial. Since code-switched data are limited, transfer learning approaches have become very important for domain adaptation.~\citet{li2012code} proposed to first train an English-Chinese bilingual model using monolingual datasets, and then fine-tune the model using a small number of code-switched data. Then, they also augment the synthetic code-mixed speech with linguistic constraints to improve the language model re-scoring.

In this chapter, we focus on applying an end-to-end neural-based approach without using any LID. Since errors in LID affect the prediction of the model, we aim to remove dependency on the language. And since code-switching is considered low-resource, we take a transfer learning approach to leverage monolingual resources for the code-switching domain. We examine different training approaches to train code-switching end-to-end speech recognition models, and explore transfer learning methods to improve code-switching speech recognition to address the low-resource issue. Commonly, high-resource monolingual speech data are used to improve the generalization ability of code-switching models as an adaptation method. Intuitively, this approach may help the model recognize individual languages, which can eventually distinguish phones from different languages. We will introduce two training strategies: multi-task training and meta-transfer learning.

\section{Model Description}
\subsection{Architecture}
\begin{figure}[!ht]
    \centering
    \includegraphics[width=0.4\linewidth]{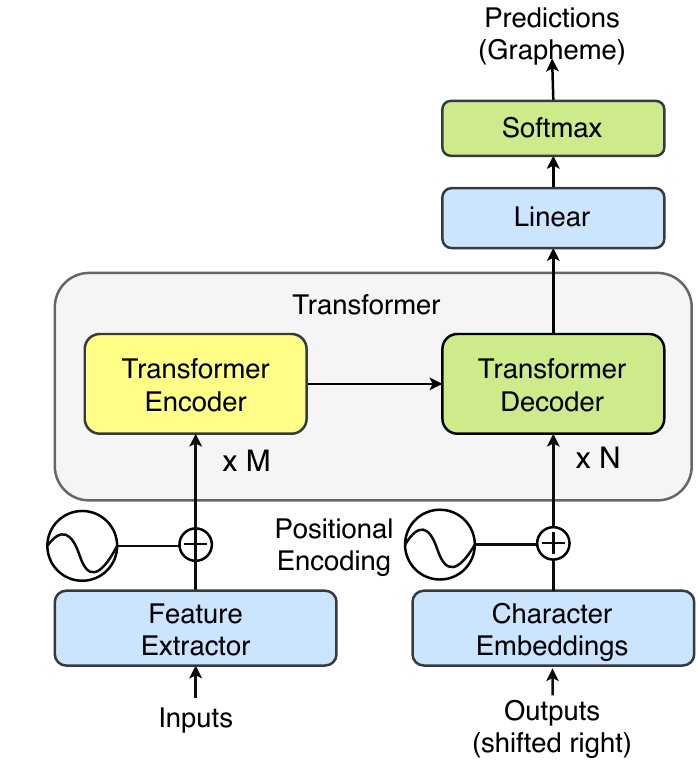}
    \caption{Transformer ASR model architecture.}
  \label{fig:transformer-asr2}
\end{figure}

We build our speech recognition model on a transformer-based encoder-decoder to learn to predict graphemes from the speech input. Our model extracts audio inputs with a learnable feature extractor module to generate input embeddings. The model is illustrated in Figure~\ref{fig:transformer-asr2}. The encoder uses input embeddings generated from the feature extractor module. The decoder receives the encoder outputs and applies multi-head attention to its input to finally calculates the logits of the outputs. To generate the probability of the outputs, we compute the softmax function of the logits. We apply a mask in the attention layer to avoid any information flow from future tokens, and train our model by optimizing the next-step prediction on the previous characters and maximizing the log probability:
\begin{equation}
\max_{\theta}\sum_{i}\log{P(y_i|x,y^{\prime}_{<i};\theta)},
\end{equation}
\noindent where $x$ is the character inputs, $y_i$ is the next predicted character, and $y^{\prime}_{<i}$ is the previous groundtruth characters. 

% In the inference time, we generate the sequence using a beam-search in the auto-regressive manner. The maximize the following scoring function: (Noisy channel?)
% \begin{equation} \alpha \sum_{i}\log{P(y_i|x,\hat{y}_{<i};\theta)} + \gamma \sqrt{wc(\hat{y}_{<i})},  
% \end{equation}

% \noindent where $\alpha$ is the parameter to control the decoding probability from the decoder and $\gamma$ is the parameter to control the effect of the word count $wc(\hat{y}_{<i})$

\subsection{Inference}
To further improve the prediction, we incorporate \textit{Pointer-Gen LM}~\cite{winata2019code} in the beam search process to select the best sub-sequence scored using the softmax probability of the characters. We define $P(Y)$ as the probability of the predicted sentence, and add the Pointer-Gen LM $P_{lm}(Y)$ to rescore the predictions. We also include word count \textit{wc(Y)} to avoid generating very short sentences. $P(Y)$ is calculated as follows:
\begin{equation} \alpha \sum_{i}\log{P(y_i|x,\hat{y}_{<i};\theta)} + \beta P_{lm}(\hat{y}_{<i}) + \gamma \sqrt{wc(\hat{y}_{<i})},  
\end{equation}

\noindent where $\alpha$ is the parameter to control the decoding probability from the decoder, $\beta$ is the parameter to control the language model probability, and $\gamma$ is the parameter to control the effect of the word count $wc(\hat{y}_{<i})$.

\section{Training Strategies}

\subsection{Standard Multi-Task Training}
To learn information from several datasets, we can jointly train all of them together~\cite{winata2018towards}. This training strategy enables the model to generalize well on all tasks. Figure~\ref{fig:meta_transfer_learning_architecture} (a) shows the intuition of how the trained parameter $\theta$ is in the center of all tasks. The algorithm is relatively simple as it draws the same number of samples from each task and trains them together in the gradient update. By learning in this way, the model can capture information from all sources, which is useful in the case of code-switching because there is overlapping information between code-switching and monolingual sources.

\subsection{Meta-Transfer Learning}
We propose our Meta-Transfer Learning by first introducing Model-Agnostic Meta-Learning (MAML).
\subsubsection{Model-Agnostic Meta-Learning}
\begin{figure}[!ht]
    \centering
    \includegraphics[scale=0.7]{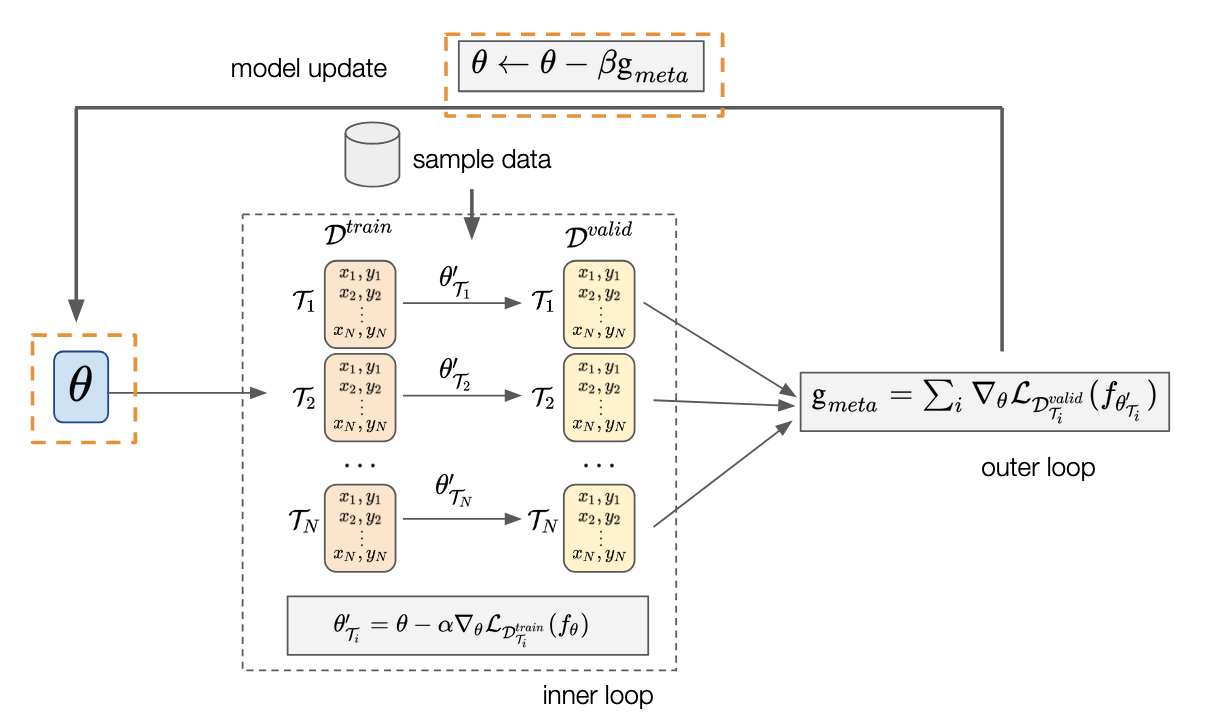}
    \caption{Model-Agnostic Meta-Learning (MAML) training mechanism.}
    \label{fig:maml}
\end{figure}

MAML~\cite{finn2017model,winata2020learning} learns to quickly adapt to a new task from a number of different tasks using a gradient descent procedure, as shown in Figure~\ref{fig:maml}. 
We apply MAML to effectively learn from a set of accents and quickly adapt to a new accent in the few-shot setting. We denote our transformer ASR as $f_{\theta}$ parameterized by $\theta$. Our dataset consists of a set of accents $\mathcal{A} = \{A_1,A_2,\cdots,A_n \}$, and for each accent $i$, we split the data into $A_i^{tra}$ and $A_i^{val}$. Then we update $\theta$ into $
\theta^{\prime}$ by computing gradient descent updates on $A_i^{tra}$:
\begin{equation}
\theta_{i}^{\prime} = \theta - \alpha \nabla_{\theta}{\mathcal{L}_{A_{i}^{tra}}(f_{\theta})},
\end{equation}
\noindent where $\alpha$ is the fast adaptation learning rate. During the training, the model parameters are trained to optimize the performance of the adapted model $f(\theta_i^{\prime})$ on unseen $A_i^{val}$. The meta-objective is defined as follows:
\begin{align}
\min_{\theta}\sum_{A_i\sim p(\mathcal{A})}\mathcal{L}_{A_i^{val}}(f_{\theta_i'}) = \sum_{A_i\sim p(\mathcal{A})}\mathcal{L}_{A_i^{val}}(f_{\theta-\alpha \nabla_{\theta} \mathcal{L}_{A_i^{tra}}(f_{\theta})}),
\end{align}

\noindent where $\mathcal{L}_{A_i^{val}}(f_{\theta_i'})$ is the loss evaluated on $A_i^{val}$. We collect the loss $\mathcal{L}_{A_i^{val}}(f_{\theta_i'})$ from a batch of accents and perform the meta-optimization as follows:
\begin{align}
\label{eq:5}
\theta \leftarrow \theta - \beta {\sum_{A_i\sim p(\mathcal{A})}\nabla_{\theta}\mathcal{L}_{A_i^{val}}(f_{\theta_i'})},
\end{align}

\noindent where $\beta$ is the meta step size and $f_{\theta_i^{\prime}}$ is the adapted network on accent $A_i$. The meta-gradient update step is performed to achieve a good initialization for our model. Then we can optimize our model with a small number of samples on target accents in the fine-tuning step. In this chapter, we use first-order approximation MAML as in~\cite{gu2018meta} and~\cite{finn2018pmaml}. Thus, Equation \ref{eq:5} is reformulated as
\begin{align}
\theta \leftarrow \theta - \beta {\sum_{A_i\sim p(\mathcal{A})}\nabla_{\theta_i'}\mathcal{L}_{A_i^{val}}(f_{\theta_i'})}.
\end{align}

\subsubsection{Our Approach}
\begin{figure}[!t]
    \centering
    \includegraphics[scale=1.9]{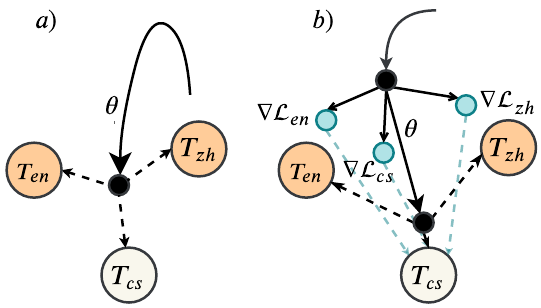}
    \caption{Illustration of (a) joint training and (b) Meta-Transfer Learning. The solid lines show the optimization path. The orange circles represent the monolingual source language, and the white circles represent the code-switching target language. The lower black circle in (b) is closer to $T_{cs}$ than that in (a).}
    \label{fig:meta_transfer_learning_architecture}
\end{figure}

Our approach extends the meta-learning paradigm to adapt knowledge learned from source domains to a specific target domain. Compared to MAML, we only assign code-switching data in $\mathcal{D}^{val}$ and take both monolingual and code-switching data in $\mathcal{D}^{train}$. So, the model is only tuned from the gradients evaluated from code-switching data. This approach captures useful information from multiple resources for the target domain and updates the model accordingly. Figure~\ref{fig:meta_transfer_learning_architecture} presents the general idea of our Meta-Transfer Learning. Its goal is not to generalize to all tasks but to focus on acquiring crucial knowledge to transfer from monolingual resources to the code-switching domain. As shown in Algorithm 1, for each adaptation step on $\mathcal{T}_i$, we compute updated parameters $\theta_i'$ via SGD as follows:
\begin{align}
    \theta_i' = \theta - \alpha \nabla_{\theta}\mathcal{L}_{\mathcal{D}_i^{tra}}(f_{\theta}),
\end{align}
where $\alpha$ is a learning hyper-parameter of the inner optimization. Then, a cross-entropy loss $\mathcal{L}_{D_i}$ is calculated from a learned model upon the generated text given the audio inputs on the target domain $j$:
\begin{align}
    \mathcal{L}_{D_i} = - \sum_{\mathcal{D}^{val}\sim \mathscr{D}_{tgt}} \log p(y_t|x_t;\theta'_i).
\end{align}
We define the objective as follows:
\begin{align}
    &\min_{\theta}\sum_{\mathcal{D}_i^{tra}, \mathcal{D}^{val}}\mathcal{L}_{\mathcal{D}^{val}}(f_{\theta_i'}) = \\
    &\sum_{\mathcal{D}_i^{tra}, \mathcal{D}^{val} }\mathcal{L}_{\mathcal{D}^{val}}(f_{\theta-\alpha \nabla_{\theta} \mathcal{L}_{D_i^{tra}}(f_{\theta})}),
\end{align}
where $\mathcal{D}_i^{tra}\sim (\mathscr{D}_{src},\mathscr{D}_{tgt})$ and $\mathcal{D}^{val}\sim \mathscr{D}_{tgt}$. We minimize the loss of the $f_{\theta'_i}$ upon $\mathcal{D}^{val}$. Then, we apply gradient descent on the meta-model parameter $\theta$ with a $\beta$ meta-learning rate.

\begin{algorithm}[!ht]
{\selectfont
\caption{Meta-Transfer Learning}
\label{alg1}
\textbf{Require:} $\mathscr{D}_{src}$, $\mathscr{D}_{tgt}$\\
\textbf{Require:} $\alpha, \beta$: step size hyper-parameters
\begin{algorithmic}[1]
\State Randomly initialize $\theta$
\While{not done}
  \State Sample batch data $\mathcal{D}^{tra}_{i} \sim (\mathscr{D}_{src}, \mathscr{D}_{tgt})$, $\mathcal{D}^{val} \sim \mathscr{D}_{tgt}$
  \For{\textbf{all} $\mathcal{D}^{tra}_{i}$}
      \State Evaluate $\nabla_{\theta}\mathcal{L}_{\mathcal{D}^{tra}_{i}}(f_{\theta})$ using $\mathcal{D}^{tra}_{i}$
      \State Compute adapted parameters with gradient descent:
    \phantom .\phantom  .
    $\theta ^ { \prime }_{i} = \theta - \alpha \nabla_{\theta}\mathcal{L}_{\mathcal{D}^{tra}_{i}}(f_{\theta})$
\EndFor
  \State $\theta \leftarrow \theta-\beta\sum_{i} \nabla_{\theta}\mathcal{L}_{\mathcal{D}^{val}}\left(f_{\theta_{i}^{\prime}}\right)$  
  
\EndWhile
\end{algorithmic}
}
\end{algorithm}

\section{Experiments and Results}

\subsection{Datasets}
We use SEAME Phase II, a conversational
English-Mandarin Chinese code-switching speech corpus that consists of spontaneously spoken interviews and conversations~\cite{SEAME2015}. 
%We split the corpus following~\citet{winata2018code}. 
The details are depicted in  Table~\ref{data-statistics-phase-2}. For the monolingual speech datasets, we use HKUST~\cite{liu2006hkust} as the monolingual Chinese dataset and Common Voice v1 as the monolingual English dataset.\footnote{The dataset is available at https://voice.mozilla.org/.} We use 16 kHz audio inputs and up-sample the HKUST data from 8 to 16 kHz.

% The training data are randomly shuffled every epoch. Our character set is the concatenation of English letters, Chinese characters found in the corpus, spaces, and apostrophes.

\subsection{Experiment Settings}
Our transformer model consists of two encoder and four decoder layers with a hidden size of 512, an embedding size of 512, a key dimension of 64, and a value dimension of 64. The input of all experiments is a spectrogram, computed with a 20 ms window, and shifted every 10 ms. Our label set has 3765 characters and includes all English and Chinese characters from the corpora, spaces, and apostrophes. We optimize our model using Adam and start the training with a learning rate of 1e-4. We fine-tune our model using SGD with a learning rate of 1e-5, and apply an early stop on the validation set. We choose $\alpha=1$, $\beta=0.1$, and $\gamma=0.1$. We draw the sample of the batch randomly with a uniform distribution every iteration. We define \textit{cs} as the code-switching dataset, \textit{en} as the English dataset, and \textit{zh} as the Chinese dataset. We conduct experiments with the following approaches: \textbf{(a)} only \textit{cs}, \textbf{(b)} joint training on \textit{en} + \textit{zh}, \textbf{(c)} joint training on \textit{cs} + \textit{en} + \textit{zh}, and \textbf{(d)} Meta-Transfer Learning. Then, we fine-tune the trained models \textbf{(b)}, \textbf{(c)}, and \textbf{(d)} on \textit{cs}, and we apply language model rescoring on our best model. We evaluate our model using beam search with a beam width of 5 and maximum sequence length of 300. The quality of our model is measured using CER.

% We evaluate our proposed sentence generation method on an end-to-end ASR system. Table \ref{asr-evaluation} shows the CER of our ASR systems, as well as the individual CER on each language. Based on the experimental results, pretraining is able to reduce the error rate by 1.64\%, as it corrects the spelling mistakes in the prediction.~After we add LM (rCS) to the decoding step, the error rate can be reduced to 32.25\%. Finally, we replace the LM with LM (Pointer-Gen $\rightarrow$ rCS), and it further decreases the error rate by 1.18\%.

\begin{CJK*}{UTF8}{gbsn}
\paragraph{Multi-task Learning End-to-end ASR} We convert the inputs into normalized frame-wise spectrograms from 16-kHz audio. Our transformer model consists of two encoder and decoder layers. The Adam optimizer and Noam warmup are used for training with an initial learning rate of 1e-4. The model has a hidden size of 1024, a key dimension of 64, and a value dimension of 64. The training data are randomly shuffled every epoch. Our character set is the concatenation of English letters, Chinese characters found in the corpus, spaces, and apostrophes.
% ~\textnormal{\tt \{a-z, space, apostrophe, 祥, 舌, ..., 底 \}}.
In the multilingual ASR pre-training, we train the model for 18 epochs. Since the sizes of the datasets are different, we over-sample the smaller dataset. The fine-tuning step takes place after the pretraining using code-switching data.~In the inference time, we explore the hypothesis using beam search with a beam width of 8 and a batch size of 1.
\end{CJK*}

\paragraph{Character Error Rate (CER)} For our ASR, we compute the overall CER and also show the individual CERs for Mandarin Chinese \textbf{(zh)} and English \textbf{(en)}. The metric calculates the distance of two sequences as the Levenshtein Distance. 

% \begin{table}[!h]
% \caption{ASR evaluation, showing the performance on all sequences: Overall, English segments (en), and Mandarin Chinese segments (zh).}
% \centering
% \resizebox{0.65\textwidth}{!}{
% \begin{tabular}{lccc}
% \toprule
% \textbf{Model} & \textbf{Overall} & \textbf{en} & \textbf{zh}\\ \midrule
% % Baseline & 34.40\% & 41.70\% & 35.84\% \\
% Baseline & 34.40\% & 41.79\% & 35.94\% \\
% + Pre-training & 32.76\% & 40.06\% & 32.44\% \\
% \hspace{3mm}\text{+ LM (rCS)} & 32.25\% & 39.45\% & 31.90\% \\ \midrule
% % \hspace{3mm}\textbf{+ LM (4c)} & \text{31.75\%} & \text{39.03\%} & \text{31.37\%} \\ \hline
% % \hspace{3mm}\textbf{+ LM (4c)} & \text{31.55\%} & \text{39.24\%} & \text{31.05\%} \\ \hline
% % \hspace{3mm}\textbf{+ LM (4c)} & \text{31.31\%} & \text{38.66\%} & \text{30.93\%} \\ \hline
% \hspace{3mm}\textbf{+ LM (Pointer-Gen $\rightarrow$ rCS)} & \textbf{31.07\%} & \textbf{38.39\%} & \textbf{30.85\%} \\ \bottomrule
% \end{tabular}
% }
% \label{asr-evaluation}
% \end{table}

\begin{table}[!t]
\centering
\caption{Results of the evaluation on CER. Lower CER is better. \textbf{\textit{Meta-Transfer Learning}} is more effective in transferring information from the monolingual datasets.}
\resizebox{0.45\textwidth}{!}{
\begin{tabular}{lc}
\toprule
 & $\downarrow$ \textbf{CER} \\ \midrule
~\citet{winata2019code} & 32.76\% \\
\hspace{3mm}+ Pointer-Gen LM & 31.07\% \\ \midrule
Only \textit{cs} & 34.51\% \\ \midrule
Joint Training (\textit{en} + \textit{zh}) & 98.29\% \\
\hspace{3mm}+ Fine-tuning & 31.22\% \\
Joint Training (\textit{en} + \textit{zh} + \textit{cs}) & 32.87\% \\
\hspace{3mm}+ Fine-tuning & 31.90\% \\ \midrule
Meta-Transfer Learning & 30.30\% \\
\hspace{3mm}+ Fine-tuning & 29.99\% \\
\hspace{3mm}+ Pointer-Gen LM & \textbf{29.30\%} \\ \bottomrule
\end{tabular}
}
\label{results}
\end{table}

\subsection{Results}
The results are shown in Table~\ref{results}. Generally, adding monolingual data \textit{en} and \textit{zh} as the training data is effective to reduce the error rate. There is a significant margin between \textbf{only \textit{cs}} and \textbf{joint training} (1.64\%) or \textbf{meta-transfer learning} (4.21\%). According to the experiment results, \textbf{meta-transfer learning} consistently outperforms the joint-training approaches, which shows its effectiveness in language adaptation. The fine-tuning approach helps to improve the performance of the trained models, especially on the jointly trained (\textit{en} + \textit{zh}). We observe that joint training on (\textit{en} + \textit{zh}) without fine-tuning cannot predict mixed-language speech, while joint training on \textit{en} + \textit{zh} + \textit{cs} is able to recognize it. However, according to Table~\ref{comparison}, adding a fine-tuning step badly affects the previously learned knowledge (e.g., \textit{en}: 11.84\%$\rightarrow$63.85\%, \textit{zh}: 31.30\%$\rightarrow$78.07\%). Interestingly, the model trained with Meta-Transfer Learning does not suffer catastrophic forgetting, even without focusing the loss objective to learn both monolingual languages. As expected, joint training on \textit{en + zh + cs} achieves decent performance on all tasks, but it does not optimally improve \textit{cs}.

The language model rescoring using Pointer-Gen LM improves the performance of the Meta-Transfer Learning model by choosing more precise code-switching sentences during beam search. Pointer-Gen LM improves the performance of the model and outperforms the model trained only in \textit{cs} by 5.21\% and the previous state-of-the-art by 1.77\%.

\begin{table}[!t]
\centering
\caption{Performance on monolingual English (\textbf{en}) and Chinese (\textbf{zh}) in terms of CER. $\Delta$ CER denotes the improvement on \textit{cs} test set relative to the baseline (Only \textit{cs}) model.}
\resizebox{0.68\textwidth}{!}{
\begin{tabular}{lccc} 
\toprule & $\uparrow$ $\Delta$\textbf{ CER}
 & $\downarrow$ \textbf{en} & $\downarrow$ \textbf{zh} \\ \midrule
Only \textit{cs} & 0 & 66.71\% & 99.66\% \\ \midrule
Joint Training (\textit{en} + \textit{zh}) & -63.78\% & 11.84\% & 31.30\% \\
\hspace{3mm}+ Fine-tuning & 3.29\% & 63.85\% & 78.07\% \\  
Joint Training (\textit{en} + \textit{zh} + \textit{cs}) & 1.64\% & 13.88\% & 30.46\% \\
\hspace{3mm}+ Fine-tuning & 2.61\% & 57.56\% & 76.20\% \\  
\midrule
\textbf{Meta-Transfer Learning} & \textbf{4.21\%} & 16.22\% & 31.39\% \\ \bottomrule
\end{tabular}
}
\label{comparison}
\end{table}

\begin{figure}[!t]
    \centering
    \includegraphics[scale=0.56]{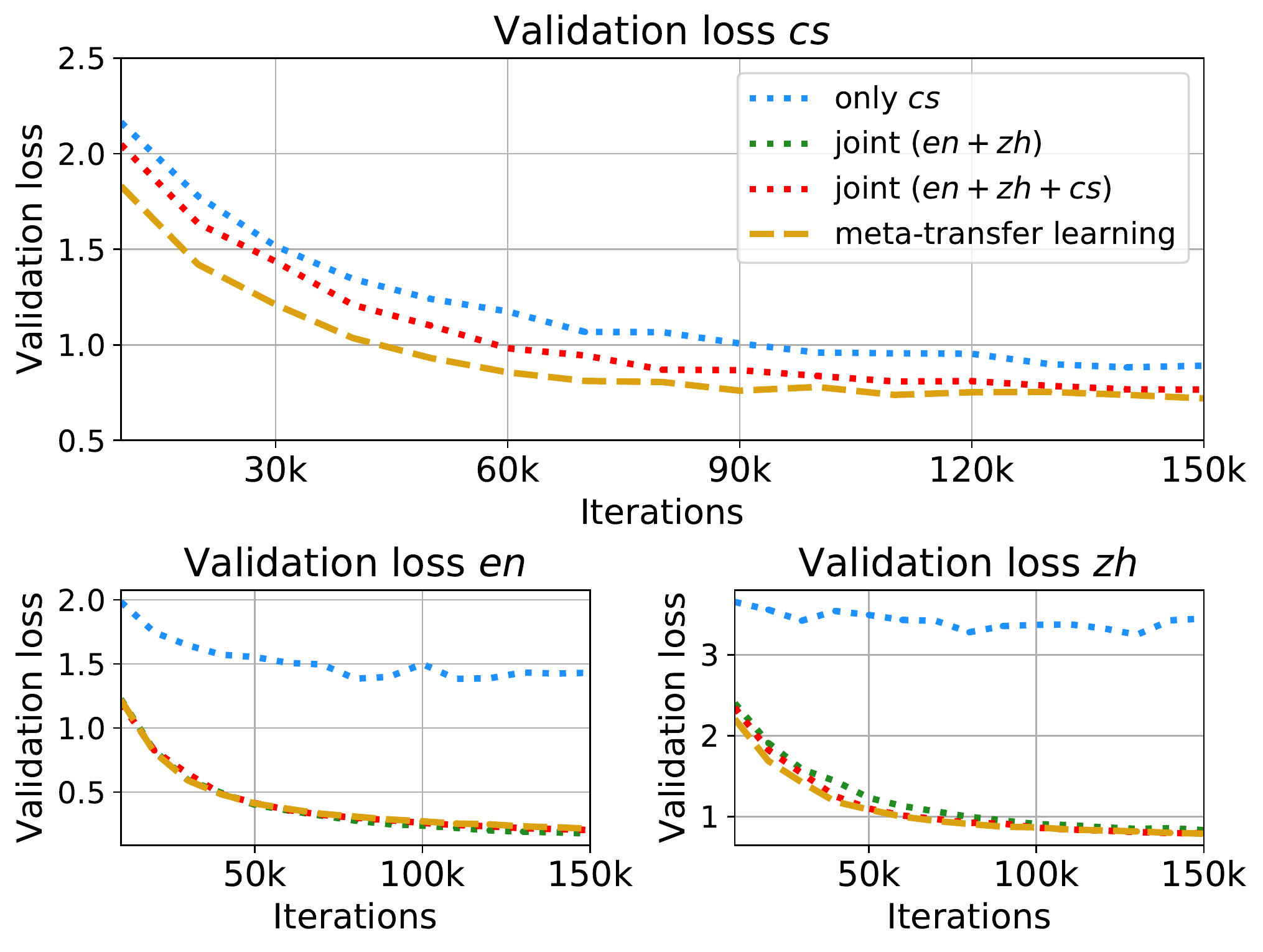}
    \caption{Validation loss per iteration. \textbf{Top:} validation loss on \textit{cs} data, 
    (joint ($en$ + $zh$) is omitted because it is higher than the range), 
    \textbf{bottom left:} validation loss on $en$ data, \textbf{bottom right:} validation loss on $zh$ data.}
    \label{fig:validation_loss}
\end{figure}

\paragraph{Convergence Rate} Figure~\ref{fig:validation_loss} depicts the dynamics of the validation loss per iteration on \textit{cs}, \textit{en}, and \textit{zh}. As we can see from the figure, Meta-Transfer Learning is able to converge faster than joint training, and ends in the lowest validation loss. For the validation losses on \textit{en} and \textit{zh}, both joint training (\textit{en} + \textit{zh} + \textit{cs}) and Meta-Transfer Learning achieve similar loss in the same iteration. This shows that Meta-Transfer Learning is not only optimized for the code-switching domain, but also preserves the generalization ability to monolingual domains, as depicted in Table~\ref{comparison}.
\paragraph{Language Model}
\begin{table}[!ht]
\centering
\caption{Results on the language modeling task in PPL.}
\resizebox{0.63\textwidth}{!}{
\begin{tabular}{lcc}
\toprule
\multirow{1}{*}{\textbf{Model}} & \multicolumn{1}{c}{\textbf{valid}} & \multirow{1}{*}{\textbf{test}} \\ \midrule
Only \textit{cs} & 72.89 & 65.71  \\ \midrule
% Pointer-Gen LM$^{\ddagger}$ & 66.08 & 59.74  \\ \hline
Joint Training (\textit{en} + \textit{zh} + \textit{cs}) & \multicolumn{1}{c}{70.99} & 63.73 \\ 
\hspace{3mm}+ Fine-tuning & 69.66 & 62.73 \\ \midrule
Meta-Transfer Learning (\textit{en} + \textit{zh} + \textit{cs}) & \multicolumn{1}{c}{68.83} & 62.14 \\ 
\hspace{3mm}+ Fine-tuning & \textbf{68.71} & \textbf{61.97} \\ \bottomrule
\end{tabular}
}
\label{lm-results-mtl}
\end{table}

We further evaluate our meta-transfer learning approach on a language model task. We simply take the transcription of the same datasets and build a two-layer LSTM-based language model following the model configuration in~\citet{winata2019code}. To further improve the performance, we apply fine-tuning with an SGD optimizer by using a learning rate of 1.0, and decay the learning rate by 0.25x for every epoch without any improvement in the validation performance. To prevent the model from over-fitting, we add an early stop of 5 epochs. As shown in Table~\ref{lm-results-mtl}, the Meta-Transfer Learning approach outperforms the joint-training approach. We find a similar trend for the language model task results on the speech recognition task, where Meta-Transfer Learning without additional fine-tuning performs better than joint training with fine-tuning. Compared to our baseline model (Only \textit{cs}), Meta-Transfer Learning is able to reduce the test set PPL by 3.57 points (65.71 $\rightarrow$ 62.14), and the post fine-tuning step reduces the test set PPL even further, from 62.14 to 61.97.

\section{Short Summary}
In this chapter, we propose a novel multi-task learning method, Meta-Transfer Learning, to transfer learn on a code-switched speech recognition system in a low-resource setting by judiciously extracting information from high-resource monolingual datasets. Our model recognizes individual languages and transfers them to better recognize mixed-language speech by conditioning the optimization objective to the code-switching domain. Experimental results show that our model outperforms existing baselines in terms of error rate, and it is also faster to converge.
\chapter{Representation Learning for Code-Switched Sequence Labeling}

Sequence labeling, such as POS tagging and NER, is a key module for NLU systems. This module is important in understanding the semantics of the word sequence. Training a code-switched model is very challenging because monolingual taggers erroneously tag tokens, and the semantics of a word may be different between languages. The other problem in code-switched sequence labeling is how to represent embeddings on code-switched sequences, and many existing code-switched datasets are not annotated with language information. In code-switching, there is ambiguity in the semantics since the same words may be found in two or more languages. The standard approach in working on code-switching is to first predict the language and then take a corresponding embedding according to the predicted language. However, this may not be accurate, when the same word appears in the code-switched languages. For instance, the word ``cola'' can be associated with a product or is the word for ``queue'' in Spanish. Thus, the challenge in code-switching is learning a language-agnostic representation without any information about the language of the word.

Learning representation for code-switching has become a prominent area of research in NLP applications to support a greater variety of language speakers. However, despite the enormous number of studies in multilingual NLP, only very few focus on code-switching. Recently, contextualized language models, such as mBERT~\cite{devlin2019bert} and XLM-R~\cite{conneau2020unsupervised} have been proposed to tackle monolingual and cross-lingual tasks in NLU benchmarks~\cite{wang2018glue,hu2020xtreme,wilie2020indonlu,liu2020attention,lin2020xpersona}, and they have achieved impressive performance. The effectiveness and efficiency of these multilingual language models on code-switching tasks, however, remains unknown. Thus, in this chapter, we will propose an effective approach to improve the representation of code-switching sentences that is efficient enough to be applied in practical scenarios.

\noindent In the following sections, we will present three key contributions:
\begin{itemize}
    \item We introduce a method to train bilingual character embeddings using a BiLSTM for learning representations on code-switched data to show the effectiveness of combining words and characters. This method is very effective to address the OOV issue.
    \item We propose multilingual meta-embedding approaches for learning representations on code-switching data using neural-based models to combine embeddings from different languages without LID. 
    \item We also present a comprehensive study on the effectiveness of multilingual models on a variety of code-switching NLU tasks to analyze the practicality of each model in terms of performance, speed, and number of parameters. We further analyze the memory footprint required by each model over different sequence lengths in a GPU. Thus, we are able to understand which model to choose in a practical scenario.
\end{itemize}
% First, we describe the existing work on representational learning on natural language processing, particularly on natural language understanding. Then, we describe our approaches to represent code-switching sequences..

\section{Bilingual Character Embeddings}
\subsection{Model Architecture}
\begin{figure}[!htb]
  \centering
  \includegraphics[width=0.52\linewidth]{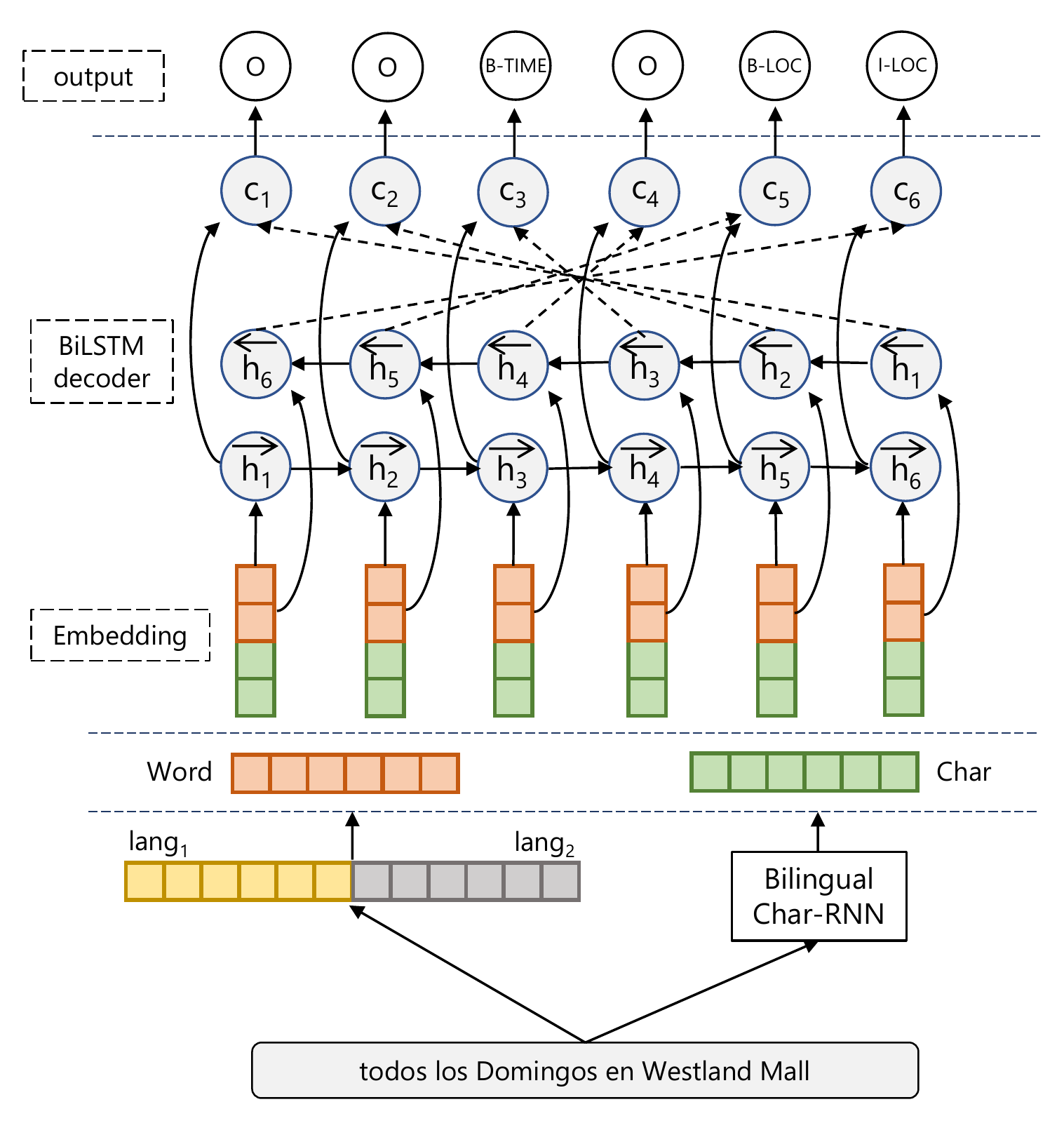}
  \caption{Word and character embeddings for sequence labeling.}
  \label{fig:word_char}
\end{figure}

Combining lexical structures (word, subword, and characters) is a common technique to improve representations by leveraging lexical composition.~\citet{lample-etal-2016-neural} proposed character-based word representations learned from a supervised corpus and unsupervised word representations learned from unannotated corpora. In the context of code-switching, we have explored a mixture of word and character embeddings~\cite{winata2018bilingual}. Figure~\ref{fig:word_char} shows the combination of word embeddings and embeddings generated from a bilingual character RNN. We use a word embedding taken from either \textit{$L_1$} or \textit{$L_2$}.
Consider a sequence of word embeddings $\textbf{x} = {(x_1, x_2, ..., x_M)}$ and character embeddings $\textbf{a} = {(a_1, a_2, ..., a_N)}$, where $M$ is the length of the word sequence and $N$ is the length of the character sequence. The word embeddings are fixed. Then, we concatenate both word and character vectors to get a richer word representation $u_t$. Afterwards, we pass the vectors to a BiLSTM. 
\begin{align}
 u_t &= x_t \oplus a_t, \\
\overrightarrow{h_t} &= \overrightarrow{\textnormal{LSTM}}(u_t, \overrightarrow{h_{t-1}}),\\ \overleftarrow{h_t} &= \overleftarrow{\textnormal{LSTM}}(u_t, \overleftarrow{h_{t-1}}), \\
c_t &= \overrightarrow{h_t} \oplus \overleftarrow{h_t},
\end{align}
where $\oplus$ denotes the concatenation operator. Dropout is applied to the recurrent layer. At each time step we make a prediction for the entity of the current token. The softmax function is used to calculate the probability distribution of all possible named-entity tags:
\begin{align}
y_t = \frac{e^{c_t}}{\sum_{j=1}^T e^{c_j}} \textnormal{, where } j = 1 \textnormal{, .., T},
\end{align}
where $\textnormal{y}_t$ is the probability distribution of tags at word $t$ and $\textnormal{T}$ is the maximum time step. Since there is a variable number of sequence lengths, we pad the sequence and apply a mask when calculating the cross-entropy loss function. Our model does not use a gazetteer or knowledge-based information, and it can be easily adapted to another language pair.

\subsubsection{Bilingual Char-RNN} 
\begin{figure}[!htb]
  \centering
  \includegraphics[width=0.7\linewidth]{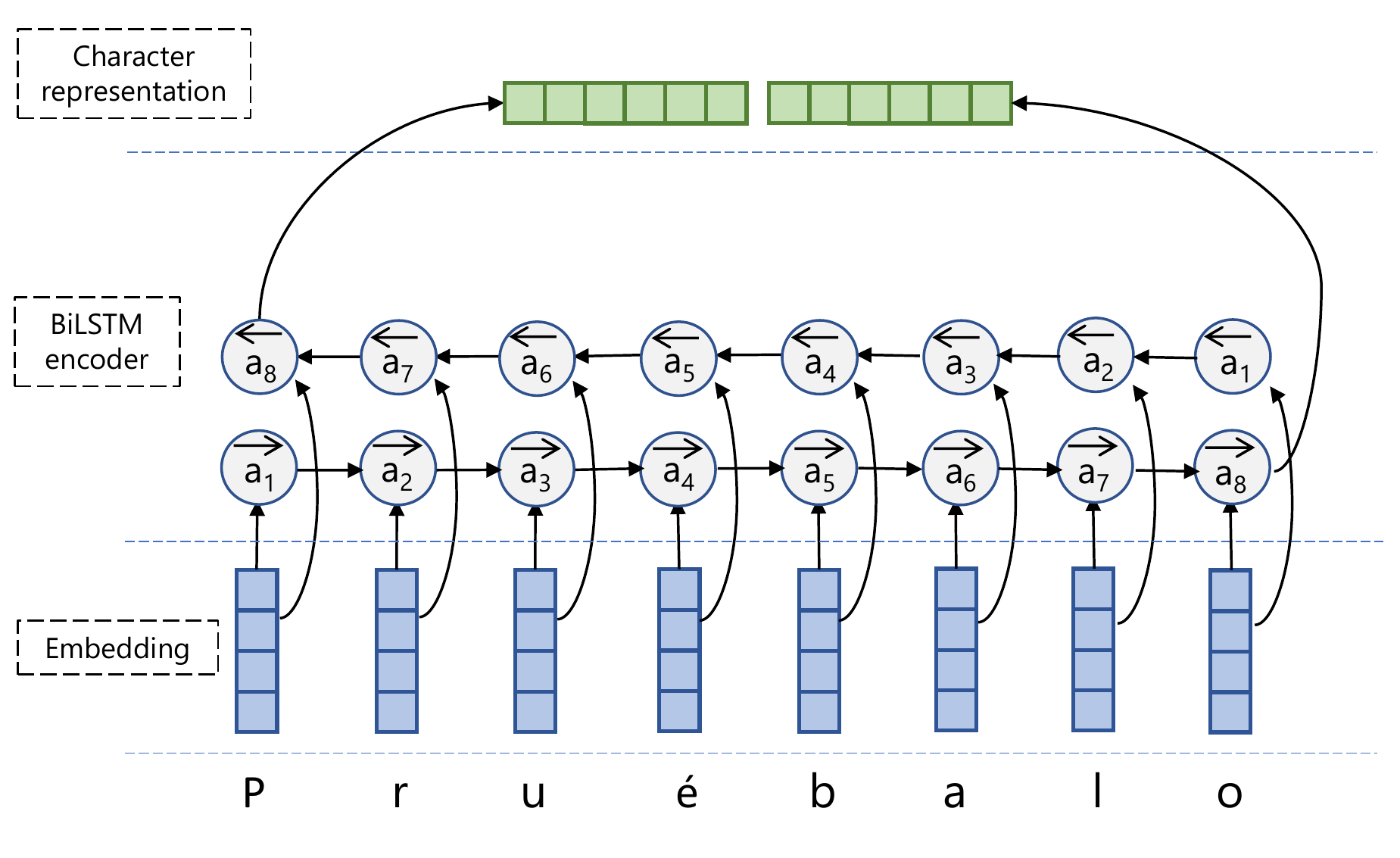}
  \caption{Bilingual Char-RNN architecture.}
  \label{fig:char-rnn}
\end{figure}

We use an RNN to represent a word with character-level information \cite{lample-etal-2016-neural}. Figure \ref{fig:char-rnn} shows the model architecture of our Char-RNN. The inputs are characters extracted from a word, and every character is embedded with a $d$-dimension vector. Then, we use this as the input for a BiLSTM as a character encoder, wherein, at every time step, a character is input to the network.

\subsubsection{Pre-processing}
In this section, we describe word-level and character-level features used in our model. The pre-processing is done before we start training the model. Words are encoded into a continuous representation. The vocabulary is built from training data. We apply preprocessing steps since our data are noisy and come from Twitter. There are many spelling mistakes, irregular word usages and repeated characters. To create the shared vocabulary, we concatenate \textit{$L_1$} and \textit{$L_2$} word vectors. For pre-processing, we propose the following steps: 
\begin{enumerate}
\item \textbf{Token replacement: } Replace user hashtags (\#user) and mentions (@user) with ``USR", and URL (https://domain.com) with ``URL".
\item \textbf{Token normalization: } Concatenate \textit{$L_1$} and \textit{$L_2$} FastText word vector vocabulary. Normalize OOV words by sequentially checking if the word exists after applying the following changes:
	\begin{enumerate}
    \item Capitalize the first character
    \item Lowercase the word
    \item Do (b) and remove repeated characters, such as \textit{``hellooooo"} to \textit{``hello"} or \textit{``lolololol"} to \textit{``lol"}
    \item Do steps (a) and (c) together
    \end{enumerate}
\end{enumerate}
As shown in Table~\ref{data-statistics-eng-spa-named-entities},  pre-processing and transfer learning are effective in handling OOV words and the pre-processing strategies dramatically decrease the number of unknown words. We concatenate all possible characters for English and Spanish, including numbers and special characters. English and Spanish have most characters in common, but, Spanish has some additional unique characters. All cases are kept as they are.

\subsection{Experimental Setup}
\subsubsection{Dataset}
In this experiment, we use English-Spanish NER tweets data from the CALCS 2018 shared task~\cite{aguilar2018named}. There are nine different named-entity labels, which all use the IOB format (Inside, Outside, Beginning), where every token is labeled with a \textnormal{\tt B-label} in the beginning, followed by an \textnormal{\tt I-label} if it is inside a named entity, or \textnormal{\tt O} otherwise. For example ``Kendrick Lamar'' is represented as \textnormal{\tt B-PER I-PER}. Table \ref{data-statistics-eng-spa} and Table \ref{data-statistics-eng-spa-named-entities} show the statistics of the dataset.

\begin{table*}[!ht]
\centering
\caption{OOV word rates on ENG-SPA dataset before and after pre-processing.}
\label{eng-spa-oov-statistics}
\begin{tabular}{@{}lccccc@{}}
\toprule
\multicolumn{1}{c}{\multirow{2}{*}{\textbf{}}} & \multicolumn{2}{c}{\textbf{Train}}  & \multicolumn{2}{c}{\textbf{Dev}} & \multicolumn{1}{c}{\textbf{Test}} \\ \cmidrule{2-6} 
\multicolumn{1}{c}{} & \multicolumn{1}{c}{All} & Entity & \multicolumn{1}{c}{All} & Entity                           & \multicolumn{1}{c}{All}                              \\ \midrule
Corpus & \multicolumn{1}{c}{-} & - & \multicolumn{1}{c}{18.91\%} & 31.84\% & \multicolumn{1}{c}{49.39\%} \\ 
FastText (eng)~\cite{mikolov2018advances} & \multicolumn{1}{c}{62.62\%} & 16.76\% & \multicolumn{1}{c}{19.12\%} & 3.91\% & \multicolumn{1}{c}{54.59\%} \\ 
+ FastText (spa)~\cite{grave2018learning} & \multicolumn{1}{c}{49.76\%} & 12.38\% & \multicolumn{1}{c}{11.98\%} & 3.91\% & \multicolumn{1}{c}{39.45\%} \\ 
+ token replacement & \multicolumn{1}{c}{12.43\%} & 12.35\% & \multicolumn{1}{c}{7.18\%} & 3.91\% & \multicolumn{1}{c}{9.60\%}  \\ 
\textbf{+ token normalization} & \multicolumn{1}{c}{\textbf{7.94\%}} & \textbf{8.38\%} & \multicolumn{1}{c}{\textbf{5.01\%}} & \textbf{1.67\%} & \multicolumn{1}{c}{\textbf{6.08\%}}  \\ \bottomrule
\end{tabular}
\end{table*}

\begin{table}[!htb]
\centering
\caption{Data statistics for ENG-SPA tweets.}
\label{data-statistics-eng-spa}
\begin{tabular}{@{}llll@{}}
\toprule
& \multicolumn{1}{c}{\textbf{Train}} & \multicolumn{1}{c}{\textbf{Dev}} & \multicolumn{1}{c}{\textbf{Test}} \\ \midrule
\# Words & \multicolumn{1}{c}{616,069} & 9,583 & \multicolumn{1}{c}{183,011} \\ \bottomrule
\end{tabular}
\end{table}

\begin{table}[!htb]
\centering
\caption{Entity statistics for ENG-SPA tweets}
\label{data-statistics-eng-spa-named-entities}
\begin{tabular}{@{}rcc@{}}
\toprule
\multicolumn{1}{r}{\textbf{Entities}} & \multicolumn{1}{c}{\textbf{Train}} & \multicolumn{1}{c}{\textbf{Dev}}  \\ \midrule
\# Person & \multicolumn{1}{c}{4701} & 75 \\ 
\# Location & \multicolumn{1}{c}{2810} & 10 \\ 
\# Product & \multicolumn{1}{c}{1369} & 16 \\ 
\# Title & \multicolumn{1}{c}{824} & 22 \\ 
\# Organization & \multicolumn{1}{c}{811} & 9 \\ 
\# Group & \multicolumn{1}{c}{718} & 4 \\ 
\# Time & \multicolumn{1}{c}{577} & 6 \\ 
\# Event & \multicolumn{1}{c}{232} & 4 \\ 
\# Other & \multicolumn{1}{c}{324} & 6 \\ \bottomrule
\end{tabular}
\end{table}

``Person'', ``Location'', and ``Product'' are the most frequent entities in the dataset, and the least common are the ``Time", ``Event", and ``Other'' categories. The ``Other'' category is the least trivial among them because it is not well clustered like the others. 
We found an issue during the prediction where some words are labeled with \textnormal{\tt O}, in between \textnormal{\tt B-label} and \textnormal{\tt I-label} tags. Our solution is to insert an \textnormal{\tt I-label} tag if the tag is surrounded by a \textnormal{\tt B-label} and \textnormal{\tt I-label} tags with the same entity category. Another problem we found is that many \textnormal{\tt I-label} tags are paired with a \textnormal{\tt B-label} in different categories. So, we replace the \textnormal{\tt B-label} category tag with the corresponding \textnormal{\tt I-label} category tag. This step improves the results of the prediction on the development set.

\subsubsection{Model Training}
We train our LSTM models with a hidden size of 200 and use a batch size equal to 64. The sentences are sorted by length in descending order. Our embedding size is 300 for words and 150 for characters. Dropout \cite{srivastava2014dropout} of 0.4 is applied to all LSTMs. The Adam optimizer is chosen to have an initial learning rate of 0.01. We apply time-based decay of a $\sqrt{2}$ decay rate and stop after two consecutive epochs without improvement. We tune our models with the development set and evaluate our best model with the test set using the harmonic mean F1-score metric with the script provided by \citet{aguilar2018named}.

\subsection{Results and Discussion}
Table \ref{results-eng-spa} gives the results for ENG-SPA tweets, showing that adding pre-trained word vectors and character-level features improves the performance. Interestingly, our initial attempts at adding character-level features did not improve the overall performance, until we applied dropout to the Char-RNN. The performance of the model improves significantly after transfer learning with FastText word vectors, while it also reduces the number of OOV words in the development and test set. We use a subword representation from Spanish FastText \cite{grave2018learning}. However, it does not improve the results since the OOV words consist of many special characters, for example, \textit{``/IAtrevido/Provocativo"}, and \textit{``Twets/wek"}, and possibly create noisy vectors, while most are also not entity words.

\begin{table*}[!htb]
\centering
\caption{Results on ENG-SPA Dataset.}
\label{results-eng-spa}
\begin{tabular}{@{}llcc@{}}
\toprule
\multicolumn{1}{c}{\textbf{Model}}                    & \multicolumn{1}{c}{\textbf{Features}} & \textbf{\begin{tabular}[c]{@{}c@{}}F1\\ Dev\end{tabular}} & \multicolumn{1}{c}{\textbf{\begin{tabular}[c]{@{}c@{}}F1\\ Test\end{tabular}}} \\ \midrule
BiLSTM & \multicolumn{1}{c}{Word} & - & \multicolumn{1}{c}{53.28\%} \\ 
BiLSTM & \multicolumn{1}{c}{Word + Char-RNN} & 46.96\% & \multicolumn{1}{c}{53.48\%} \\
BiLSTM & \multicolumn{1}{c}{FastText (eng)} & 57.72\% & \multicolumn{1}{c}{59.91\%} \\ 
BiLSTM & \multicolumn{1}{c}{FastText (eng-spa)} & 57.42\% & \multicolumn{1}{c}{60.24\%} \\ 
BiLSTM & \multicolumn{1}{c}{+ Char-RNN} & 65.22\%  & \multicolumn{1}{c}{61.96\%}                                                  \\ 
+ post                                         & \multicolumn{1}{c}{}                                      & \textbf{65.39\%}                                        & \multicolumn{1}{c}{\textbf{62.76\%}}                                      \\ \bottomrule
\end{tabular}
\end{table*}

\section{Meta-Embeddings}

\subsection{Simple Meta-Embeddings}
\begin{figure}[!htb]
  \centering
  \includegraphics[width=0.9\linewidth]{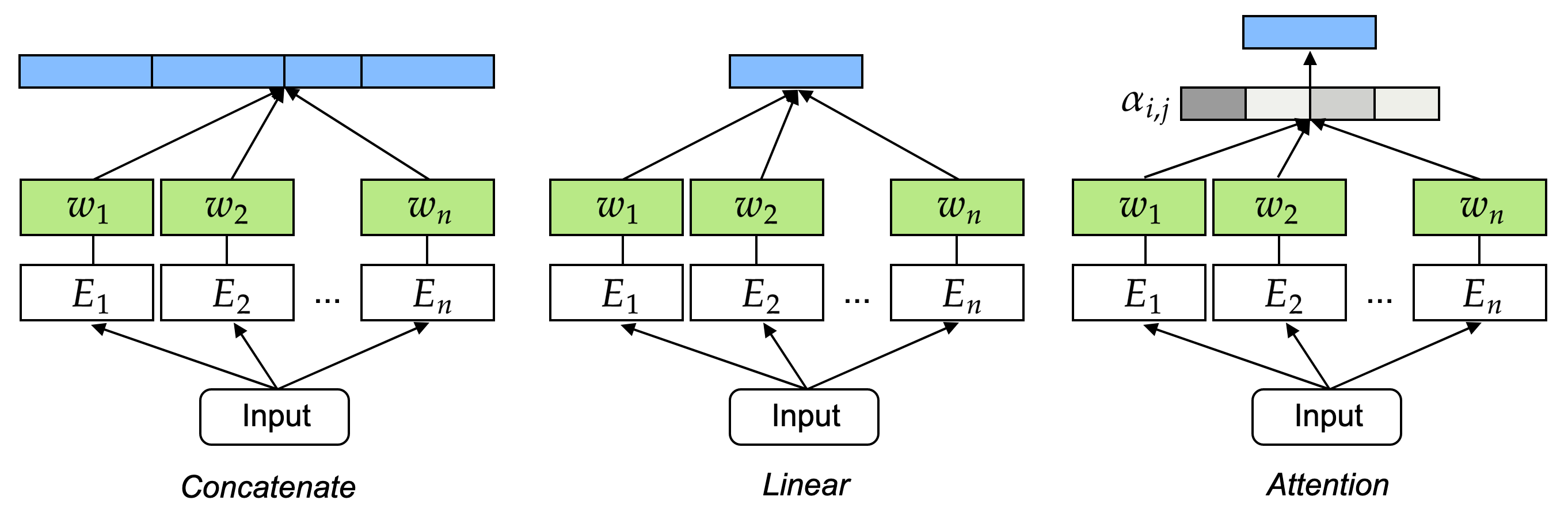}
  \caption{Meta-embeddings architecture.}
  \label{fig:meta_embedding_architectures}
\end{figure}

Creating new embeddings by combining existing embeddings has been shown to be an important research direction in NLP. \citet{bansal2014tailoring} show that an ensemble of word representations outperforms single word embeddings, which implies the complementarity of different word embeddings and suggests that adding more embeddings could be a useful technique to build a better word context. Meta-embeddings aim to learn how to effectively combine pre-trained word embeddings in supervised training into a single dense representation \cite{yin2016learning,muromagi2017linear,bollegala2018think,coates2018frustratingly,kiela2018dynamic}. This method is known to be effective to overcome domain and modality limitations. Given a set of $n$ embeddings $\mathbf{E} = \{ E_1, E_2, ..., E_k, ..., E_{n-1}, E_n \}$, where each has a different embedding size $d_k$, there are many ways to form meta-embeddings. Figure~\ref{fig:meta_embedding_architectures} shows the three main architectures of meta-embeddings: concatenated, linear, and attention-based. These architectures are introduced in the following.

\paragraph{Concatenation}
A naive method to combine multiple sources of embeddings is concatenation. It is a very simple method, but it has shown good performance in semantic similarity measurement and word analogy detection, as shown by~\citet{bollegala2018think}. Here, we can say that it creates a higher embedding dimension and the computation is more expensive according to the number of embeddings combined. The final embeddings is
\begin{align}
    \mathbf{w}_{meta} = [\mathbf{w}_1,\mathbf{w}_2,...,\mathbf{w}_n].
\end{align}

\paragraph{Linear}
Instead of concatenating the embeddings, we can simply sum all of them with an equal weight by assuming the embedding sizes are the same. If the embedding sizes are different, we need to project the embeddings beforehand. The final embeddings is
\begin{align}
    \mathbf{w}_{meta} = \sum_{i=0}^n{\phi(\mathbf{w}_i)}.
\end{align}

\paragraph{Attention}
We can learn a simple attention mechanism to weight embeddings from several sources, and apply the weights to embeddings. The final embeddings is
\begin{align}
    \mathbf{w}_{meta} = \sum_{i=0}^n{\alpha_i \phi(\mathbf{w}_i)}.
\end{align}

In the context of code-switching,
~\citet{wang2018code} concatenate the weighted embeddings. If a word embedding is available in one language, but not in the other, the second language embeddings are initialized as a zero-vector.

\subsection{Multilingual Meta-Embeddings}
Word embedding pre-training is a well-known method to transfer knowledge from previous tasks to a target task that has fewer high-quality training data. Word embeddings are commonly used as features in supervised learning problems. We extend the idea of meta-embeddings to solve a multilingual task, and take a new perspective by combining embeddings from different languages. A word may appear in more than one language, and by taking embeddings from different language sources, we conjecture commonalities in semantics among languages.
\begin{figure}[!htb]
  \centering
  \includegraphics[width=0.6\linewidth]{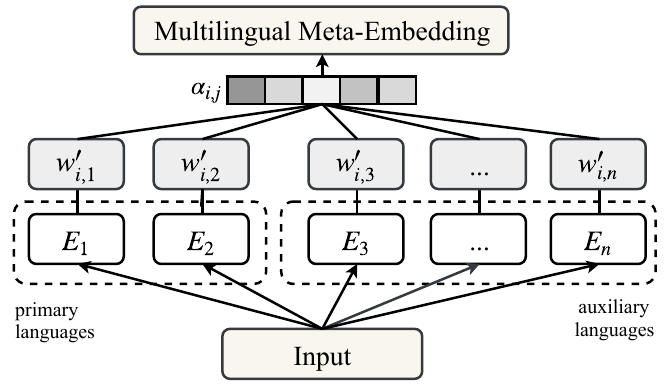}
  \caption{Multilingual meta-embeddings architecture.}
  \label{fig:mme}
\end{figure}
We propose to generate a single word representation by extracting information from different pre-trained embeddings. We apply a non-linear gating function to attend to pre-trained embeddings and explore the possibility of utilizing more languages for generating multilingual meta-embeddings. 
We define a sentence that consists of $m$ words $\{\mathbf{x}_j\}_{j=1}^m$, and $\{\mathbf{w}_{i,j}\}_{j=1}^n$ word vectors from $n$ pre-trained word embeddings. We generate a multilingual vector representation for each word by taking a weighted sum of monolingual embeddings. Each embedding $\textbf{w}_{i,j}$ is projected with a fully connected layer with a non-linear scoring function $\phi$ (e.g., tanh) into a $d$-dimensional vector and an attention mechanism to calculate the attention weight $\alpha_{i,j} \in \mathbb{R}^{d}$:
\begin{align}
\mathbf{w}_i^{MME} &= \sum_{j=1}^n {\alpha_{i,j} 
\mathbf{w'}_{i,j}},\\
\alpha_{i,j} &= \frac{e^{\phi(\mathbf{w'}_{i,j})}}{\sum_{j=1}^n  e^{\phi(\mathbf{w'}_{i,j}))}}.
\end{align}

\subsection{Hierarchical Meta-Embeddings}
Previous works have mostly focused on applying pre-trained word embeddings from each language in order to represent noisy mixed-language text, and combining them with character-level representations~\cite{trivedi2018iit,wang2018code,winata2018bilingual}. However, despite the effectiveness of such word-level approaches, they neglect the importance of subword-level characteristics shared across different languages. Such information is often hard to capture with word embeddings or randomly initialized character-level embeddings. Naturally, we can turn towards subword-level embeddings such as FastText~\cite{grave2018learning} to help in this task, which would allow us to leverage the morphological structures shared across different languages. Despite their likely usefulness, little attention has been focused around using subword-level features in this task. 
This is partly because of the non-triviality of combining language embeddings in the subword space, which arises from the distinct segmentation into subwords for different languages. 
The lack of attention leads us to explore the literature of meta-embeddings~\cite{yin2016learning,muromagi2017linear,bollegala2018think,coates2018frustratingly,kiela2018dynamic,winata2019learning}, a method to learn how to combine embeddings.
\begin{figure}[!htb]
  \centering
  \includegraphics[width=0.65\linewidth]{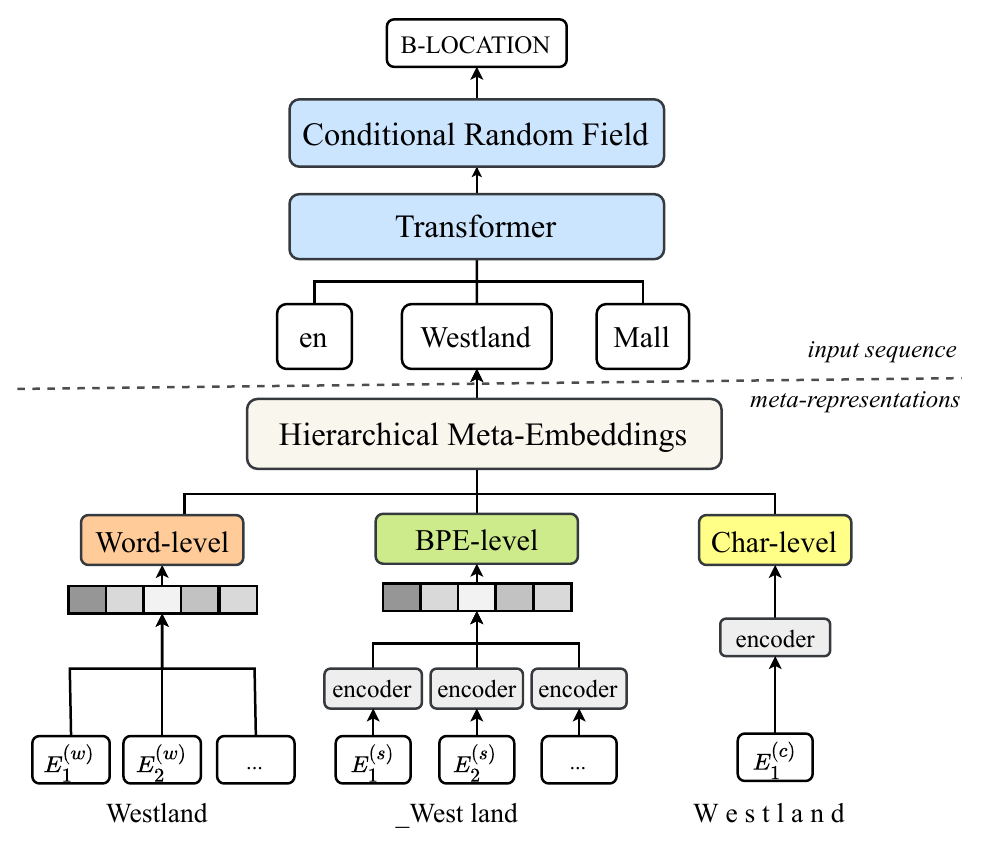}
  \caption{Hierarchical meta-embeddings architecture.}
  \label{fig:hme}
\end{figure}

We propose a method to combine word, subword, and character representations to create a mixture of embeddings~\cite{winata2019hierarchical}. We generate multilingual meta-embeddings of words and subwords, and then concatenate them with character-level embeddings to generate final word representations, as shown in Figure~\ref{fig:hme}. Let $\mathbf{w}$ be a sequence of words with $n$ elements, where $\mathbf{w} = [w_1,\dots,w_n]$. Each word can be tokenized into a list of subwords $\mathbf{s} = [s_1,\dots,s_m]$ and a list of characters $\mathbf{c} = [c_1,\dots,c_p]$. The list of subwords $\mathbf{s}$ is generated using a function $f$; $\mathbf{s} = f(\mathbf{w})$. Function $f$ maps a word into a sequence of subwords. Further, let $E^{(w)}$, $E^{(s)}$, and $E^{(c)}$ be a set of word, subword, and character embedding lookup tables. Each set consists of different monolingual embeddings, and each element is transformed into an embedding vector in $\mathbb{R}^d$. We denote subscripts $_{\{i,j\}}$ as element and embedding language indices, and superscripts $^{({w,s,c})}$ as words, subwords, and characters.

\paragraph{Mapping Subwords and Characters to Word-Level Representations} 
We propose to map subwords into word representations and choose BPEs~\cite{sennrich2016neural} since they have a compact vocabulary. First, we apply $f$ to segment words into sets of subwords, and then extract the pre-trained subword embedding vectors $\mathbf{x}^{(s)}_{i,j} \in \mathbb{R}^d$ for language $j$. Since each language has a different $f$, we replace the projection matrix with the transformer~\cite{vaswani2017attention} to learn and combine important subwords into a single vector representation. Then, we create $\mathbf{u}_i^{(s)} \in \mathbb{R}^{d'}$ which represents the subword-level MME by taking the weighted sum of $\mathbf{x}'^{(s)}_{i,j} \in \mathbb{R}^{d'}$:
\begin{align}
\mathbf{x}'^{(s)}_{i,j} = \text{Encoder}(\mathbf{x}^{(s)}_{i,j}),\\
\mathbf{u}_{i}^{(s)} = \sum_{j=1}^n{\alpha_{i,j} \mathbf{x}'^{(s)}_{i,j}}.
\end{align}

To combine character-level representations, we apply an encoder to each character:
\begin{align}
    \mathbf{u}_i^{(c)} = \textnormal{Encoder}(\mathbf{x}_{i}) \in \mathbb{R}^{d'}.
\end{align}

We combine the word-level, subword-level, and character-level representations by concatenation $\mathbf{u}^{HME}_i = (\mathbf{u}^{(w)}_i, \mathbf{u}^{(s)}_i, \mathbf{u}^{(c)}_i)$, where $\mathbf{u}^{(w)}_i \in \mathbb{R}^{d'}$ and $\mathbf{u}^{(s)}_i \in \mathbb{R}^{d'}$ are the word-level MME and BPE-level MME, and $\mathbf{u}^{(c)}_i$ is a character embedding. We randomly initialize the character embedding and keep it trainable. We fix all subword and word pre-trained embeddings during the training.

\subsection{Experimental Setup}

\subsubsection{Datasets}
Similar to the previous chapter, we use English-Spanish NER tweets data from the CALCS 2018 shared task~\cite{aguilar2018named}. It is also the same dataset as used in the previous section.

\subsubsection{Model Training}
Our model is trained using the Noam optimizer with a dropout of 0.1 for the multilingual setting and 0.3 for the cross-lingual setting. Our model contains four layers of transformer blocks with a hidden size of 200 and four heads. We start the training with a learning rate of 0.1, and replace user hashtags (\#user) and mentions (@user) with \texttt{<USR>}, and URL (https://domain.com) with \texttt{<URL>}, similarly to \citet{winata2018bilingual}. 
\subsubsection{Pre-trained Word Embeddings}
We use FastText word embeddings trained from Common Crawl and Wikipedia \cite{grave2018learning} for English (\textit{en}) and Spanish (\textit{es}), as well as \textbf{four Romance languages}, Catalan (\textit{ca}), Portuguese (\textit{pt}), French (\textit{fr}), and Italian (\textit{it}), and \textbf{a Germanic language}, German (\textit{de}). \textbf{Five Celtic languages} are included as the distant language group: Breton (\textit{br}), Welsh (\textit{cy}), Irish (\textit{ga}), Scottish Gaelic (\textit{gd}), and Manx (\textit{gv}). We also add English Twitter GloVe word embeddings \cite{pennington2014glove} and BPE-based subword embeddings from \citet{heinzerling2018bpemb}. We generate the vector representation on all unknown words using FastText. We train our model in two different settings: \textbf{(1) the multilingual setting}, where we combine the main languages (\textit{en-es}) with the Romance languages and Germanic language, and \textbf{(2) the cross-lingual setting}, where we use the Romance and Germanic languages without the main languages. Our model contains four layers of transformer encoders with a hidden size of 200, four heads, and a dropout of 0.1. We use the Adam optimizer and start the training with a learning rate of 0.1 and an early stop of 15 iterations. We replace user hashtags and mentions with \texttt{<USR>}, emoji with \texttt{<EMOJI>}, and URL with \texttt{<URL>}. We evaluate our model using the \textit{absolute F1 score} metric.

\subsubsection{Baselines}
We evaluate our model using flat combinations of word-embeddings by concatenating or summing all word-level embeddings.
\paragraph{CONCAT} We concatenate word embeddings by merging the dimensions of word representations. This method combines embeddings into a high-dimensional input that may cause inefficient computation.

\paragraph{LINEAR} We sum all word embeddings into a single word vector with equal weight. This method combines embeddings without considering the importance of each.

\paragraph{Random embeddings} We use randomly initialized word embeddings and keep them trainable to calculate the lower-bound performance.

\paragraph{Aligned embeddings} We align embeddings using CSLS. We run MUSE using the code from~\citet{lample2018word}.~\footnote{https://github.com/facebookresearch/MUSE}

\begin{table}[!th]
\centering
\caption{Meta-embeddings results on CALCS 2018 English-Spanish NER test set.}
\resizebox{\textwidth}{!}{
\begin{tabular}{llllllll}
\toprule
\multicolumn{1}{c}{\multirow{4}{*}{Model}} & \multicolumn{4}{c}{Multilingual embeddings} & \multicolumn{3}{c}{Cross-lingual  embeddings} \\ \cmidrule{2-8} 
\multicolumn{1}{c}{} & \multicolumn{1}{c}{\begin{tabular}[c]{@{}c@{}}main\\ languages\end{tabular}} & \multicolumn{2}{c}{\begin{tabular}[c]{@{}c@{}}+ closely-related\\ languages\end{tabular}} & \multicolumn{1}{c}{\begin{tabular}[c]{@{}c@{}}+ distant\\ languages\end{tabular}} & \multicolumn{2}{c}{\begin{tabular}[c]{@{}c@{}}closely-related\\ languages\end{tabular}} & \multicolumn{1}{c}{\begin{tabular}[c]{@{}c@{}}distant\\ languages\end{tabular}} \\ \cmidrule{2-8} 
\multicolumn{1}{c}{} & \multicolumn{1}{c}{en-es} & \multicolumn{1}{c}{ca-pt} & \multicolumn{1}{c}{ca-pt-de-fr-it} & \multicolumn{1}{c}{br-cy-ga-gd-gv} & \multicolumn{1}{c}{ca-pt} & \multicolumn{1}{c}{ca-pt-de-fr-it} & br-cy-ga-gd-gv \\ \midrule
\multicolumn{8}{l}{Flat word-level embeddings} \\ \midrule
\multicolumn{1}{l}{CONCAT} &  \multicolumn{1}{c}{65.3 \small{$\pm$ 0.38}} & \multicolumn{1}{c}{64.99 \small{$\pm$ 1.06}} & \multicolumn{1}{c}{65.91 \small{$\pm$ 1.16}} & \multicolumn{1}{c}{65.79 \small{$\pm$ 1.36}} & \multicolumn{1}{c}{58.28 \small{$\pm$ 2.66}} & \multicolumn{1}{c}{64.02 \small{$\pm$ 0.26}} & \multicolumn{1}{c}{50.77 \small{$\pm$ 1.55}} \\
\multicolumn{1}{l}{LINEAR} & \multicolumn{1}{c}{64.61 \small{$\pm$ 0.77}} & \multicolumn{1}{c}{65.33 \small{$\pm$ 0.87}} & \multicolumn{1}{c}{65.63 \small{$\pm$ 0.92}} & \multicolumn{1}{c}{64.95 \small{$\pm$ 0.77}} & \multicolumn{1}{c}{60.72 \small{$\pm$ 0.84}} & \multicolumn{1}{c}{62.37 \small{$\pm$ 1.01}} & \multicolumn{1}{c}{53.28 \small{$\pm$ 0.41}}\\ \midrule
\multicolumn{8}{l}{Multilingual Meta-Embeddings (MME)} \\ \midrule
\multicolumn{1}{l}{Word} &
\multicolumn{1}{c}{65.43 \small{$\pm$ 0.67}} & \multicolumn{1}{c}{66.63 \small{$\pm$ 0.94}} & 
\multicolumn{1}{c}{66.8 \small{$\pm$ 0.43}} & \multicolumn{1}{c}{66.56 \small{$\pm$ 0.4}} & \multicolumn{1}{c}{61.75 \small{$\pm$ 0.56}} & \multicolumn{1}{c}{63.23 \small{$\pm$ 0.29}} & \multicolumn{1}{c}{53.43 \small{$\pm$ 0.37}} \\ \midrule  
\multicolumn{8}{l}{Hierarchical Meta-Embeddings (HME)} \\ \midrule
\multicolumn{1}{l}{+ BPE} & \multicolumn{1}{c}{65.90 \small{$\pm$ 0.72}} & \multicolumn{1}{c}{67.31 \small{$\pm$ 0.34}} & \multicolumn{1}{c}{67.26 \small{$\pm$ 0.54}} & \multicolumn{1}{c}{66.88 \small{$\pm$ 0.37}} & \multicolumn{1}{c}{63.44 \small{$\pm$ 0.33}} & \multicolumn{1}{c}{63.78 \small{$\pm$ 0.62}} & \multicolumn{1}{c}{60.19 \small{$\pm$ 0.63}}\\
\multicolumn{1}{l}{+ Char} & \multicolumn{1}{c}{65.88 \small{$\pm$ 1.02}} & \multicolumn{1}{c}{67.38 \small{$\pm$ 0.84}} & \multicolumn{1}{c}{65.94 \small{$\pm$ 0.47}} & 
\multicolumn{1}{c}{66.10 \small{$\pm$ 0.33}} & 
\multicolumn{1}{c}{61.97 \small{$\pm$ 0.60}} & 
\multicolumn{1}{c}{63.06 \small{$\pm$ 0.69}} & 
\multicolumn{1}{c}{57.50 \small{$\pm$ 0.56}}\\
\multicolumn{1}{l}{+ BPE + Char} & \multicolumn{1}{c}{66.55 \small{$\pm$ 0.72}} & \multicolumn{1}{c}{\textbf{67.80 \small{$\pm$ 0.31}}} & \multicolumn{1}{c}{67.07 \small{$\pm$ 0.49}} & \multicolumn{1}{c}{67.02 \small{$\pm$ 0.16}} & \multicolumn{1}{c}{63.9 \small{$\pm$ 0.22}} & \multicolumn{1}{c}{ \textbf{64.52 \small{$\pm$ 0.35 }}} & \multicolumn{1}{c}{60.88 \small{$\pm$ 0.84}} \\ \bottomrule
\end{tabular}
}
\label{tab:results_calcs}
\end{table}
\begin{table}[!ht]
\centering
\caption{Overall results on CALCS 2018 English-Spanish NER test set.}
\resizebox{0.5\textwidth}{!}{
\begin{tabular}{lc}
\toprule
\multicolumn{1}{c}{Model} & \multicolumn{1}{c}{F1} \\ \midrule
Random & 46.68 \small{$\pm$ 0.79} \\ \midrule
\multicolumn{1}{l}{\citet{wang2018code}} & 62.39 \\
\multicolumn{1}{l}{\citet{wang2018code} (Ensemble)} & 62.67 \\
\multicolumn{1}{l}{\citet{trivedi2018iit} } & 61.89 \\
\multicolumn{1}{l}{\citet{trivedi2018iit} (Ensemble)} & 63.76 \\
\multicolumn{1}{l}{Bilingual Char RNN~\cite{winata2018bilingual}} & 62.76 \\ \midrule
% \multicolumn{1}{l|}{\citet{winata2019learning} MME (Ensemble)} & 68.34 \\ \hline
% \multicolumn{2}{l}{Flat word-level embeddings} \\ 
% CONCAT & 65.91 \small{$\pm$ 1.16}\\ 
% LINEAR & 65.63 \small{$\pm$ 0.92}\\ \midrule
\textbf{Bilingual embeddings}\\
MUSE (es $\rightarrow$ en)~\cite{lample2017unsupervised} & 60.89 \small{$\pm$ 0.37} \\
MUSE (en $\rightarrow$ es)~\cite{lample2017unsupervised} & 61.49 \small{$\pm$ 0.62} \\ 
biCVM~\cite{hermann-blunsom-2014-multilingual} & 51.60 \\
biSkip~\cite{luong-etal-2015-bilingual} & 52.98 \\
GCM~\cite{pratapa-etal-2018-word} & 53.57 \\ \midrule
\textbf{Multilingual embeddings} & \\
mBERT~\cite{khanuja-etal-2020-gluecos} & 59.69 \\
mBERT (modified)~\cite{khanuja-etal-2020-gluecos} & 61.77 \\
MME & 66.63 \small{$\pm$ 0.94} \\
HME & 67.80 \small{$\pm$ 0.31} \\ 
HME (Ensemble)$^\dagger$ & \textbf{69.17} \\ \midrule
\textbf{Cross-lingual embeddings} & \\
MME & 63.23 \small{$\pm$ 0.29} \\
HME & 64.52 \small{$\pm$ 0.35} \\ 
HME (Ensemble)$^\dagger$ & \textbf{65.99} \\ \bottomrule
\end{tabular}
}
% \caption{Comparison to existing works. $^\dagger$ Ensemble five different models. Our approaches are best performing ones from Table~\ref{tab:res}.}
\label{tab:related_work_calcs}
\end{table}

\subsection{Results and Discussion}
In Table~\ref{tab:results_calcs}, we report the average and std. F1-score results are obtained on the CALCS 2018 English-Spanish dataset, which is run five times. We compare our results on flat word-level embeddings, multilingual meta-embeddings, and hierarchical meta-embeddings. In general, we can see that word-level meta-embeddings even without subword or character-level information, consistently perform better than the flat baselines (e.g., CONCAT and LINEAR) in all settings. This is mainly because of the attention layer, which does not require additional parameters. Furthermore, comparing our approaches to previous state-of-the-art models, we can clearly see that ours all significantly outperform the others.

From Table~\ref{tab:results_calcs}, in the multilingual setting, in which the model is trained with the main languages, it is evident that adding both closely-related and distant language embeddings improves the performance. This shows us that our model is able to leverage the lexical similarity between the languages. This is more distinctly shown in the cross-lingual setting, as using distant languages performs significantly less well than using closely-related ones (e.g., \textit{ca-pt}). Interestingly, for distant languages, when adding subwords, we can still see a drastic performance increase. We hypothesize that even though the characters are mostly different, the lexical structure is similar to that of our main languages. On the other hand, adding subword inputs to the model is consistently better than adding characters. This is due to the transfer of the information from the pre-trained subword embeddings. As shown in Table~\ref{tab:results_calcs}, subword embeddings are more effective for the distant languages (Celtic languages) than the closely-related languages such as Catalan or Portuguese. 

Table~\ref{tab:related_work_calcs} shows a comparison of our proposed method with existing work. For the ensemble model, we run a majority voting scheme from five different models. Interestingly, multilingual language models such as mBERT underperform many of the baselines. It is clear that mBERT is only trained on monolingual datasets without any supervision on parallel datasets. Thus, the performance of the model on code-switching data is very poor.
\begin{figure}[!ht]
  \centering
  \includegraphics[width=\linewidth]{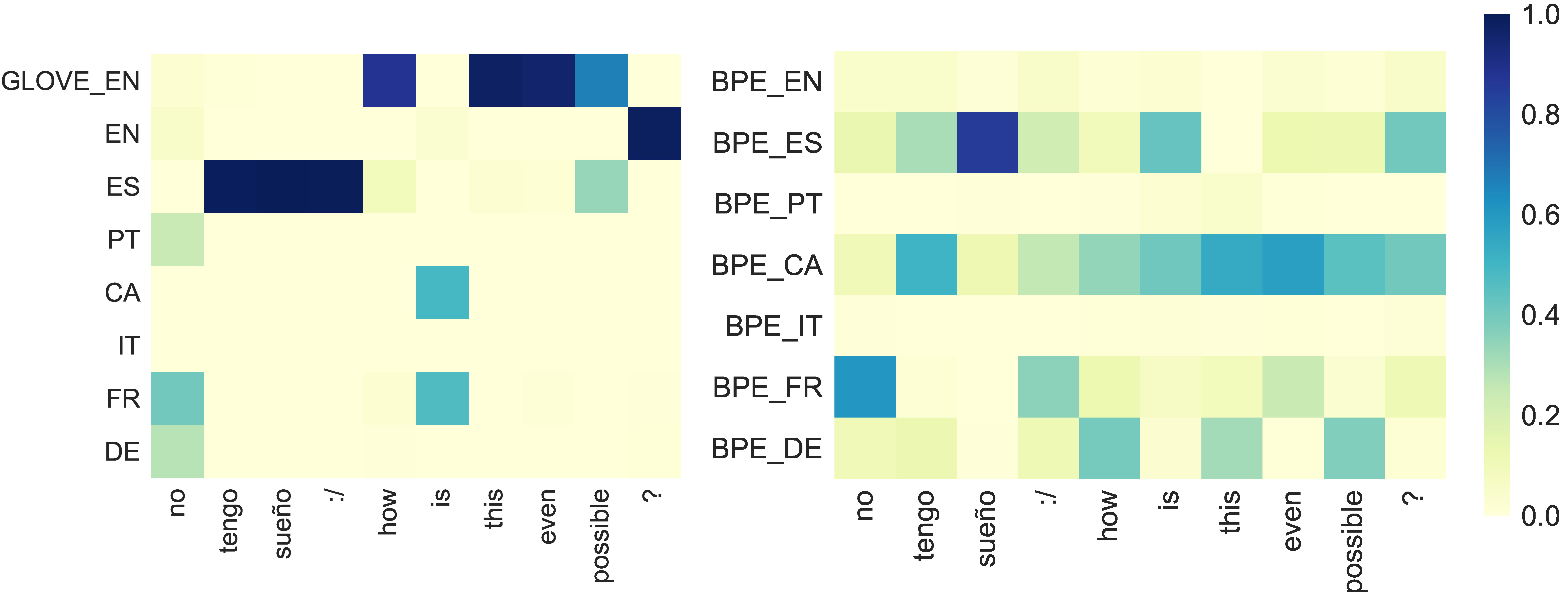}
  \caption{Heatmap of attention over languages from a validation sample of the word-level MME and BPE-level MME extracted from the attention weights of the multilingual model (\textit{en-es-ca-pt-de-fr-it}).}
  \label{fig:attention_heatmap}
\end{figure}
\begin{figure}[!ht]
  \centering
  \includegraphics[width=0.5\linewidth]{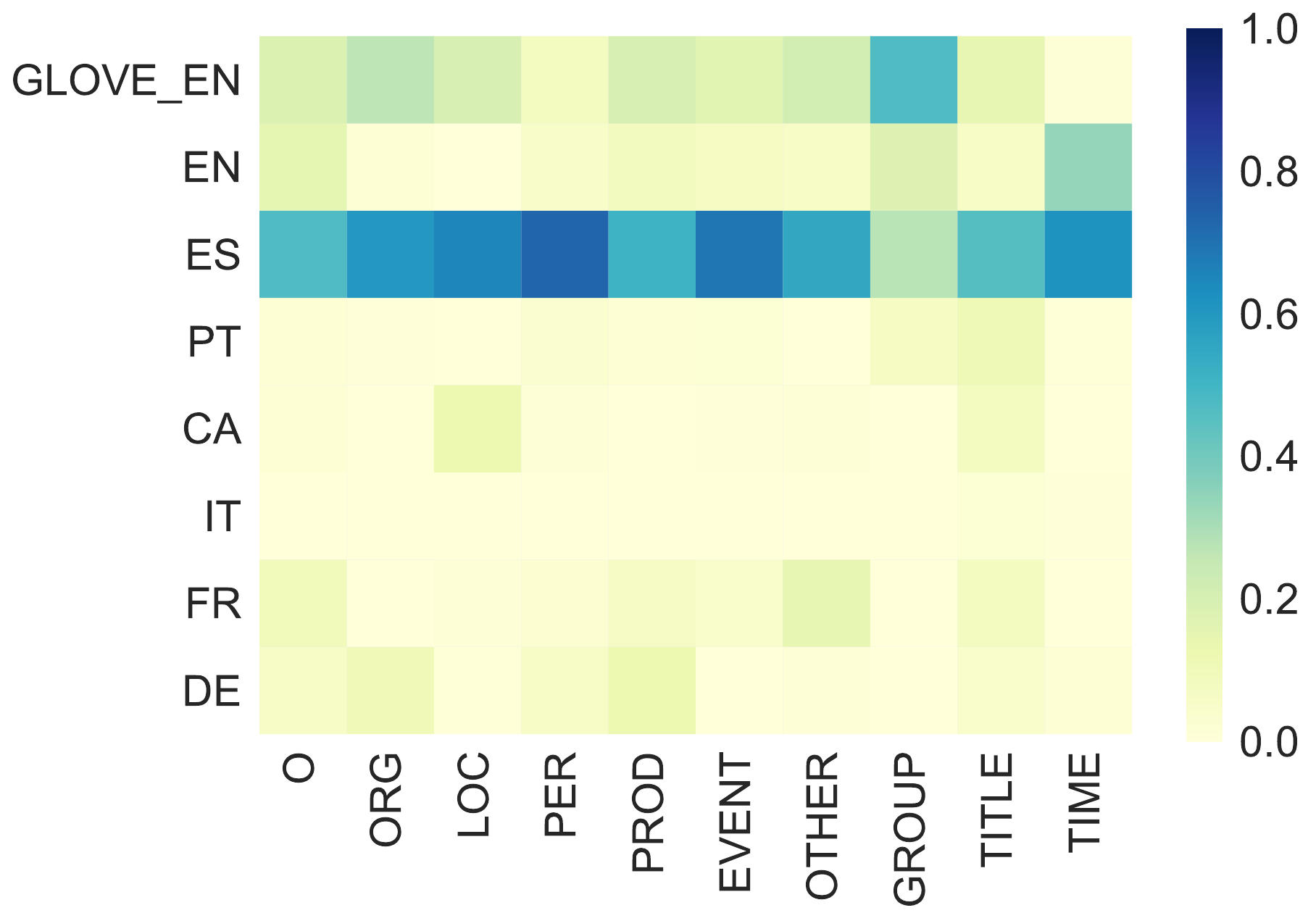}
  \caption{The average of attention weights for word embeddings versus NER tags from the validation set.}
  \label{fig:attention_lang_class}
\end{figure}

\subsubsection{Visualization}
Moreover, we visualize the attention weights of the model at the word and subword-level to interpret the model dynamics. As can be seen from the left-hand side of Figure~\ref{fig:attention_heatmap}, at the word-level, the model mostly chooses the correct language embedding for each word, but also combines different languages. Without any language identifiers, it is impressive to see that our model learns to attend to the right languages. The right-hand side of Figure~\ref{fig:attention_heatmap}, which shows attention weight distributions for the subword-level, demonstrates interesting behaviors of the model, in which, for most English subwords, it leverages \textit{ca}, \textit{fr}, and \textit{de} embeddings. We hypothesize that this is because the dataset is mainly constructed with Spanish words, which can also be verified from Figure~\ref{fig:attention_lang_class}, in which most NER tags are classified as \textit{es}, and only some as \textit{en}, such as Group and Time. 

\section{Effectiveness and Efficiency of Multilingual Models}

In this section, we provide an analysis on the power of existing pre-trained multilingual models to understand their capability and adaptability in the code-switching setting~\cite{winata2021multilingual}. Here, we would like to answer the question which models are effective in representing code-switching text, and why?. In order to find the answer, we conduct a comparative study on various existing multilingual models with our proposed meta-embeddings by evaluating the performance, activation memory, and number of parameters. Moreover, we further analyze the memory footprint required by each model over different sequence lengths in a GPU. Thus, we are able to understand which model to choose in a practical scenario.

\subsection{Experimental Setup}
\subsubsection{Datasets}
We evaluate our models on five downstream tasks and three language pairs in the LinCE Benchmark~\cite{aguilar2020lince}. We choose three NER tasks, Hindi-English (HIN-ENG)~\cite{singh2018language}, Spanish-English (SPA-ENG)~\cite{aguilar2018named} and Modern Standard Arabic (MSA-EA)~\cite{aguilar2018named}, and two POS tagging tasks, Hindi-English (HIN-ENG)~\cite{singh2018twitter} and Spanish-English (SPA-ENG)~\cite{soto2017crowdsourcing}. For the Hindi-English datasets, we apply Roman-to-Devanagari transliteration. We show the number of tokens of each language in Table~\ref{dataset}, where we classify the language with more tokens as ML and the other as EL, where ML is the matrix, or the primary, language, and EL is the embedded, or the secondary, language. We replace some words with special tokens, such as user hashtags and mentions with \texttt{<USR>}, emoji with \texttt{<EMOJI>}, and URL with \texttt{<URL>}, for models that use word-embeddings, similarly to~\citet{winata2019learning}. We evaluate our models using the micro-F1 score for NER and accuracy for POS tagging, following~\citet{aguilar2020lince}.

\begin{table}[!ht]
\centering
\caption{Dataset statistics are taken from~\citet{aguilar2020lince}. We define \textit{$L_1$} and  \textit{$L_2$} as the languages found in the dataset. For example, in HIN-ENG,  \textit{$L_1$} is HIN and  \textit{$L_2$} is ENG. $^{\dagger}$We define MSA as ML and EA as EL. \#\textit{$L_1$} represents the number of tokens in the first language and \#\textit{$L_2$} represents the number of tokens in the second language.}
\resizebox{0.52\textwidth}{!}{
\begin{tabular}{lrrll}
\toprule
 & \multicolumn{1}{c}{\textbf{\#\textit{$L_1$}}} & \multicolumn{1}{c}{\textbf{\#\textit{$L_2$}}} & \textbf{ML} & \textbf{EL} \\ \midrule
\multicolumn{5}{l}{NER} \\ \midrule
HIN-ENG & 13,860 & 11,391 & HIN & ENG \\
SPA-ENG & 163,824 & 402,923 & \multicolumn{1}{l}{ENG} & \multicolumn{1}{l}{SPA} \\
MSA-EA$^{\dagger}$ & \multicolumn{1}{c}{-} & \multicolumn{1}{c}{-} & \multicolumn{1}{l}{MSA} & \multicolumn{1}{l}{EA} \\ \midrule
\multicolumn{5}{l}{POS} \\ \midrule
HIN-ENG & 12,589 & 9,882 & HIN & ENG \\
SPA-ENG & 178,135 & 92,517 & SPA & ENG \\ \bottomrule
\end{tabular}
}
\label{dataset}
\end{table}

\subsubsection{Models}
\begin{figure}[!t]
  \centering
  \includegraphics[width=\linewidth]{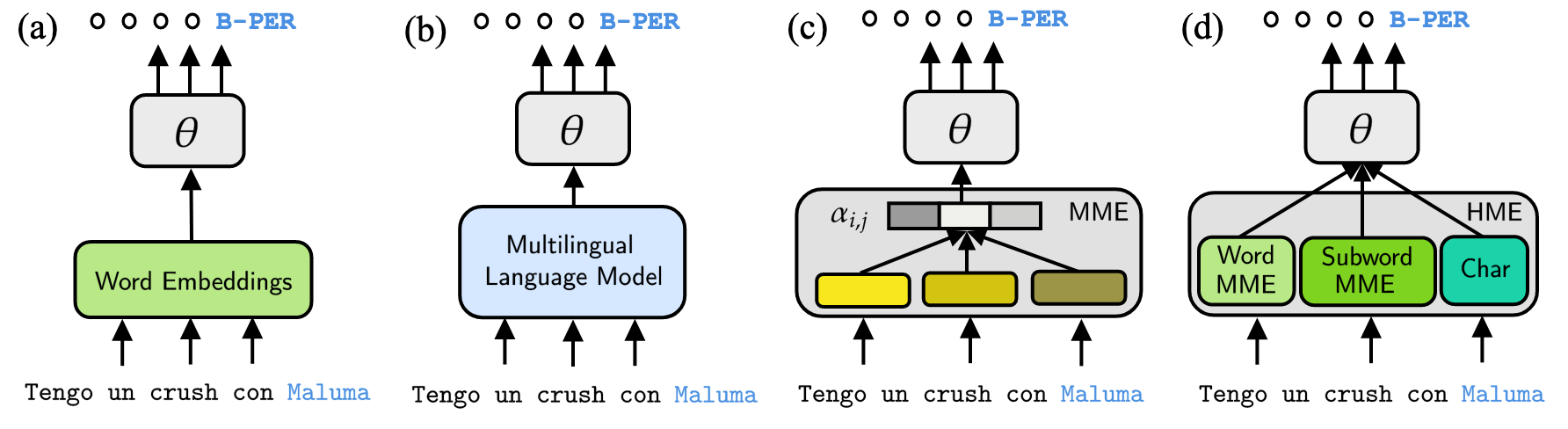}
  \caption{Model architectures for code-switched sequence labeling: (a) model using word embeddings, (b) model using multilingual language model, (c) model using multilingual meta-embeddings (MME), and (d) model using hierarchical meta-embeddings (HME).}
  \label{fig:representation_learning_architecture}
\end{figure}

Here, we compare models:  word embedings, bilingual embeddings, and multilingual pre-trained models. We show the general architectures for code-switched sequence labeling in 
Figure~\ref{fig:representation_learning_architecture}.

\paragraph{Scratch}
We train transformer-based models by following the mBERT model structure, and all of the parameters are randomly initialized. We train transformer models with four and six layers with a hidden size of 768. We want to measure the effectiveness of pre-training on multilingual models. We start the training with a learning rate of 1e-4 and an early stop of 10 epochs.

\paragraph{Word Embeddings}
We use FastText embeddings~\cite{grave2018learning,mikolov2018advances} to represent our input in our transformer models. The model consists of a four-layer transformer encoder with four heads and a hidden size of 200. We train a transformer followed by a CRF layer \cite{lafferty2001conditional}. We train our model with a learning rate of 0.1, a batch size of 32 and an early stop of 10 epochs. We also train our model with only ML and EL embeddings. We freeze all embeddings and only keep the classifier trainable.

\paragraph{Bilingual embeddings}
We want to use bilingual embeddings, such as MUSE~\cite{lample2018word}, to align the embeddings space between the ML and EL. We first conduct adversarial training using the SGD optimizer with a learning rate of 0.1, and then we perform the refinement procedure for five iterations using the Procrustes solution and CSLS~\cite{lample2018word}. After we align the embeddings, we train our model with the aligned word embeddings (MUSE (ML $\rightarrow$ EL) or MUSE (EL $\rightarrow$ ML)) on the code-switching tasks. 

\paragraph{Pre-trained Multilingual Models}
We use pre-trained models from Huggingface.~\footnote{\href{https://github.com/huggingface/transformers}{https://github.com/huggingface/transformers}} On top of each model, we put a fully-connected layer classifier. We train the model with a learning rate between [1e-5, 5e-5] with a decay of 0.1 and a batch size of 8. For large models, such as $\text{XLM-R}_{\text{LARGE}}$ and $\text{XLM-MLM}_{\text{LARGE}}$, we freeze the embeddings layer to fit into a single GPU. 

\paragraph{Multilingual Meta-Embeddings (MME)}
We use pre-trained word embeddings to train our MME. Table~\ref{embedding-list} shows the embeddings used for each dataset. We freeze all embeddings and train a transformer classifier with the CRF. The transformer classifier consists of a hidden size of 200, a head of 4, and four layers. All models are trained with a learning rate of 0.1, an early stop of 10 epochs, and a batch size of 32. We release the implementation in a public code repository.~\footnote{\href{https://github.com/gentaiscool/meta-emb}{https://github.com/gentaiscool/meta-emb}} Table~\ref{embedding-list} shows the list of word embeddings used in MME.

\paragraph{Hierarchical Meta-Embeddings (HME)}
We train our HME model using the same embeddings as MME and pre-trained subword embeddings from~\citet{heinzerling2018bpemb}. The word embeddings used in the experiment are shown in Table~\ref{embedding-list} and the subword embeddings for each language pair are shown in Table~\ref{subword-embedding-list}. We follow the same hyper-parameters we use when we train MME models.

\paragraph{HME-Ensemble}
In order to improve the model's robustness from multiple predictions, we combine predictions from five HME models, and compute the final prediction by majority voting to achieve a consensus. This method has shown to be very effective in improving the robustness of an unseen test set~\cite{winata2019caire_hkust}. The advantage of applying this method is very simple to implement and can be easily spawned in multiple machines, as in parallel processes.

\begin{table}[!ht]
\centering
\caption{Embeddings list for MME.}
\resizebox{0.51\textwidth}{!}{
\begin{tabular}{ll}
\toprule
& \textbf{Word Embeddings List} \\ \midrule
\multicolumn{2}{l}{NER} \\ \midrule
HIN-ENG & FastText: Hindi, English ~\cite{grave2018learning}\\
SPA-ENG & FastText: Spanish, English, Catalan,
\\
& Portugese~\cite{grave2018learning}\\ 
& GLoVe: English-Twitter~\cite{pennington2014glove} \\
MSA-EA & FastText: Arabic, Egyptian~\cite{grave2018learning} \\ \midrule
\multicolumn{2}{l}{POS} \\ \midrule
HIN-ENG & FastText: Hindi, English~\cite{grave2018learning} \\
SPA-ENG & FastText: Spanish, English, Catalan, \\
& Portugese~\cite{grave2018learning}\\ 
& GLoVe: English-Twitter~\cite{pennington2014glove} \\ \bottomrule
\end{tabular}
}
\label{embedding-list}
\end{table}

\begin{table}[!ht]
\centering
\caption{Subword embeddings list for HME.}
\resizebox{0.51\textwidth}{!}{
\begin{tabular}{ll}
\toprule
 & \textbf{Subword Embeddings List} \\ \midrule
\multicolumn{2}{l}{NER} \\ \midrule
HIN-ENG & Hindi, English \\
SPA-ENG & Spanish, English, Catalan, Portugese\\ 
MSA-EA & Arabic, Egyptian \\ \midrule
\multicolumn{2}{l}{POS} \\ \midrule
HIN-ENG & Hindi, English \\
SPA-ENG & Spanish, English, Catalan, Portugese \\ 
\bottomrule
\end{tabular}
}
\label{subword-embedding-list}
\end{table}

\paragraph{Char2Subword}
We take the results of the Char2Subword model reported in~\citet{aguilar2020char2subword}.

\subsection{Results and Discussion}

We find that multilingual pre-trained language models, such as $\text{XLM-R}_{\text{BASE}}$, achieves similar or sometimes better results than the HME model. Note that HME uses word and subword pre-trained embeddings that are trained using significantly fewer data than mBERT and $\text{XLM-R}_{\text{BASE}}$ and can achieve on par performance to theirs. Interestingly, we also observed that $\text{XLM-R}_{\text{LARGE}}$ improves the performance significantly, but with a trade-off in the training and inference time, with 13x more parameters than HME-Ensemble for a marginal improvement (2\%). We evaluate all the models on the LinCE benchmark, and the development set results are shown in Table \ref{table:dev_result}. As expected, the Scratch models perform significantly worse than the other pre-trained models. Both FastText and MME use pre-trained word embeddings, but MME achieves a consistently higher F1 score than FastText in both NER and POS tasks. This demonstrates the importance of the contextualized self-attentive encoder. HME further improves on the F1 score of the MME models, suggesting that encoding hierarchical information from subword-level, word-level, and sentence-level representations can improve the code-switching task performance. Comparing HME with mBERT and XLM-R, we find that the HME models are able to obtain comparable F1 scores with up to 10x smaller model sizes, and this indicates that pre-trained multilingual word embeddings can achieve a good balance between performance and model size in code-switching tasks. We show the model's performance on the LinCE benchmark test set in Table \ref{table:test_result}. $\text{XLM-R}_{\text{LARGE}}$ achieves the best-average performance, with a 13x larger model size when we compare it to the HME-Ensemble model.

\begin{table*}[!ht]
\centering
\caption{Results on the development set of the LinCE benchmark. $^{\ddagger}$ The results are taken from~\citet{aguilar2020char2subword}. The number of parameters of mBERT (cased) is calculated by approximation.}
\resizebox{\textwidth}{!}{
\begin{tabular}{lc|rlrlrl|rlrl}
\toprule
& \multicolumn{1}{c}{} & \multicolumn{6}{c}{NER} & \multicolumn{4}{c}{POS} \\
\cmidrule(l{2pt}r{2pt}){3-8} \cmidrule(l{2pt}r{2pt}){9-12}
& \multicolumn{1}{c}{} & \multicolumn{2}{c}{HIN-ENG} & \multicolumn{2}{c}{SPA-ENG} & \multicolumn{2}{c}{MSA-EA} & \multicolumn{2}{c}{HIN-ENG} & \multicolumn{2}{c}{SPA-ENG} \\
 \cmidrule(l{2pt}r{2pt}){3-4} \cmidrule(l{2pt}r{2pt}){5-6}
 \cmidrule(l{2pt}r{2pt}){7-8}
 \cmidrule(l{2pt}r{2pt}){9-10}
 \cmidrule(l{2pt}r{2pt}){11-12}
\multicolumn{1}{l}{\textbf{Method}} & \multicolumn{1}{l}{\textbf{Avg Perf.}} &\multicolumn{1}{|c}{Params} & \multicolumn{1}{c}{F1} &  \multicolumn{1}{c}{Params} & \multicolumn{1}{c}{F1} & \multicolumn{1}{c}{Params} & \multicolumn{1}{c|}{F1} & \multicolumn{1}{c}{Params} & \multicolumn{1}{c}{Acc} & \multicolumn{1}{c}{Params} & \multicolumn{1}{c}{Acc} \\
 \midrule
Scratch (2L) & 63.40 & 96M & 46.51 & 96M & 32.75 & 96M & 60.14 & 96M & 83.20 & 96M & 94.39 \\
Scratch (4L) & 60.93 & 111M & 47.01 & 111M & 19.06 & 111M & 60.24 & 111M & 83.72 & 111M & 94.64 \\ \midrule
\multicolumn{11}{l}{Mono/Multilingual Word Embeddings} \\ \midrule
FastText (ML) & 76.43 & 4M & 63.58 & 18M & 57.10 & 16M & 78.42 & 4M & 84.63 & 6M & 98.41 \\ 
FastText (EL) & 76.71 & 4M & 69.79 & 18M & 58.34 & 16M & 72.68 & 4M & 84.40 & 6M & 98.36 \\ 
MUSE (ML $\rightarrow$ EL) & 76.54 & 4M & 64.05 & 18M & 58.00 & 16M & 78.50 & 4M & 83.82& 6M & 98.34 \\
MUSE (EL $\rightarrow$ ML) & 75.58 & 4M & 64.86 & 18M & 57.08 & 16M & 73.95 & 4M & 83.62 & 6M & 98.38 \\\midrule
\multicolumn{11}{l}{Pre-Trained Multilingual Models} \\\midrule
mBERT (uncased) & 79.46 & 167M & 68.08 & 167M & 63.73 & 167M & 78.61 & 167M & 90.42 & 167M & 96.48 \\
mBERT (cased){$^\ddagger$} & 79.97 & 177M & 72.94 & 177M & 62.66 & 177M & 78.93 & 177M & 87.86 & 177M & 97.29 \\
Char2Subword{$^\ddagger$} & 81.07 & 136M & 74.91 & 136M & 63.32 & 136M & 80.45 & 136M & 89.64 & 136M & 97.03 \\
$\text{XLM-R}_{\text{BASE}}$ & 81.90 & 278M & 76.85 & 278M & 62.76 & 278M & 81.24 & 278M & 91.51 & 278M & 97.12 \\
XLM-R$_{\text{LARGE}}$ & \textbf{84.39} & 565M & \textbf{79.62} & 565M & \textbf{67.18} & 565M & \textbf{85.19} & 565M & 92.78 & 565M & 97.20 \\ 
$\text{XLM-MLM}_{\text{LARGE}}$ & 81.41 & 572M & 73.91 & 572M & 62.89 & 572M & 82.72 & 572M & 90.33 & 572M & 97.19 \\ \midrule
\multicolumn{11}{l}{Multilingual Meta-Embeddings} \\ \midrule
Concat & 79.70 & 10M & 70.76 & 86M & 61.65 & 31M & 79.33 & 8M & 88.14 & 23M & 98.61 \\
Linear & 79.60 & 10M & 69.68 & 86M & 61.74 & 31M & 79.42 & 8M & 88.58 & 23M & 98.58 \\
Attention (MME) & 79.86 & 10M & 71.69 & 86M & 61.23 & 31M & 79.41 & 8M & 88.34 & 23M & 98.65 \\
HME & 81.60 & 12M & 73.98 & 92M & 62.09 & 35M & 81.26 & 12M & 92.01 & 30M & 98.66 \\
HME-Ensemble & \textbf{82.44} & 20M & 76.16 & 103M & 62.80 & 43M & 81.67 & 20M &	\textbf{92.84} & 40M & \textbf{98.74} \\ \bottomrule
\end{tabular}
}
\label{table:dev_result}
\end{table*}

\begin{table*}[!ht]
\centering
\caption{Results on the test set of the LinCE benchmark.$^{\ddagger}$ The results are taken from~\citet{aguilar2020char2subword}. $^{\dagger}$ The result is taken from the LinCE leaderboard.}
\resizebox{\textwidth}{!}{
\begin{tabular}{lrc|ccc|cc}
\toprule
& \multicolumn{2}{c}{} & \multicolumn{3}{c}{NER} & \multicolumn{2}{c}{POS} \\
\cmidrule(l{2pt}r{2pt}){4-6} \cmidrule(l{2pt}r{2pt}){7-8}
\multicolumn{1}{l}{\textbf{Method}} & \multicolumn{1}{c}{\textbf{Avg Params}} & \multicolumn{1}{l}{\textbf{Avg Perf.$\uparrow$}} &  \multicolumn{1}{c}{HIN-ENG} & \multicolumn{1}{c}{SPA-ENG} & \multicolumn{1}{c}{MSA-EA} & \multicolumn{1}{c}{HIN-ENG} & \multicolumn{1}{c}{SPA-ENG} \\ \midrule
English BERT (cased)$^{\dagger}$ & 108M & 75.80 & 74.46 & 61.15 & 59.44 & 87.02 & 96.92 \\
mBERT (cased)$^{\ddagger}$ & 177M & 77.08 & 72.57 & 64.05 & 65.39 &  86.30 & 97.07 \\
HME & 36M & 77.64 & 73.78 & 63.06 & 66.14 & 88.55 & 96.66 \\
Char2Subword$^{\ddagger}$ & 136M & 77.85 & 73.38 & 64.65 & 66.13 & 88.23 & 96.88 \\
XLM-MLM$_{\text{LARGE}}$ & 572M & 78.40 & 74.49 & 64.16 & 67.22 & 89.10 & 97.04 \\
XLM-R$_{\text{BASE}}$ & 278M & 78.75 & 75.72 & 64.95 & 65.13 & 91.00 & 96.96 \\
HME-Ensemble & \underline{45M} & \underline{79.17} & 75.97 & 65.11 & \textbf{68.71} & 89.30 & 96.78 \\
XLM-R$_{\text{LARGE}}$ & 565M & \textbf{80.96} & \textbf{80.70} & \textbf{69.55} & 65.78 & \textbf{91.59} & \textbf{97.18} \\
\bottomrule
\end{tabular}
}
\label{table:test_result}
\end{table*}

% \begin{table*}[!ht]
% \centering
% \caption{Results on LinCE benchmark.}
% \resizebox{\textwidth}{!}{
% \begin{tabular}{lccccccc}
% \toprule
% \multicolumn{1}{c}{\multirow{2}{*}{Model}} & \multicolumn{3}{c}{NER} & \multicolumn{2}{c}{POS} & \multirow{2}{*}{Avg}\\ \cmidrule{2-6}
% & SPA-ENG & HIN-ENG & MSA-EA & SPA-ENG & HIN-ENG \\ \midrule
% ELMo small~\cite{aguilar2020lince} & 52.58 & 68.79 & 56.68 & 96.34 & 86.71 & 72.22 \\
% BERT base cased~\cite{aguilar2020lince} & 61.15 & 74.46 & 59.44 & 96.92 & 87.02 & 75.80 \\
% mBERT~\cite{aguilar2020lince} & 64.05 & 72.57 & 65.39 & \textbf{97.07} & 86.30 & 77.08 \\ 
% Char2Subword~\cite{aguilar2020char2subword} & 64.65 & 73.38 & 66.13 & 96.88 & 88.23 & 77.85 \\ \midrule
% HME & \textbf{65.30} & \textbf{75.92} & \textbf{68.36} & 96.80 & \textbf{89.42} & \textbf{79.16} \\ \bottomrule
% \end{tabular}
% }
% \label{tab:results_lince}
% \end{table*}

% \subsubsection{LinCE Benchmark}

\begin{figure*}
    \centering
    \resizebox{0.48\textwidth}{!}{
        \includegraphics{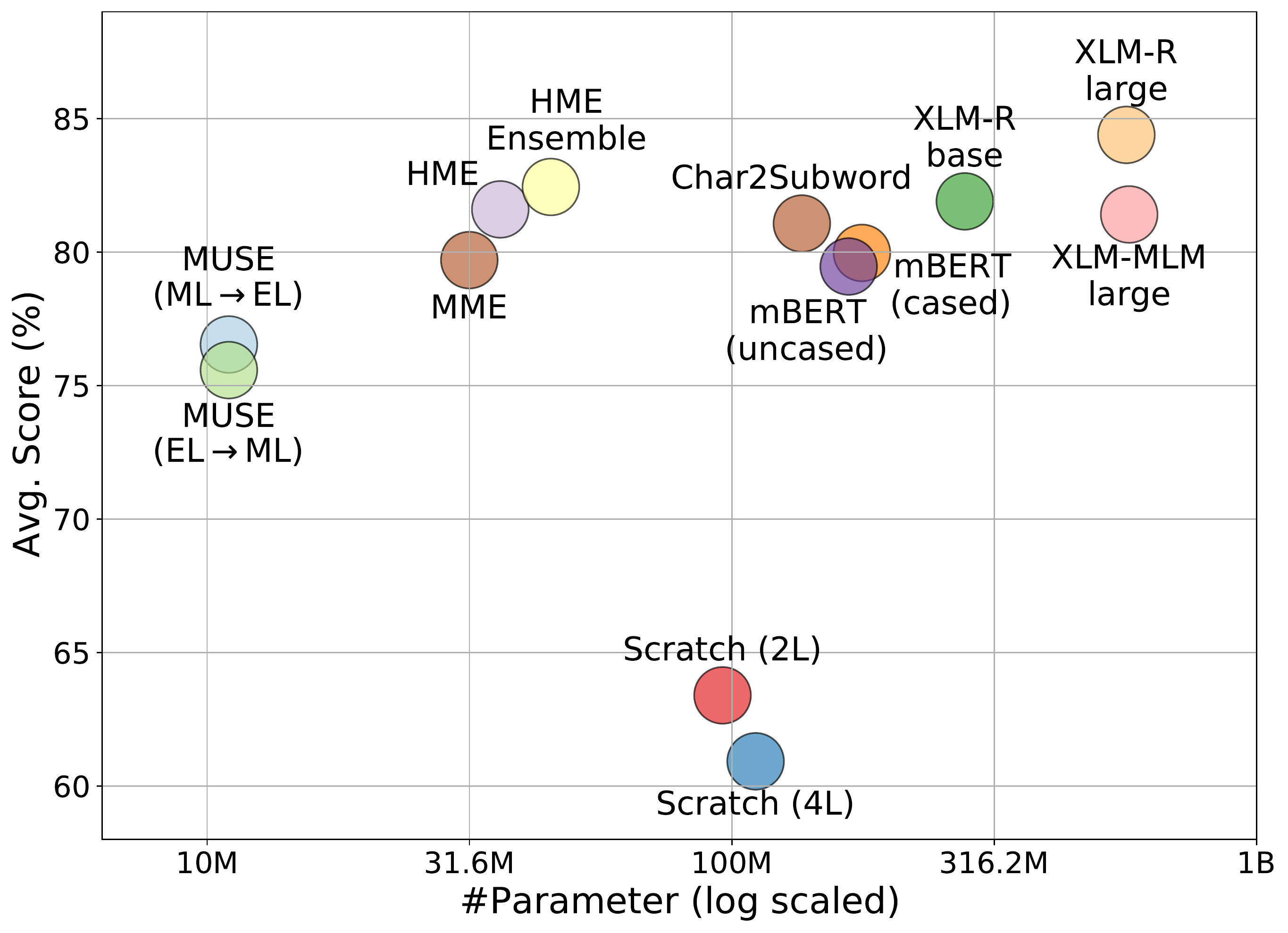}
    }
    \hspace{10pt}
    \resizebox{0.48\textwidth}{!}{
        \includegraphics{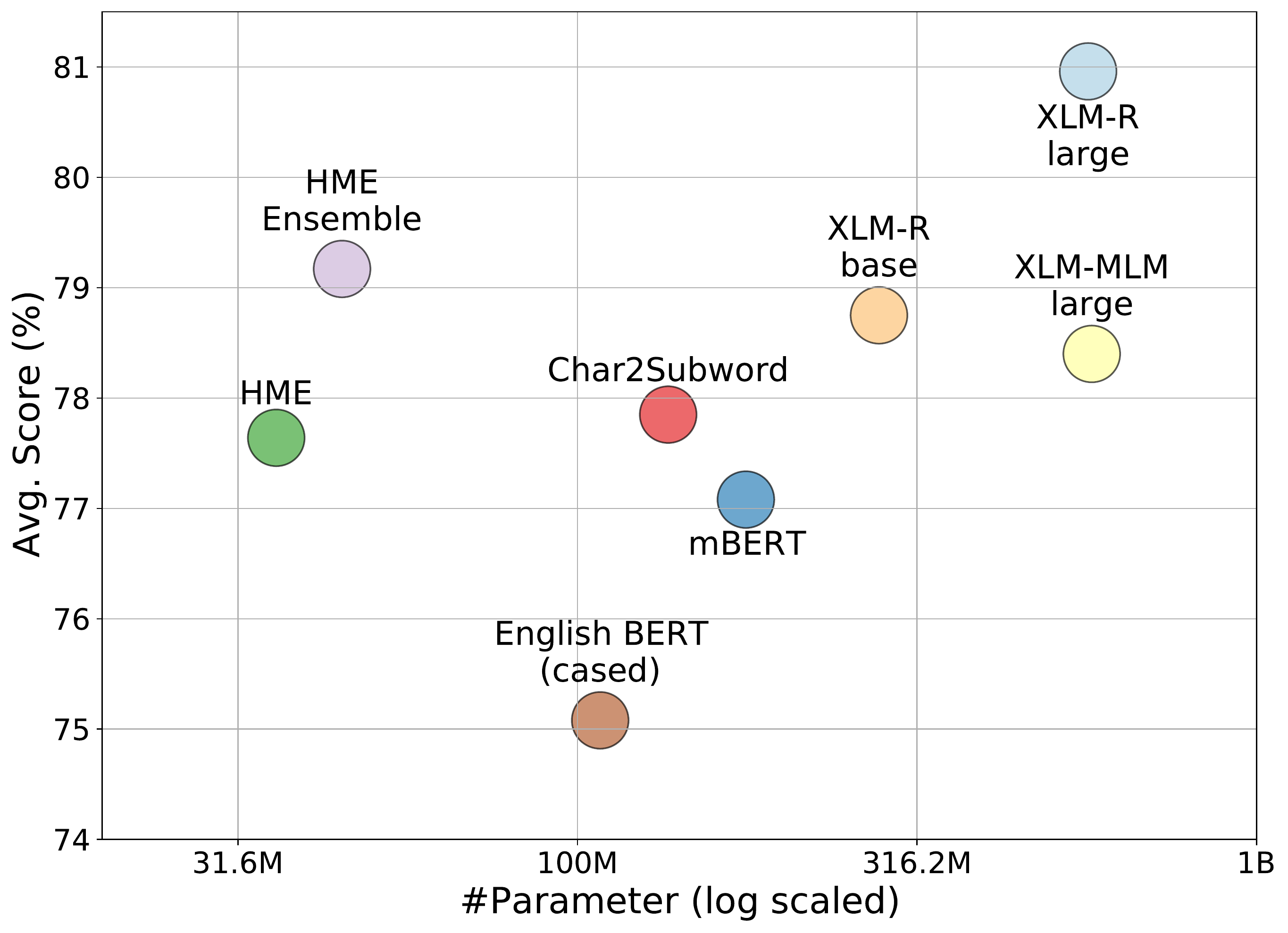}
    }
    \caption{Validation set (left) and test set (right) evaluation performance (y-axis) and parameter (x-axis) of different models on LinCE benchmark.}
    \label{fig:performance_vs_parameters}
\end{figure*}
\subsubsection{Performance vs. Model Size} As shown in Figure \ref{fig:performance_vs_parameters}, the Scratch models yield the worst average score, at 60.93 points. The model performance can be improved by around 10 points by using pre-trained embeddings FastText with only 10M parameters on average. On the other hand, the MME models, which have 31.6M parameters on average, achieve similar results to the mBERT models, with around 170M parameters. Interestingly, adding subwords and character embeddings to MME, such as in the HME models, further improves the performance of the MME models and achieves a 81.60 average score, similar to that of the XLM-R$_{\text{BASE}}$ and XLM-MLM$_{\text{LARGE}}$ models, but with less than one-fifth the number of parameters, at around 42.25M. The ensemble method adds further performance improvement of the HME model by around 1\%, with an additional 2.5M parameters compared to its non-ensemble counterparts.

\subsubsection{Inference Time} To compare the speed of different models, we generate data with random values and various sequence lengths: [16, 32, 64, 128, 256, 512, 1024, 2048, 4096]. We measure each model's inference time and collect the statistics of each at a particular sequence length by running the model 100 times. The experiment is performed on a single NVIDIA GTX1080Ti GPU. We do not include the pre-processing time in our analysis. It is evident that the pre-processing time for the meta-embeddings models is longer than for other models as pre-processing requires a tokenization step to be conducted for the input multiple times with different tokenizers. The reported sequence lengths are counted based on the input tokens of each model, and we use words for the MME and HME models, and subwords for the other models.
% As the inputs of FastText, MME, and HME models are in word-level, and other implementations are based on subwords (except for Char2Subword since the implementation code has not been released yet), we apply the sequence length for word-level in MME and HME models and subword-level in the other models. 
% As the inputs of MME and HME models, we use two BPE tokens and five character tokens.
% For each word in the dummy inputs of MME and HME models, we use two BPE tokens and five character tokens. This means that both MME and HME models encode double the other models' size on our analysis, which might be enough to compensate for the preprocessing time for the meta embedding model. As MME and HME models have different configurations for each task, we measure the inference time from all the configurations, each with 100 iterations, and take the combined statistics over all the configurations.

\begin{figure}
    \centering
    \resizebox{0.65\textwidth}{!}{
        \includegraphics{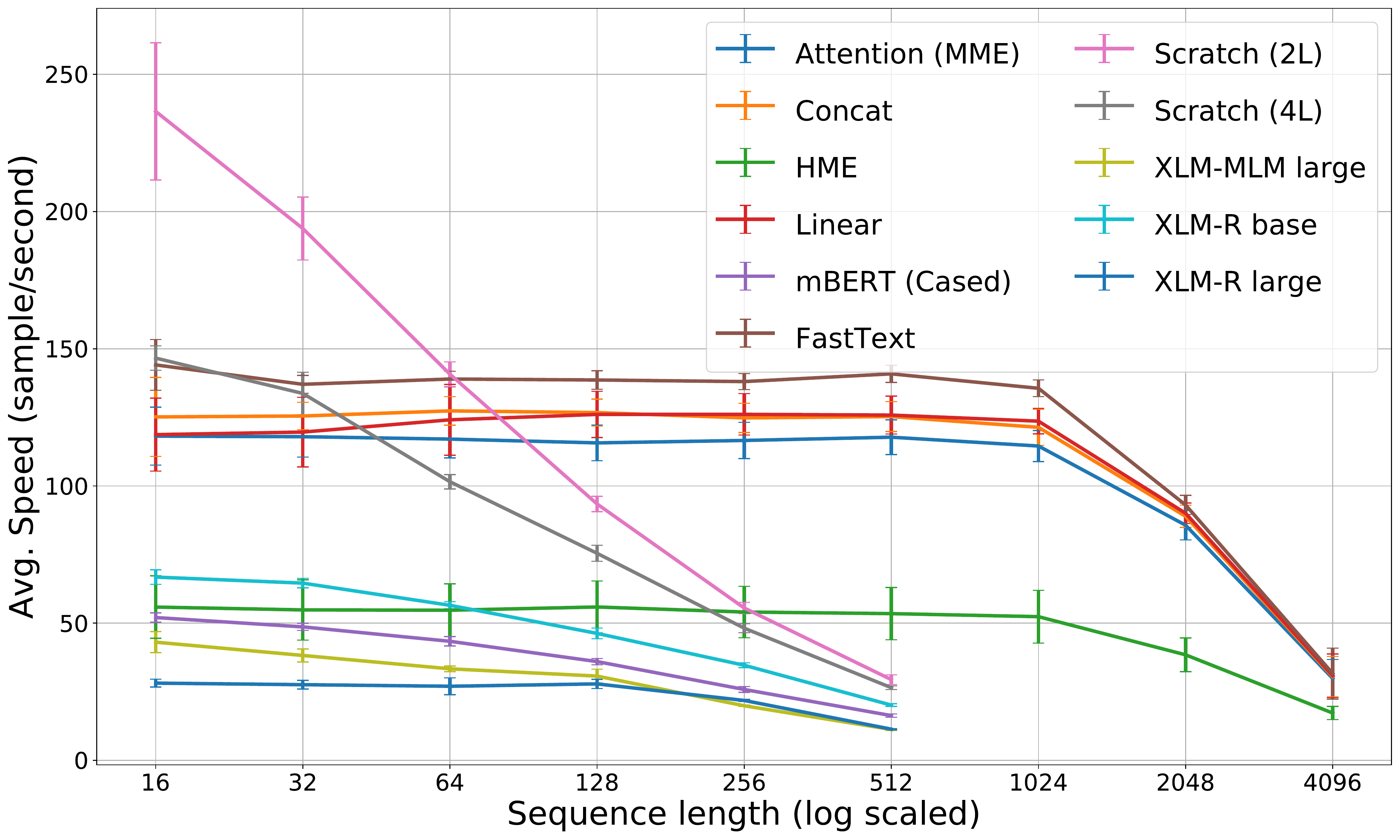}
    }
    \caption{Speed-to-sequence length comparison of different models.}
    \label{fig:speed_vs_seqlen}
\end{figure}

We show the results of the inference speed test in Figure \ref{fig:speed_vs_seqlen}. Although all pre-trained contextualized language models yield a high validation score, these models are the slowest in terms of inference time. The larger the size of the pre-trained model, the longer it takes to run a prediction. For shorter sequences, the HME model performs as quickly as the mBERT and XLM-R$_{\text{BASE}}$ models, and can retain its speed as the sequence length increases because of the smaller model dimensions in each layer. The FastText, MME, and Scratch models have high throughput in short-sequence settings by processing more than 150 samples per second. For longer sequences, the same behavior occurs, with the throughput of the Scratch models reducing as the sequence length increases, even becoming lower than that of the HME model when the sequence length is greater than or equal to 256. In addition, for the FastText, MME, and HME models, the throughput remains steady when the sequence length is less than 1024, and it starts to decrease afterwards.

\begin{table}[!t]
\centering
\caption{GPU memory consumption of different models with input size of 512.}
\resizebox{0.42\textwidth}{!}{
    \begin{tabular}{lrr}
    \toprule
    \textbf{Model} &  \textbf{Activation (MB)} \\
    \midrule
    FastText &               79.0 \\
    Concat &               85.3 \\
    Linear &                80.8 \\
    Attention (MME) &      88.0 \\
    HME &           154.8 \\
    Scratch (2L) &             133.0 \\
    Scratch (4L) &            264.0 \\
    mBERT &            597.0 \\
    XLM-R$_{\text{BASE}}$ &           597.0 \\
    XLM-R$_{\text{LARGE}}$ &           1541.0 \\
    XLM-MLM$_{\text{LARGE}}$ &        1158.0 \\
    \bottomrule
    \end{tabular}
}
\label{tab:memory_512}
\end{table}

\subsubsection{Memory Footprint} We record the memory footprint over different sequence lengths, and use the same setting for the FastText, MME, and HME models as in the inference time analysis. We record the size of each model on a GPU machine, and the size of the activation memory after performing a single forward operation on a sample with a certain sequence length. The result of the memory footprint analysis for a sequence length of 512 is shown in Table \ref{tab:memory_512}. Based on the results, we can see that the meta-embedding models use a significantly smaller memory footprint to store the model and activation memory. For example, the memory footprint of the HME model is less than that of the Scratch (4L) model, which has only four transformer encoder layers, a model dimension of 768 and a feed-forward dimension of 3,072. On the other hand, large pre-trained language models, such as XLM-MLM$_{\text{LARGE}}$ and XLM-R$_{\text{LARGE}}$, use a much larger memory for storing the activation memory compared to all other models. 
The complete results of the memory footprint analysis can be found in Appendix.

\section{Short Summary}
In this chapter, we propose two approaches for learning representations for code-switching data. First, we introduce a BiLSTM-based model with a hierarchical architecture using a bilingual character RNN to address the OOV words issue. Moreover, token replacement, token normalization, and transfer learning reduce the OOV words rate even further and significantly improves the performance. We show that our model can achieve similar performance to models using gazetteers and hand-picked features. Then, we propose Hierarchical Meta-Embeddings (HME), which learns how to combine multiple monolingual word-level and subword-level embeddings to create language-agnostic representations without specific language information. We achieve the state-of-the-art results on the task of NER for English-Spanish code-switching data. We also show that our model can leverage subword information very effectively from languages from different roots to generate better word representations. 

We study the effectiveness of multilingual language models so as to understand their capability and adaptability to the code-switching setting. We run experiments on NER and POS tagging on three different language pairs, and find that a pre-trained multilingual model does not necessarily guarantee high-quality representations on code-switching, while the HME model achieves similar results to mBERT and $\text{XLM-R}_{\text{BASE}}$, but with significantly fewer parameters. We also find that, while $\text{XLM-R}_{\text{LARGE}}$ has better performance by a large margin, this comes with a substantial cost in the training and inference time, using 13x more parameters than HME-Ensemble for only a 2\% improvement. 
\chapter{Conclusion}

This thesis focuses on transfer learning methods for code-switching NLP and speech processing tasks by leveraging monolingual resources. We discussed the main challenges in code-switching and introduced novel transfer learning approaches for language modeling, speech recognition, and representation learning for sequence labeling tasks. We proposed  multilingual transfer learning by utilizing monolingual data to improve code-switching representations to address the need for huge amounts of data. In this chapter, we conclude the thesis, summarize our contributions, and discuss possible future work.

We examined data augmentation approaches to train language models for code-switching by generating synthetic code-switching data using a neural network language model with a copy mechanism, called Pointer-Gen. Pointer-Gen leverages information from the input to ensure high-quality code-switching sentence generation and eliminates the dependence on the aligner or tagger. It learns the distribution of the real code-switching data instead of relying on the linguistic prior information. The proposed approach increases the variance of the corpus to increase the robustness of our language models. Pointer-Gen samples new code-switching sentences from the distribution of the code-switching data in a zero-shot setting. The model works effectively for the English-Mandarin language pair, in which the languages are inherently different in terms of their grammatical structure. We found that our data augmentation method outperforms the method with an equivalence constraint and other neural-based augmentation methods.

We presented approaches to train language models with multi-task training that leverages syntactic information. We train our model by jointly learning the language modeling task and part-of-speech sequence tagging task on code-switched utterances. We incorporate language information into part-of-speech tags to create bilingual tags that distinguish between languages, and learn the correlation between language modeling and the next part-of-speech prediction. Indeed, the syntactic information helps the model to be aware of code-switching points, and it boosts the performance over the language model.

We introduced multi-task learning approaches to train code-switching speech recognition by proposing transfer learning methods. Our methods apply meta-learning by judiciously extracting information from high-resource monolingual datasets. The optimization conditions the model to retrieve useful learned information that is focused on the code-switching domain. The meta-transfer learning quickly adapts the model to the code-switching task from a number of monolingual tasks by learning to learn. Experimental results showed that our proposed method preserves the monolingual tasks' performance, and it is faster to converge.

We discussed state-of-the-art multilingual representation learning methods for code-switched named entity recognition. We introduced meta-embeddings considering the commonalities across languages and lexical compositionality. We found that this method is language-agnostic, and it is very effective and efficient compared to large contextual language models, such as mBERT and XLM-R in the code-switching domain. We also found that the hierarchical meta-embeddings model obtains comparable F1 and accuracy scores with up to 10x smaller model sizes compared to pre-trained multilingual models.

Lastly, the main contribution of this thesis is that it emphasizes the importance of multilingual transfer learning on code-switching tasks. We proposed language-agnostic end-to-end approaches that are not dependent on particular languages to improve the generalization of our models on code-switched data. And, the transfer learning from monolingual data is very effective for code-switching. It is also worth noting that linguistic theory can be used as prior information to understand how code-switches are triggered.

In future work, we expect to develop a better training objective to learn a multilingual contextualized language model in a self-supervised fashion that can well represent mixed languages since, according to our empirical observation, the current pre-trained models lack cross-linguality. We would also like to explore a more effective transfer learning method to leverage monolingual data and parallel data.

% We adapted ideas from linguistics into statistical approaches, utilized transfer learning from monolingual data to improve code-switching representations and effectively address the challenges. 

\newpage
\addcontentsline{toc}{chapter}{Reference}
\bibliographystyle{IEEEtranN}
\bibliography{reference} 

\newpage
\addcontentsline{toc}{chapter}{Publication}
\null\skip0.2in
\begin{center}
{\bf \Large \underline{List of Publications}}
\end{center}
\vspace{12mm}

% \paragraph{Main Conference Papers}
(* denotes equal contribution)
\begin{itemize}
    \item Zihan Liu, \textbf{Genta Indra Winata}, Samuel Cahyawijaya, Andrea Madotto, Zhaojiang Lin, Pascale Fung. "On the Importance of Word Order Information in Cross-lingual Sequence Labeling." In AAAI, 2021.
    \item Andrea Madotto, Samuel Cahyawijaya, \textbf{Genta Indra Winata}, Yan Xu, Zihan Liu, Zhaojiang Lin, Pascale Fung. "Learning Knowledge Bases with Parameters for Task-Oriented Dialogue Systems." In Proceedings of the 2020 Conference on Empirical Methods in Natural Language Processing: Findings, 2020.
    \item Zihan Liu, \textbf{Genta Indra Winata}, Peng Xu, Zhaojiang Lin, Pascale Fung. "Cross-lingual Spoken Language Understanding with Regularized Representation Alignment." In Proceedings of the 2020 Conference on Empirical Methods in Natural Language Processing (EMNLP), 2020.
    \item Zhaojiang Lin, Andrea Madotto, \textbf{Genta Indra Winata}, Pascale Fung. "MinTL: Minimalist transfer learning for task-oriented dialogue systems." Proceedings of the 2020 Conference on Empirical Methods in Natural Language Processing (EMNLP), 2020.
    \item Bryan Wilie*, Karissa Vincentio*, \textbf{Genta Indra Winata}*, Samuel Cahyawijaya*, Xiaohong Li, Zhi Yuan Lim, Sidik Soleman, Rahmad Mahendra, Pascale Fung, Syafri Bahar, Ayu Purwarianti. "IndoNLU: Benchmark and resources for evaluating indonesian natural language understanding." In Proceedings of the 1st Conference of the Asia-Pacific Chapter of the Association for Computational Linguistics and the 10th International Joint Conference on Natural Language Processing, 2020.
    \item \textbf{Genta Indra Winata}*, Samuel Cahyawijaya*, Zhaojiang Lin, Zihan Liu, Peng Xu, Pascale Fung. "Meta-transfer learning for code-switched speech recognition." In Proceedings of the 58th Annual Meeting of the Association for Computational Linguistics, 2020.
    \item Zihan Liu, \textbf{Genta Indra Winata}, Peng Xu, Pascale Fung. "Coach: A Coarse-to-Fine Approach for Cross-domain Slot Filling." In Proceedings of the 58th Annual Meeting of the Association for Computational Linguistics, 2020.
    \item \textbf{Genta Indra Winata}*, Samuel Cahyawijaya*, Zihan Liu*, Zhaojiang Lin, Andrea Madotto, Peng Xu, Pascale Fung. "Learning Fast Adaptation on Cross-Accented Speech Recognition." In INTERSPEECH, 2020.
    \item Zihan Liu*, \textbf{Genta Indra Winata}*, Zhaojiang Lin, Peng Xu, Pascale Fung. "Attention-Informed Mixed-Language Training for Zero-shot Cross-lingual Task-oriented Dialogue Systems." In AAAI, 2020.
    \item \textbf{Genta Indra Winata}*, Samuel Cahyawijaya*, Zhaojiang Lin, Zihan Liu, and Pascale Fung. "Lightweight and Efficient End-to-End Speech Recognition Using Low-Rank Transformer." In ICASSP 2020-2020 IEEE International Conference on Acoustics, Speech and Signal Processing (ICASSP), pp. 6144-6148. IEEE, 2020.
    \item Zhaojiang Lin, Peng Xu, \textbf{Genta Indra Winata}, Farhad Bin Siddique, Zihan Liu, Jamin Shin, and Pascale Fung. "CAiRE: An End-to-End Empathetic Chatbot." In AAAI, pp. 13622-13623. 2020.
    \item Zihan Liu, Jamin Shin, Yan Xu, \textbf{Genta Indra Winata}, Peng Xu, Andrea Madotto, and Pascale Fung. "Zero-shot Cross-lingual Dialogue Systems with Transferable Latent Variables." In Proceedings of the 2019 Conference on Empirical Methods in Natural Language Processing and the 9th International Joint Conference on Natural Language Processing (EMNLP-IJCNLP), pp. 1297-1303. 2019.
    \item \textbf{Genta Indra Winata}, Zhaojiang Lin, Jamin Shin, Zihan Liu, and Pascale Fung. "Hierarchical Meta-Embeddings for Code-Switching Named Entity Recognition." In Proceedings of the 2019 Conference on Empirical Methods in Natural Language Processing and the 9th International Joint Conference on Natural Language Processing (EMNLP-IJCNLP), pp. 3532-3538. 2019.
    \item \textbf{Genta Indra Winata}, Andrea Madotto, Chien-Sheng Wu, and Pascale Fung. "Code-Switched Language Models Using Neural Based Synthetic Data from Parallel Sentences." In Proceedings of the 23rd Conference on Computational Natural Language Learning (CoNLL), pp. 271-280. 2019.
    \item \textbf{Genta Indra Winata}, Andrea Madotto, Jamin Shin, Elham J. Barezi, and Pascale Fung. "On the Effectiveness of Low-Rank Matrix Factorization for LSTM Model Compression." Proceedings of the 33rd Pacific Asia Conference on Language, Information and Computation (PACLIC), 2019.
    \item Zhaojiang Lin, \textbf{Genta Indra Winata}, and Pascale Fung. "Learning comment generation by leveraging user-generated data." In ICASSP 2019-2019 IEEE International Conference on Acoustics, Speech and Signal Processing (ICASSP), pp. 7225-7229. IEEE, 2019.
% \end{itemize}

% \paragraph{Shared Task Workshop Papers}
% \begin{itemize}
    \item Dan Su, Yan Xu, \textbf{Genta Indra Winata}, Peng Xu, Hyeondey Kim, Zihan Liu, and Pascale Fung. "Generalizing Question Answering System with Pre-trained Language Model Fine-tuning." In EMNLP 2019 MRQA Workshop, p. 203. 2019.
    \item Zihan Liu, Yan Xu, \textbf{Genta Indra Winata}, and Pascale Fung. "Incorporating Word and Subword Units in Unsupervised Machine Translation Using Language Model Rescoring." In Proceedings of the Fourth Conference on Machine Translation (Volume 2: Shared Task Papers, Day 1), pp. 275-282. 2019.
    \item \textbf{Genta Indra Winata}*, Andrea Madotto*, Zhaojiang Lin, Jamin Shin, Yan Xu, Peng Xu, and Pascale Fung. "CAiRE\_HKUST at SemEval-2019 Task 3: Hierarchical Attention for Dialogue Emotion Classification." In Proceedings of the 13th International Workshop on Semantic Evaluation, pp. 142-147. 2019.
    \item \textbf{Genta Indra Winata}, Zhaojiang Lin, and Pascale Fung. "Learning multilingual meta-embeddings for code-switching named entity recognition." In Proceedings of the 4th Workshop on Representation Learning for NLP (RepL4NLP-2019), pp. 181-186. 2019.
    \item Zhaojiang Lin, Andrea Madotto, \textbf{Genta Indra Winata}, Zihan Liu, Yan Xu, Cong Gao, and Pascale Fung. "Learning to learn sales prediction with social media sentiment." In Proceedings of the First Workshop on Financial Technology and Natural Language Processing, pp. 47-53. 2019.
    \item \textbf{Genta Indra Winata}, Chien-Sheng Wu, Andrea Madotto, and Pascale Fung. "Bilingual Character Representation for Efficiently Addressing Out-of-Vocabulary Words in Code-Switching Named Entity Recognition." In Proceedings of the Third Workshop on Computational Approaches to Linguistic Code-Switching, pp. 110-114. 2018.
% \end{itemize}

% \paragraph{Main Workshop Papers}
% \begin{itemize}
    \item \textbf{Genta Indra Winata}, Andrea Madotto, Chien-Sheng Wu, and Pascale Fung. "Code-Switching Language Modeling using Syntax-Aware Multi-Task Learning." In Proceedings of the Third Workshop on Computational Approaches to Linguistic Code-Switching, pp. 62-67. 2018.
    \item Chien-Sheng Wu, Andrea Madotto, \textbf{Genta Indra Winata}, and Pascale Fung. "End-to-end dynamic query memory network for entity-value independent task-oriented dialog." In 2018 IEEE International Conference on Acoustics, Speech and Signal Processing (ICASSP), pp. 6154-6158. IEEE, 2018.
    \item \textbf{Genta Indra Winata}, Onno Pepijn Kampman, and Pascale Fung. "Attention-based lstm for psychological stress detection from spoken language using distant supervision." In 2018 IEEE International Conference on Acoustics, Speech and Signal Processing (ICASSP), pp. 6204-6208. IEEE, 2018.
    \item \textbf{Genta Indra Winata}, Onno Kampman, Yang Yang, Anik Dey, and Pascale Fung. "Nora the Empathetic Psychologist." In INTERSPEECH, pp. 3437-3438. 2017.
    \item Chien-Sheng Wu*, Andrea Madotto*, \textbf{Genta Winata}, and Pascale Fung. "End-to-end recurrent entity network for entity-value independent goal-oriented dialog learning." In Dialog System Technology Challenges Workshop, DSTC6. 2017.
\end{itemize}

\newpage
\addcontentsline{toc}{chapter}{Appendix}
\appendix
\chapter*{Appendix}

\section*{Multi-Task Learning Results}
Results with different hyper-parameter settings

\begin{table}[!htb]
\centering
\caption{Language model results in Phase I}
\label{appendix-results1}
\begin{tabular}{@{}lcccccc@{}}
\hline
\textbf{Model} & \multicolumn{1}{|c|}{\textbf{\begin{tabular}[c]{@{}c@{}}Hidden\\ size\end{tabular}}} & \textbf{\begin{tabular}[c]{@{}c@{}}Embedding\\ size\end{tabular}} & \multicolumn{1}{|c|}{\textbf{Dropout}} & \textbf{POS tag dropout} & \multicolumn{1}{|c|}{\textbf{\begin{tabular}[c]{@{}c@{}}PPL \\ dev\end{tabular}}} & \textbf{\begin{tabular}[c]{@{}c@{}}PPL\\ test\end{tabular}} \\ \hline
\multirow{2}{*}{LSTM} & \multicolumn{1}{|c|}{200} & 200 & \multicolumn{1}{|c|}{0.2} & - & \multicolumn{1}{|c|}{197.5} & 196.84 \\ \cline{2-7} 
& \multicolumn{1}{|c|}{500} & 500 & \multicolumn{1}{|c|}{0.4} & - & \multicolumn{1}{|c|}{190.33} & 185.91 \\ \hline
% & 1000 & 0.6 & - & x & x \\ \hline
\multirow{2}{*}{+ syntactic features} & \multicolumn{1}{|c|}{200} & 200 & \multicolumn{1}{|c|}{0.2} & - & \multicolumn{1}{|c|}{187.37} & 184.87 \\ \cline{2-7} 
& \multicolumn{1}{|c|}{500} & 500 & \multicolumn{1}{|c|}{0.4} & - & \multicolumn{1}{|c|}{178.51} & 176.57 \\ \hline
% & 1000 & 0.6 & - & x & x \\ \hline
\multirow{2}{*}{Multi-task ($p=0.25$)} & \multicolumn{1}{|c|}{200} & 200 & \multicolumn{1}{|c|}{0.4} & 0.2 & \multicolumn{1}{|c|}{180.91} & 178.18 \\ \cline{2-7} 
& \multicolumn{1}{|c|}{\textbf{500}} & \textbf{500} & \multicolumn{1}{|c|}{\textbf{0.4}} & \textbf{0.4} & \multicolumn{1}{|c|}{\textbf{173.55}} & \textbf{174.96} \\ \hline
% & 1000 & 0.6 & 0.6 & x & x \\ \hline
\multirow{2}{*}{Multi-task ($p=0.50$)} & \multicolumn{1}{|c|}{200} & 200 & \multicolumn{1}{|c|}{0.4} & 0.2 & \multicolumn{1}{|c|}{182.6} & 178.75 \\ \cline{2-7} 
& \multicolumn{1}{|c|}{500} & 500 & \multicolumn{1}{|c|}{0.4} & 0.4 & \multicolumn{1}{|c|}{175.23} & 173.89 \\ \hline 
% & 1000 & 0.6 & 0.6 & 171.70 & 168.46 \\ \hline
\multirow{2}{*}{Multi-task ($p=0.75$)} & \multicolumn{1}{|c|}{200} & 200 & \multicolumn{1}{|c|}{0.4} & 0.2 & \multicolumn{1}{|c|}{180.90} & 178.18 \\ \cline{2-7} 
& \multicolumn{1}{|c|}{500} & 500 & \multicolumn{1}{|c|}{0.4} & 0.4 & \multicolumn{1}{|c|}{185.83} & 178.49 \\ \hline
% & 1000 & 0.6 & 0.6 & 174.20 & 168.51 \\ \hline
\end{tabular}
\end{table}

\begin{table}[!htb]
\centering
\caption{Language model results in SEAME Phase II}
\label{appendix-results2}
\begin{tabular}{@{}lcccccc@{}}
\hline
\textbf{Model} & \multicolumn{1}{|c|}{\textbf{\begin{tabular}[c]{@{}c@{}}Hidden\\ size\end{tabular}}} & \textbf{\begin{tabular}[c]{@{}c@{}}Embedding\\ size\end{tabular}} & \multicolumn{1}{|c|}{\textbf{Dropout}} & \textbf{POS tag dropout} & \multicolumn{1}{|c|}{\textbf{\begin{tabular}[c]{@{}c@{}}PPL \\ dev\end{tabular}}} & \textbf{\begin{tabular}[c]{@{}c@{}}PPL\\ test\end{tabular}} \\ \hline
\multirow{2}{*}{RNNLM} & \multicolumn{1}{|c|}{200} & 200 & \multicolumn{1}{|c|}{-} & - & \multicolumn{1}{|c|}{181.87} & 176.80 \\ \cline{2-7} 
& \multicolumn{1}{|c|}{500} & 500 & \multicolumn{1}{|c|}{-} & - & \multicolumn{1}{|c|}{178.35} & 171.27 \\ \hline 
% & 1000 & - & - & 176.92 & 170.18 \\ \hline
\multirow{2}{*}{LSTM} & \multicolumn{1}{|c|}{200} & 200 & \multicolumn{1}{|c|}{0.2} & - & \multicolumn{1}{|c|}{156.77} & 159.58 \\ \cline{2-7} 
& \multicolumn{1}{|c|}{500} & 500 & \multicolumn{1}{|c|}{0.4} & - & \multicolumn{1}{|c|}{150.65} & 153.06 \\ \hline
% & 1000 & 0.6 & - & x & x \\ \hline
\multirow{2}{*}{+ syntactic features} & \multicolumn{1}{|c|}{200} & 200 & \multicolumn{1}{|c|}{0.2} & - & \multicolumn{1}{|c|}{153.6} & 152.66 \\ \cline{2-7} 
& \multicolumn{1}{|c|}{500} & 500 & \multicolumn{1}{|c|}{0.4} & - & \multicolumn{1}{|c|}{147.44} & 148.38 \\ \hline
% & 1000 & 0.6 & - & x & x \\ \hline
\multirow{2}{*}{Multi-task ($p=0.25$)} & \multicolumn{1}{|c|}{200} & 200 & \multicolumn{1}{|c|}{0.4} & 0.2 & \multicolumn{1}{|c|}{149.68} & 149.84 \\ \cline{2-7} 
& \multicolumn{1}{|c|}{\textbf{500}} & \textbf{500} & \multicolumn{1}{|c|}{\textbf{0.4}} & \textbf{0.4} & \multicolumn{1}{|c|}{\textbf{141.86}} & \textbf{141.71} \\ \hline
% & 1000 & 0.6 & 0.6 & x & x \\ \hline
\multirow{2}{*}{Multi-task ($p=0.50$)} & \multicolumn{1}{|c|}{200} & 200 & \multicolumn{1}{|c|}{0.4} & 0.2 & \multicolumn{1}{|c|}{150.92} & 152.38 \\ \cline{2-7} 
& \multicolumn{1}{|c|}{500} & 500 & \multicolumn{1}{|c|}{0.4} & 0.4 & \multicolumn{1}{|c|}{144.18} & 144.27 \\ \hline
% & 1000 & 0.6 & 0.6 & x & x \\ \hline
\multirow{2}{*}{Multi-task ($p=0.75$)} & \multicolumn{1}{|c|}{200} & 200 & \multicolumn{1}{|c|}{0.4} & 0.2 & \multicolumn{1}{|c|}{150.32} & 151.22 \\ \cline{2-7} 
& \multicolumn{1}{|c|}{500} & 500 & \multicolumn{1}{|c|}{0.4} & 0.4 & \multicolumn{1}{|c|}{145.08} & 144.85 \\ \hline
% & 1000 & 0.6 & 0.6 & x & x \\ \hline
\end{tabular}
\end{table}

\clearpage
\subsection*{SEAME Data Split}
We split the recording ids into train, development, and test set as the following:

\begin{table}[!htb]
\centering
\caption{Recording distribution in Phase I}
\label{recording-list-phase1}
\begin{tabular}{@{}ccc@{}}
\toprule
\multirow{2}{*}{\textbf{Data}} & \multicolumn{2}{c}{\textbf{Recording list}} \\ \cmidrule{2-3} 
 & \multicolumn{1}{c}{\textbf{Conversation}} & \textbf{Interview} \\ \midrule
\textbf{Train} & \multicolumn{1}{c}{\begin{tabular}[c]{@{}c@{}}
02NC03FBX, 02NC04FBY, 03NC05FAX \\
03NC06FAY,04NC07FBX,04NC08FBY\\
06NC11MAX,06NC12MAY,07NC13MBP\\
07NC14FBQ,08NC15MBP,08NC16FBQ\\
09NC17FBP,09NC18MBQ,10NC19MBP\\
10NC20MBQ,11NC21FBP,11NC22MBQ\\
12NC23FBP,12NC24FBQ,13NC25MBP\\
13NC26MBQ,14NC27MBP,14NC28MBQ\\
16NC31FBP,16NC32FBQ,18NC35FBP\\
18NC36MBQ,19NC37MBP,19NC38FBQ\\
21NC41MBP,22NC43FBP,22NC44MBQ\\
23NC35FBQ,23NC45MBP,24NC35FBQ\\
24NC45MBP,25NC43FBQ,25NC47MBP\\
26NC48FBP,26NC49FBQ,27NC47MBQ\\
27NC50FBP,28NC51MBP,28NC52FBQ\\
29NC53MBP,29NC54FBQ,30NC48FBP\\
30NC49FBQ,31NC35FBQ,31NC50XFB\\
32NC36MBQ,32NC50FBP,33NC37MBP\\
33NC43FBQ,34NC37MBP,35NC56MBP\\
36NC46FBQ,37NC45MBP,38NC50FBP\\
39NC57FBX,40NC58FAY,41NC59MAX\\
42NC60FBQ,44NC44MBQ,45NC22MBQ\\
46NC41MBP,46NC41MBP\\
\end{tabular}} & \begin{tabular}[c]{@{}c@{}}
NI02FAX,NI04FBX,NI05MBQ\\
NI06FBP,NI07FBQ,NI08FBP\\
NI09FBP,NI10FBP,NI11FBP\\
NI12MAP,NI13MBQ,NI14MBP\\
NI15FBQ,NI16FBP,NI17FBQ\\
NI18MBP,NI19MBQ,NI20MBP\\
NI21MBQ,NI22FBP,NI23FBQ\\
NI24MBP,NI25MBQ,NI26FBP\\
NI27MBQ,NI28MBP,NI29MBP\\
NI30MBQ,NI31FBP,NI32FBQ\\
NI33MBP,NI34FBQ,NI35FBP\\
NI36MBQ,NI37MBP,NI39FBP\\
NI40FBQ,NI41MBP,NI42FBQ\\
NI43FBP,NI44MBQ,NI45FBP\\
NI46FBQ,NI47MBP,NI48FBQ\\
NI49MBP,NI50FBQ,NI51MBP\\
NI52MBQ,NI53FBP,NI54FBQ\\
NI55FBP,NI56MBX,NI57FBQ\\
NI58FBP,NI59FBQ,NI60MBP\\
NI61FBP,NI62MBQ,NI63MBP\\
NI64FBQ,NI65MBP,NI66MBQ\\
NI67MBQ,UI02FAZ,UI03FAZ\\
UI04FAZ,UI05MAZ,UI06MAZ\\
UI07FAZ,UI08MAZ,UI10FAZ\\
UI11FAZ,UI12FAZ,UI13FAZ\\
UI14MAZ,UI15FAZ,UI16MAZ\\
UI17FAZ,UI18MAZ,UI19MAZ\\
UI20MAZ,UI21MAZ,UI22MAZ\\
UI23FAZ,UI24MAZ,UI25FAZ\\
UI26MAZ,UI27FAZ,UI28FAZ\\
UI29FAZ\\
\end{tabular} \\ \midrule
\textbf{Dev} & \multicolumn{1}{c}{\begin{tabular}[c]{@{}c@{}}01NC01FBX, 01NC02FBY, 15NC29FBP\\ 15NC30MBQ, 21NC42MBQ, 43NC61FBQ\end{tabular}} & \begin{tabular}[c]{@{}c@{}}UI01FAZ, UI09MAZ\end{tabular} \\ \midrule
\textbf{Test} & \multicolumn{1}{c}{\begin{tabular}[c]{@{}c@{}}05NC09FAX, 05NC10MAY, 17NC33FBP\\ 17NC34FBQ, 20NC39MBP, 20NC40FBQ\end{tabular}} & \begin{tabular}[c]{@{}c@{}}NI01MAX, NI03FBX\end{tabular} \\ \bottomrule
\end{tabular}
\end{table}

\clearpage
\begin{table}[!htb]
\centering
\caption{Recording distribution in Phase II}
\label{recording-list-phase2}
\begin{tabular}{@{}ccc@{}}
\toprule
\multirow{2}{*}{\textbf{Data}} & \multicolumn{2}{c}{\textbf{Recording list}} \\ \cmidrule{2-3} 
 & \multicolumn{1}{c}{\textbf{Conversation}} & \textbf{Interview} \\ \hline
\textbf{Train} & \multicolumn{1}{c}{\begin{tabular}[c]{@{}c@{}}
02NC03FBX, 02NC04FBY, 03NC05FAX\\ 
03NC06FAY, 04NC07FBX, 04NC08FBY\\ 
06NC11MAX, 06NC12MAY, 07NC13MBP\\ 
07NC14FBQ, 08NC15MBP, 08NC16FBQ \\ 
09NC17FBP, 09NC18MBQ, 10NC19MBP\\ 
10NC20MBQ, 11NC21FBP, 11NC22MBQ \\ 
12NC23FBP, 12NC24FBQ, 13NC25MBP\\ 
13NC26MBQ, 14NC27MBP, 14NC28MBQ \\ 
16NC31FBP, 16NC32FBQ, 18NC35FBP\\ 
18NC36MBQ, 19NC37MBP, 19NC38FBQ \\ 
21NC41MBP, 22NC43FBP, 22NC44MBQ\\ 
23NC35FBQ, 23NC45MBP, 24NC35FBQ \\ 
24NC45MBP, 25NC43FBQ, 25NC47MBP\\ 
26NC48FBP, 26NC49FBQ, 27NC47MBQ \\ 
27NC50FBP, 28NC51MBP, 28NC52FBQ\\ 
29NC53MBP, 29NC54FBQ, 30NC48FBP \\ 
30NC49FBQ, 31NC35FBQ, 31NC50XFB\\ 
32NC36MBQ, 32NC50FBP, 33NC37MBP \\ 
33NC43FBQ, 34NC37MBP, 35NC56MBP\\ 
36NC46FBQ, 37NC45MBP, 38NC50FBP \\ 
39NC57FBX, 40NC58FAY, 41NC59MAX\\ 
42NC60FBQ, 44NC44MBQ, 45NC22MBQ \\ 
46NC41MBP\end{tabular}} & \begin{tabular}[c]{@{}c@{}}
NI02FAX, NI04FBX, NI05MBQ\\ 
NI06FBP, NI07FBQ, NI08FBP\\ 
NI09FBP, NI10FBP, NI12MAP\\ 
NI13MBQ, NI14MBP, NI15FBQ\\ 
NI16FBP, NI17FBQ, NI18MBP\\ 
NI19MBQ, NI20MBP, NI21MBQ\\ 
NI22FBP, NI23FBQ, NI24MBP\\ 
NI25MBQ, NI26FBP, NI27MBQ\\ 
NI28MBP, NI29MBP, NI30MBQ\\ 
NI31FBP, NI32FBQ, NI33MBP\\ 
NI34FBQ, NI35FBP, NI36MBQ\\
NI37MBP, NI39FBP, NI40FBQ\\
NI41MBP, NI42FBQ, NI43FBP\\
NI44MBQ, NI45FBP, NI46FBQ\\
NI47MBP, NI48FBQ, NI49MBP\\ 
NI50FBQ, NI51MBP, NI52MBQ\\
NI53FBP, NI54FBQ, NI55FBP\\
NI56MBX, NI57FBQ, NI58FBP\\
NI59FBQ, NI60MBP, NI61FBP\\
NI62MBQ, NI63MBP, NI64FBQ\\
NI65MBP, NI66MBQ, NI67MBQ\\ 
UI02FAZ, UI03FAZ, UI04FAZ\\ 
UI05MAZ, UI06MAZ, UI07FAZ\\ 
UI08MAZ, UI10FAZ, UI11FAZ\\
UI12FAZ, UI13FAZ, UI14MAZ\\
UI15FAZ, UI16MAZ, UI17FAZ\\
UI18MAZ, UI19MAZ, UI20MAZ\\
UI21MAZ, UI22MAZ, UI23FAZ\\
UI24MAZ, UI25FAZ, UI26MAZ\\
UI27FAZ, UI28FAZ, UI29FAZ\end{tabular} \\ \midrule
\textbf{Dev} & \multicolumn{1}{c}{\begin{tabular}[c]{@{}c@{}}01NC01FBX, 01NC02FBY, 15NC29FBP\\ 15NC30MBQ, 21NC42MBQ, 43NC61FBQ\end{tabular}} & \begin{tabular}[c]{@{}c@{}}UI01FAZ, UI09MAZ\end{tabular} \\ \midrule
\textbf{Test} & \multicolumn{1}{c}{\begin{tabular}[c]{@{}c@{}}05NC09FAX, 05NC10MAY, 17NC33FBP\\ 17NC34FBQ, 20NC39MBP, 20NC40FBQ\end{tabular}} & \begin{tabular}[c]{@{}c@{}}NI01MAX, NI03FBX\end{tabular} \\ \bottomrule
\end{tabular}
\end{table}

\clearpage
\section*{Memory Footprint Analysis}
\label{sec:appendix}
We show the complete results of our memory footprint analysis of word-embeddings, meta-embeddings, and multilingual models in Table \ref{tab:memory_all}.
\begin{table*}[!ht]
\centering
\caption{Memory footprint (MB) for storing the activations for a given sequence length.}
\resizebox{0.98\textwidth}{!}{
\begin{tabular}{l|rrrrrrrrr}
\toprule
\multirow{2}{*}{\textbf{Model}} &  \multicolumn{9}{c}{\textbf{Activation (MB)}} \\
 & \textbf{16}   &    \textbf{32}  & \textbf{64}  & \textbf{128} & \textbf{256} & \textbf{512} & \textbf{1024} & \textbf{2048} & \textbf{4096} \\
\midrule
FastText   &   1.0 &   2.0 &    4.0 &   10.0 &   26.0 &    79.0 &  261.0 &   941.0 &  3547.0 \\
Linear  &   1.0 &   2.0 &    4.0 &   10.0 &   27.4 &    80.8 &  265.6 &   950.0 &  3562.0 \\
Concat   &   1.0 &   2.0 &    5.0 &   11.2 &   29.2 &    85.2 &  274.5 &   967.5 &  3596.5 \\
Attention (MME)  &   1.0 &   2.0 &    5.4 &   12.4 &   31.0 &    89.0 &  283.2 &   985.6 &  3630.6 \\
HME  &   3.2 &   6.6 &   13.4 &   28.6 &   64.2 &   154.8 &  416.4 &  1252.0 &  4155.0 \\
Scratch (2L) &   2.0 &   4.0 &    8.0 &   20.0 &   46.0 &   133.0 &         - &          - &          - \\
Scratch (4L)  &   3.0 &   7.0 &   15.0 &   38.0 &   90.0 &   264.0 &         - &          - &          - \\
mBERT (uncased) &  10.0 &  20.0 &   41.0 &  100.0 &  218.0 &   597.0 &         - &          - &          - \\
XLM-R$_{\text{BASE}}$ &  10.0 &  20.0 &   41.0 &  100.0 &  218.0 &   597.0 &         - &          - &          - \\
XLM-R$_{\text{LARGE}}$ &  25.0 &  52.0 &  109.0 &  241.0 &  579.0 &  1541.0 &         - &          - &          - \\
XLM-MLM$_{\text{LARGE}}$ &  20.0 &  42.0 &   89.0 &  193.0 &  467.0 &  1158.0 &         - &          - &          - \\
\bottomrule
\end{tabular}
}
\label{tab:memory_all}
\end{table*}

\end{document}